\documentclass[10pt,twocolumn,letterpaper]{article}

\usepackage{cvpr}              
\usepackage{graphicx}
\usepackage{amsmath}
\usepackage{amssymb}
\usepackage{booktabs}

\usepackage{acronym}
\usepackage{caption}
\usepackage{subcaption}
\usepackage{cuted}
\usepackage{adjustbox}

\usepackage{siunitx}
\usepackage{booktabs,makecell}
\usepackage{etoolbox}
\usepackage{color, colortbl}

\usepackage[ruled,vlined]{algorithm2e}

\usepackage{setspace}

\usepackage[title]{appendix}

\usepackage{enumitem}
\setlist{topsep=0pt, leftmargin=*,noitemsep,topsep=0pt,parsep=0pt,partopsep=0pt}

\usepackage[pagebackref,breaklinks,colorlinks]{hyperref}

\usepackage[capitalize]{cleveref}
\crefname{section}{Sec.}{Secs.}
\Crefname{section}{Section}{Sections}
\Crefname{table}{Table}{Tables}
\crefname{table}{Tab.}{Tabs.}

\acrodef{nec}[NEC]{normal epipolar constraint}
\acrodef{pnec}[PNEC]{probabilistic normal epipolar constraint}
\acrodef{klt}[KLT]{Kanade-Lucas-Tomasi}
\acrodef{grq}[GRQ]{generalized Rayleigh quotient}
\acrodef{scf}[SCF]{self-consistent-field}
\acrodef{irls}[IRLS]{iteratively reweighted least squares}
\acrodef{dnls}[DNLS]{differentiable nonlinear least squares}
\acrodef{ransac}[RANSAC]{random sample consensus}

\newcommand{\pseudoparagraph}[1]{\textbf{#1}}

\usepackage{xcolor}
\definecolor{Gray}{gray}{0.96}

\def\cvprPaperID{1932} \def\confName{CVPR}
\def\confYear{2023}

\newcommand\blfootnote[1]{  \begingroup
  \renewcommand\thefootnote{}\footnote{#1}  \addtocounter{footnote}{-1}  \endgroup
}

\begin{document}

\title{Learning Correspondence Uncertainty\\[1mm]via Differentiable Nonlinear Least Squares}
\author{Dominik Muhle$^{1,2}$ \hspace{1cm} Lukas Koestler$^{1,2}$ \hspace{1cm} Krishna Murthy Jatavallabhula$^4$ \hspace{1cm} Daniel Cremers$^{1,2,3}$\\
$^1$TU Munich \hspace{1cm} $^2$Munich Center for Machine Learning \hspace{1cm} $^3$University of Oxford \hspace{1cm} $^4$MIT \\
{\tt\small \{dominik.muhle, lukas.koestler, cremers\}@tum.de \hspace{.5cm} jkrishna@mit.edu}
\vspace{-0.5cm}
}
\maketitle

\begin{strip}
\vspace{-0.7cm}
\centering
\captionsetup{type=figure}
\includegraphics[trim={0cm 22.5cm 0cm 0cm},clip,width=\linewidth]{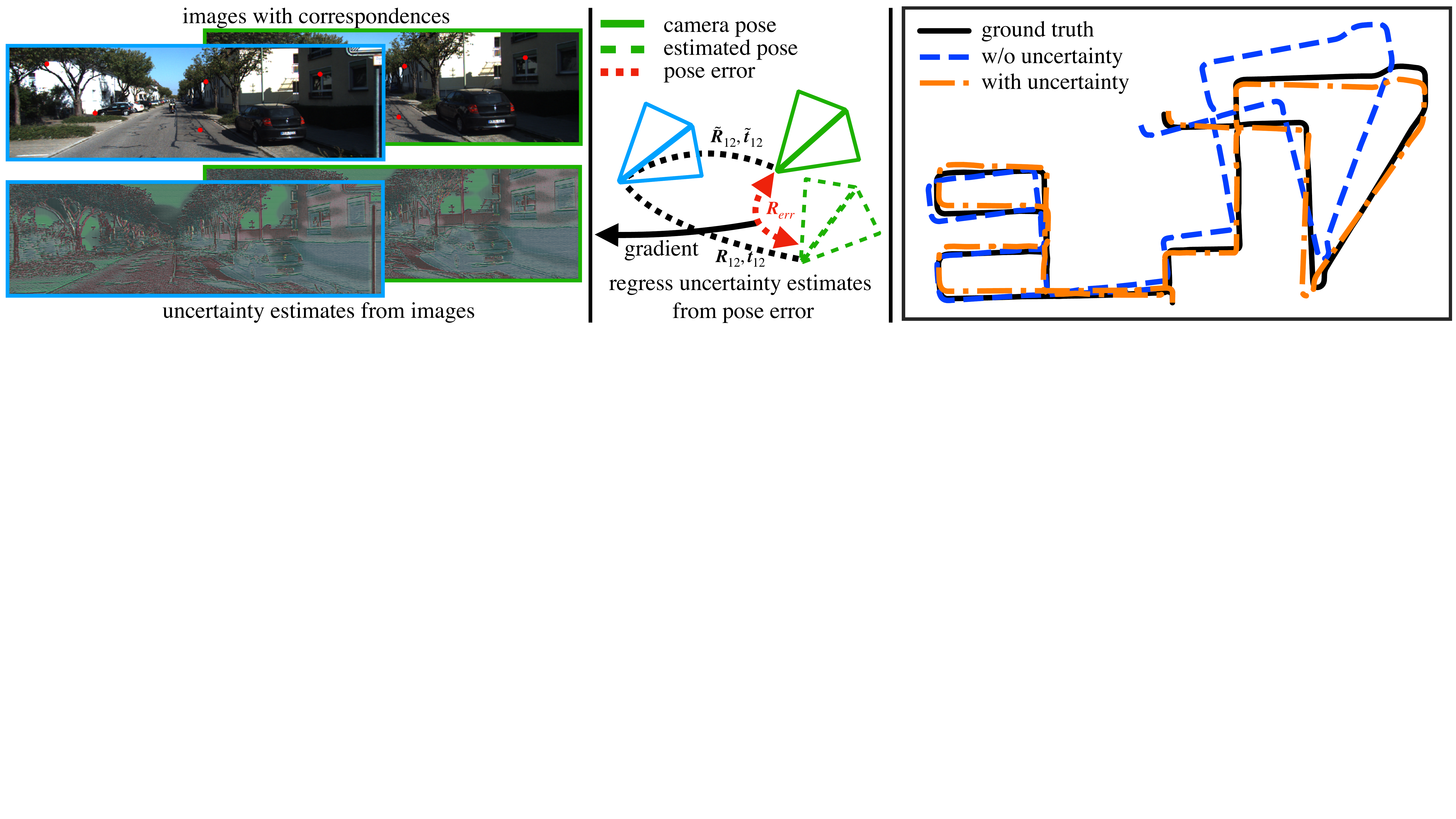}
\vspace{-0.7cm}
\captionof{figure}{
We present a \textbf{\ac{dnls} framework for learning feature correspondence quality} by computing per-feature positional uncertainty. 
The uncertainty estimates (left, bottom images) are regressed from a pose estimation error (middle), enabling the framework across a range of (handcrafted, learned) feature extractors.
Our learned covariances (right, orange trajectory) improve orientation estimation by up to $11\%$ over state-of-the-art probabilistic pose estimation methods on the KITTI dataset~\cite{Geiger2012CVPR}.
}
\label{fig:teaser}
\vspace{-0.2cm}
\end{strip}

\begin{abstract}
We propose a differentiable nonlinear least squares framework to account for uncertainty in relative pose estimation from feature correspondences. Specifically, we introduce a symmetric version of the probabilistic normal epipolar constraint, and an approach to estimate the covariance of feature positions by differentiating through the camera pose estimation procedure.
We evaluate our approach on synthetic, as well as the KITTI and EuRoC real-world datasets.
On the synthetic dataset, we confirm that our learned covariances accurately approximate the true noise distribution.
In real world experiments, we find that our approach consistently outperforms state-of-the-art non-probabilistic and probabilistic approaches, regardless of the feature extraction algorithm of choice.
\end{abstract}

\blfootnote{
\href{https://dominikmuhle.github.io/dnls_covs/}{Project Page}
} 

\vspace{-0.7cm}
\section{Introduction}
Estimating the relative pose between two images given mutual feature correspondences is a fundamental problem in computer vision. It is a key component of structure from motion (SfM) and visual odometry (VO) methods which in turn fuel a plethora of applications from autonomous vehicles or robots to augmented and virtual reality.
\begin{figure*}[t]
    \centering
    \begin{subfigure}[b]{0.44\textwidth}
        \centering
        {\includegraphics[trim={0cm 0cm 0cm 0cm},clip,width=\textwidth]{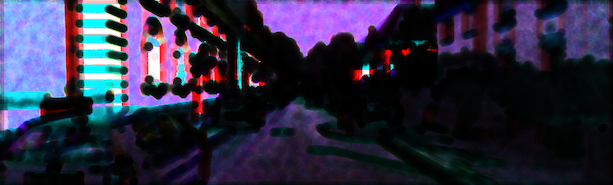}}
        \caption{covariances from \cite{muhle2022pnec} per pixel}
        \label{fig:dense_klt}
    \end{subfigure}
    \begin{subfigure}[b]{0.44\textwidth}
        \centering
        {\includegraphics[trim={0cm 0cm 0cm 0cm},clip,width=\textwidth]{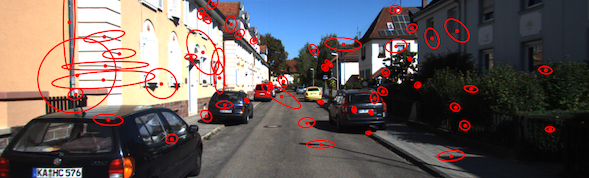}}
        \caption{points and covariances \cite{muhle2022pnec}}
        \label{fig:key_klt}
    \end{subfigure}
    \vspace{0.0cm}
    \begin{subfigure}[b]{0.44\textwidth}
        \centering
        {\includegraphics[trim={0cm 0cm 0cm 0cm},clip,width=\textwidth]{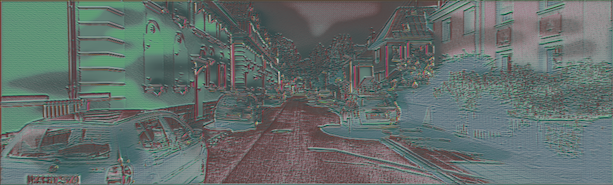}}
        \caption{learned covariances per pixel (Ours)}
        \label{fig:dense}
    \end{subfigure}
    \begin{subfigure}[b]{0.44\textwidth}
        \centering
        {\includegraphics[trim={0cm 0cm 0cm 0cm},clip,width=\textwidth]{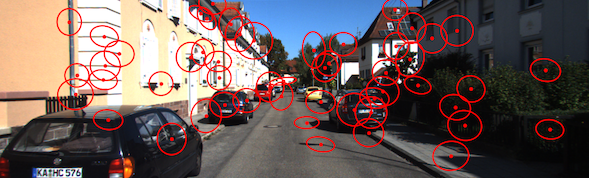}}
        \caption{points and learned covariances (Ours)}
        \label{fig:key}
    \end{subfigure}
    \vspace{-0.3cm}
    \caption{
        Comparison between covariances used in \cite{muhle2022pnec} (first row) and our learned covariances (second row). The first column shows a dense color coded ($s, \alpha, \beta$ mapped to HLS with $\gamma$ correction) representation for each pixel, while the second column shows subsampled keypoints and their corresponding (enlarged) covariances. The higher saturation in (a) shows that the covariances are more anisotropic. The learned covariances (c) show a more fine-grained detail in the scale (brightness) and less blurring than the covariances in (a). 
    }
\label{fig:cov_comp}
\vspace{-0.6cm}
\end{figure*}

Estimating the relative pose -- rotation and translation -- between two images, is often formulated as a geometric problem that can be solved by estimating the essential matrix~\cite{longuet1987readings} for calibrated cameras, or the fundamental matrix~\cite{Hartley2004} for uncalibrated cameras. Related algorithms like the eight-point algorithm~\cite{longuet1987readings,hartley1997defense}  provide fast solutions.
However, essential matrix based approaches suffer issues such as {\em solution multiplicity}~\cite{Faugeras1989mul_solutions, Hartley2004} and {\em planar degeneracy} \cite{EigenNEC_Kneip2013}.
The \acf{nec}\cite{OrigNEC_Kneip2012} addresses the issues by estimating the rotation independently of the translation, leading to more accurate relative poses \cite{EigenNEC_Kneip2013}. 

Neither of the aforementioned algorithms takes into account the \emph{quality} of feature correspondences -- an important cue that potentially improves pose estimation accuracy. Instead, feature correspondences are classified into inliers and outliers through a RANSAC scheme \cite{MRO_Chng2020}. However, keypoint detectors \cite{ORB_Rublee2011, detone2018superpoint} for feature correspondences or tracking algorithms \cite{KLT_Tomasi1991} yield imperfect points \cite{lindenberger2021pixsfm} that exhibit a richer family of error distributions, as opposed to an inlier-outlier distribution family. Algorithms, that make use of feature correspondence quality have been proposed for essential/fundamental matrix estimation \cite{pro_cov_Brooks2001, ranftl2018deep} and for the \ac{nec} \cite{muhle2022pnec}, respectively. 

While estimating the relative pose can be formulated as a classical optimization problem \cite{EigenNEC_Kneip2013, DSO_Engel2016}, the rise in popularity of deep learning has led to several works augmenting VO or visual simultaneous localisation and mapping (VSLAM) pipelines with learned components. GN-Net \cite{von2020gn} learns robust feature representations for direct methods like DSO \cite{DSO_Engel2016}. For feature based methods Superpoint \cite{detone2018superpoint} provides learned features, while Superglue \cite{sarlin2020superglue} uses graph neural networks to find corresponding matches between feature points in two images. DSAC introduces a differential relaxation to RANSAC that allows gradient flow through the otherwise non-differentiable operation. In \cite{ranftl2018deep} a network learns to re-weight correspondences for estimating the fundamental matrix. PixLoc \cite{sarlin2021back} estimates the pose from an image and a 3D model based on direct alignment. 

In this work we combine the predictive power of deep learning with the precision of geometric modeling for highly accurate relative pose estimation. Estimating the noise distributions for the feature positions of different feature extractors allows us to incorporate this information into relative pose estimation. Instead of modeling the noise for each feature extractor explicitly, we present a method to learn these distributions from data, using the same domain that the feature extractors work with - images. We achieve this based on the following technical contributions:
\begin{itemize}
    \item We introduce a symmetric version of the \acf{pnec}, that more accurately models the geometry of relative pose estimation with uncertain feature positions.
    \item We propose a learning strategy to minimize the relative pose error by learning feature position uncertainty through \acf{dnls}, see \autoref{fig:teaser}.
    \item We show with synthetic experiments, that using the gradient from the relative pose error leads to meaningful estimates of the positional uncertainty that reflect the correct error distribution.
    \item We validate our approach on real-world data in a visual odometry setting and compare our method to non-probabilistic relative pose estimation algorithms, namely Nist\'er 5pt~\cite{EM_Nister2003}, and \ac{nec}~\cite{EigenNEC_Kneip2013}, as well as to the \ac{pnec} with non-learned covariances \cite{muhle2022pnec}. 
    \item We show that our method is able to generalize to different feature extraction algorithms such as SuperPoint \cite{detone2018superpoint} and feature tracking approaches on real-world data.
    \item We release the code for all experiments and the training setup to facilitate future research.
\end{itemize}

\section{Related Work}
This work is on deep learning for improving frame-to-frame relative pose estimation by incorporating feature position uncertainty with applications to visual odometry. We therefore restrict our discussion of related work to relative pose estimation in visual odometry, weighting correspondences for relative pose estimation, and deep learning in the context of VSLAM. For a broader overview over VSLAM we refer the reader to more topic-specific overview papers~\cite{triggs1999bundle,cadena2016past} and to the excellent books by Hartley and Zisserman~\cite{Hartley2004} and by Szeliski~\cite{szeliski2010computer}.

\pseudoparagraph{Relative Pose Estimation in Visual Odometry.}
Finding the relative pose between two images has a long history in computer vision, with the first solution for perspective images reaching back to 1913 by Kruppa~\cite{kruppa1913ermittlung}. Modern methods for solving this problem can be classified into \emph{feature-based} and \emph{direct} methods. The former rely on feature points extracted in the images together with geometric constraints like the \emph{epipolar constraint} or the \emph{normal epipolar constraint} \cite{OrigNEC_Kneip2012} to calculate the relative pose. The latter optimize the pose by directly considering the intensity differences between the two images and rose to popularity with LSD-SLAM~\cite{engel14eccv} and DSO\cite{DSO_Engel2016}. Since direct methods work on the assumption of brightness or irradiance constancy they require the appearance to be somewhat similar across images. In turn, keypoint based methods rely on suitable feature extractors which can exhibit significant amounts of noise and uncertainty. In this paper we propose a method to learn the intrinsic noise of keypoint detectors -- therefore, the following will focus on feature based relative pose estimation.

One of the most widely used parameterizations for reconstructing the relative pose from feature correspondences is the essential matrix, given calibrated cameras, or the fundamental matrix in the general setting. Several solutions based on the essential matrix have been proposed ~\cite{longuet1987readings,EM_Nister2003,stewenius2006recent,li2006five,kukelova2008polynomial}.
They include the linear solver by Longuet-Higgins \cite{longuet1987readings}, requiring 8 correspondences, or the solver by Nist{\'e}r \etal~\cite{Nistr2004VisualO} requiring the minimal number of 5 correspondences. However, due to their construction, essential matrix methods deteriorate for purely rotational motion with noise-free correspondences \cite{EigenNEC_Kneip2013}.
As an alternative, methods that do not use the essential matrix have been proposed -- they either estimate the relative pose using quaternions \cite{fathian2018quest} or make use of the \acf{nec} by Kneip and Lynen~\cite{OrigNEC_Kneip2012, EigenNEC_Kneip2013}. The latter addresses the problems of the essential matrix by estimating rotation independent of the translation. \cite{Briales2018globalNEC} shows how to obtain the global minimum for the \ac{nec}. Further work, that disentangles rotation and translation can be found in \cite{lim2010estimating}.

\pseudoparagraph{Weighting of Feature Correspondences.}
Keypoints in images can exhibit significant noise, deteriorating the performance for pose estimation significantly \cite{germain2020s2dnet}. The noise characteristics of the keypoint positions depend on the feature extractor. For \ac{klt}~tracking~\cite{KLT_Lukas1981,KLT_Tomasi1991} approaches, the position uncertainty has been investigated in several works \cite{forstner1987fast,Steele2005Foerster,sheorey2014uncertainty,zhang2017uncertainty}. The uncertainty was directly integrated into the tracking in \cite{dorini2011unscented}. \cite{SIFTCOV_Zeisl2009} proposed a method to obtain anisotropic and inhomogeneous covariances for SIFT~\cite{lowe2004distinctive} and SURF~\cite{bay2006surf}. 

\begin{figure*}[t]
\begin{center}
    \includegraphics[trim={0.0cm 18.9cm 0.0cm 0.1cm},clip,width=0.92\textwidth]{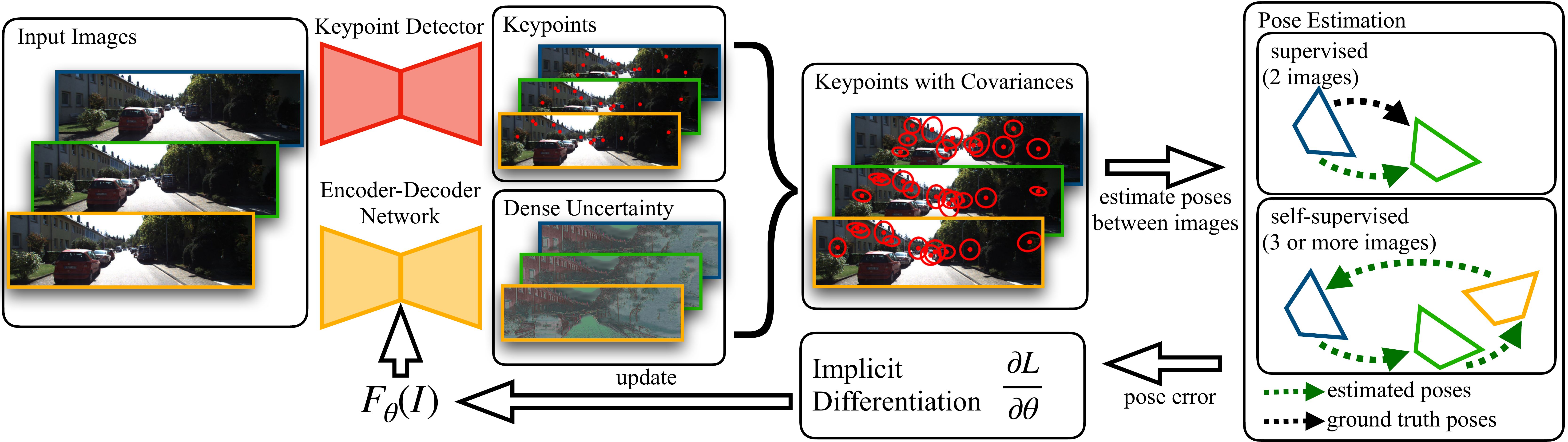}
    \vspace{-0.6cm}
\end{center}
    \caption{\textbf{Architecture}: We extract the uncertainty per image for every pixel using a UNet \cite{ronneberger2015u} backbone. Using keypoint locations from a keypoint detector, we obtain the keypoints with their estimated positional uncertainty. The relative pose is then estimated using a \ac{dnls} optimization. The UNet is updated by backpropagating the gradient (obtained by implicit differentiation)
        to the network output.}
    \vspace{-0.6cm}
\label{fig:architecture}
\end{figure*}

Given the imperfect keypoint positions, not all correspondences are equally well suited for estimating the relative pose. \cite{germain2020s2dnet} showed the effect of the noise level on the accuracy of the pose estimation. Limiting the influence of bad feature correspondences has been studied from a geometrical and a probabilistic perspective. 
\ac{ransac} \cite{fischler1981random} is a popular method to classify datapoints into inliers and outliers that can be easily integrated into feature based relative pose estimation pipelines. 
Ranftl \etal~\cite{ranftl2018deep} relax the hard classification for inlier and outlier and use deep learning to find a robust fundamental matrix estimator in the presence of outliers in an \ac{irls} fashion. DSAC \cite{brachmann2017dsac} models \ac{ransac} as a probabilistic process to make it differentiable.
Other lines of work integrate information about position uncertainty directly into the alignment problem. For radar based SLAM, \cite{burnett2021radar} incorporates keypoint uncertainty in radar images, with a deep network predicting the uncertainty. 
Image based position uncertainty was investigated from the statistical, \cite{kanatani2004geometric,kanatani2008statistical}, the photogrammetry \cite{meidow2009reasoning} and the computer vision perspective \cite{pro_cov_Brooks2001,con_cov_Kanazawa2001}. \cite{pro_cov_Brooks2001} and \cite{con_cov_Kanazawa2001} debated the benefit of incorporating position uncertainty into fundamental matrix estimation. 
We base our method on the \acf{pnec} \cite{muhle2022pnec}, that improved on the \ac{nec} by extending it to a probabilistic view. It achieved better results on real-world data with covariances approximated using the Boltzmann distribution~\cite{DBLP:journals/jei/BishopN07}. 
We expand on this idea by learning covariances (see \autoref{fig:cov_comp}) agnostic of the keypoints extractor used to further improve pose estimation.

\pseudoparagraph{Deep Learning in VSLAM.}
Deep Learning has transformed computer vision in the last decade. While deep networks have been successfully used for tasks like detection \cite{redmon2016you}, semantic segmentation \cite{long2015fully}, and recently novel view synthesis \cite{mildenhall2021nerf}, they have also found application in VSLAM pipelines. DVSO \cite{yang2018dvso} and D3VO \cite{yang20d3vo} leveraged deep learning to improve the precision for direct methods, while GN-Net \cite{von2020gn} predicts robust and dense feature maps. Several works proposed to learn keypoint extractors, for feature based pose estimation, such as SuperPoint \cite{detone2018superpoint} and LIFT \cite{yi2016lift}.
 SuperGlue \cite{sarlin2020superglue} enabled feature matching with graph neural networks. Other lines of work leverage deep learning for localization by making parts of the pose estimation pipeline differentiable \cite{torii201524, sarlin2021back, brachmann2017dsac, arandjelovic2016netvlad}. Works, that directly predicting the pose include PoseNet \cite{kendall2015posenet} and CTCNet \cite{ctcnet} that uses self-supervised learning with a cycle-consistency loss for VO. \cite{lindenberger2021pixsfm} learns image representations by refining keypoint positions and camera poses in a post-processing step of a structure-from-motion pipeline. $\nabla$SLAM \cite{jatavallabhula2020slam} presents a differentiable dense SLAM system with several components (e.g., the Levenberg-Marquardt~\cite{levenberg1944method,marquardt1963algorithm} optimizer).

 \section{Method}
 In the following, we present our framework to estimate positonal uncertainty of feature points by leveraging \ac{dnls}. We learn the noise covariances through a forward and backward step. In the forward step, the covariances are used in a probabilistic pose estimation optimization, namely the \ac{pnec}. In the backward step, the gradient from the pose error is back-propagated through the optimization to the covariances. From there we can train a neural network to predict the keypoint position uncertainty from the images. We start by summarizing the asymmetric \ac{pnec} \cite{muhle2022pnec} and for the first time introduce its symmetric counterpart.

\subsection{Prerequisites}
 \pseudoparagraph{Notation.}
 We follow the notation of \cite{muhle2022pnec}. Bold lowercase letters (\eg $\boldsymbol{f}$) denote vectors, whereas bold uppercase letters (\eg  $\boldsymbol{\Sigma}$) denote matrices. $\hat{\boldsymbol{u}} \in \mathbb{R}^{3\times 3}$ represents the skew-symmetric matrix of the vector $\boldsymbol{u} \in \mathbb{R}^3$ such that the cross product between two vectors can be rewritten as a matrix-vector operation,~i.e.~$\boldsymbol{u} \times \boldsymbol{v} = \hat{\boldsymbol{u}} \boldsymbol{v}$. The transpose is denoted by the superscript $^\top$. We deviate from \cite{muhle2022pnec} in the following: variables of the second frame are marked with the ${}^\prime$ superscript, while variables of the first frame do not have a superscript. We represent the relative pose between images as a rigid-body transformation consisting of a rotation matrix $\boldsymbol{R} \in SO(3)$ and a unit length translation $\boldsymbol{t} \in \mathbb{R}^3$ ($\|\boldsymbol{t}\| = 1$ is imposed due to scale-invariance). 

\subsection{The Probabilistic Normal Epipolar Constraint}
 The asymetric \acf{pnec} estimates the relative pose, give two images $\boldsymbol{I}, \boldsymbol{I}^\prime$ of the same scene under the assumption of uncertain feature positions in the second image. A feature correspondences is given by $\boldsymbol{p}_{i}, \boldsymbol{p}^\prime_{i}$ in the image plane, where the uncertainty of $\boldsymbol{p}^\prime_{i}$ is represented by the corresponding covariance $\boldsymbol{\Sigma}_{2\text{D}, i}^\prime$. To get the epipolar geometry for the \ac{pnec} the feature points are unprojected using the camera intrinsics, giving unit length bearing vectors $\boldsymbol{f}_{i}, \boldsymbol{f}^\prime_{i}$. The uncertainty of $\boldsymbol{f}^\prime_{i}$ is now represented by $\boldsymbol{\Sigma}_i^\prime$. 
 Estimating the relative pose is done by minimizing the \ac{pnec} cost function as defined in \cite{muhle2022pnec}. For convenience we recap the energy function 
 \begin{equation} \label{eqn:pnec-cost}
      E(\boldsymbol{R}, \boldsymbol{t}) = \sum_i \frac{e_i^2}{\sigma_i^2} = \sum_i \frac{| \boldsymbol{t}^\top (\boldsymbol{f}_i \times \boldsymbol{R} \boldsymbol{f}^\prime_i) |^2}{\boldsymbol{t}^\top \hat{\boldsymbol{f}_i} \boldsymbol{R} \boldsymbol{\Sigma}^\prime_i \boldsymbol{R}^\top \hat{\boldsymbol{f}_i}{}^\top \boldsymbol{t}} \,,
 \end{equation}
 in our notation. As mentioned previously, this asymmetric \ac{pnec} in \cite{muhle2022pnec} only considers uncertainties $\boldsymbol{\Sigma}^\prime$ in the second frame. While this assumption might hold for the \ac{klt} tracking \cite{Basalt_Usenko2020} used in \cite{muhle2022pnec}, this leaves out important information when using other keypoint detectors like ORB \cite{ORB_Rublee2011} or SuperPoint \cite{detone2018superpoint}. Therefore, we will introduce a symmetric version of the \ac{pnec} that is more suitable for our task in the following. 

\pseudoparagraph{Making the PNEC symmetric.}
 As in \cite{muhle2022pnec} we assume the covariance of the bearing vectors $\boldsymbol{f}_i$ and $\boldsymbol{f}^\prime_i$ to be gaussian, their covariance matrices denoted by $\boldsymbol{\Sigma}_i$ and $\boldsymbol{\Sigma}_i^\prime$, respectively. The new variance can be approximated as
 \begin{equation}
     \sigma_{s,i}^2 = \boldsymbol{t}^\top (\hat{(\boldsymbol{R} \boldsymbol{f}^\prime_i)} \boldsymbol{\Sigma}_i \hat{(\boldsymbol{R} \boldsymbol{f}^\prime_i)}{}^\top + \hat{\boldsymbol{f}_i} \boldsymbol{R} \boldsymbol{\Sigma}^\prime_i \boldsymbol{R}^\top \hat{\boldsymbol{f}_i}{}^\top) \boldsymbol{t}\, .
 \end{equation} 
 In the supplementary material (see \autoref{sec:approx}), we derive the variance and show the validity of this approximation given the geometry of the problem. This new variance now gives us the new symmetric \ac{pnec} with its following energy function
 \begin{equation} \label{eqn:symm-pnec-cost}
     E_{s}(\boldsymbol{R}, \boldsymbol{t}) = \sum_i \frac{e_i^2}{\sigma_{s,i}^2}
 \end{equation}

\subsection{\ac{dnls} for Learning Covariances}
 We want to estimate covariances $\boldsymbol{\Sigma}_{2\text{D}}$ and $\boldsymbol{\Sigma}_{2\text{D}}^\prime$ (in the following collectively denoted as $\boldsymbol{\Sigma}_{2\text{D}}$ for better readability) in the image plane 
 \begin{equation}
    \boldsymbol{\Sigma}_{2\text{D}} = \arg \min_{\boldsymbol{\Sigma}_{2\text{D}}} \mathcal{L}\,,
 \end{equation}
 such that they minimize a loss function $\mathcal{L}$ of the estimated pose. Since we found that the rotational error of the \ac{pnec} is more stable than the translational error, we chose to minimize only the rotational error
 \begin{align}
     e_{\text{rot}} & = \angle  \tilde{\boldsymbol{R}}{}^\top \boldsymbol{R} \\
     \mathcal{L}(\tilde{\boldsymbol{R}}, \boldsymbol{R}; \boldsymbol{\Sigma}_{2\text{D}}) & = e_{\text{rot}}
 \end{align}
 between the ground truth rotation $\tilde{\boldsymbol{R}}$ and the estimated rotation $\boldsymbol{R}$. We obtain
 \begin{equation}
      \boldsymbol{R} = \arg \min_{\boldsymbol{R}} E_{s}(\boldsymbol{R}, \boldsymbol{t}; \boldsymbol{\Sigma}_{2\text{D}})
 \end{equation}
 by minimizing \autoref{eqn:symm-pnec-cost}. To learn the covariances that minimize the rotational error, we can follow the gradient $d \mathcal{L} / d \boldsymbol{\Sigma}_{2\text{D}}$. Implicit differentiation allows us to compute the gradient as
 \begin{equation} \label{eqn:grad}
     \frac{d \mathcal{L}}{d \boldsymbol{\Sigma}_{2\text{D}}} = - \frac{\partial^2 E_s}{\partial \boldsymbol{\Sigma}_{2\text{D}} \partial \boldsymbol{R}{}^\top} \left(\frac{\partial^2 E_s}{\partial \boldsymbol{R} \partial \boldsymbol{R}{}^\top} \right)^{-1} \frac{e_{\text{rot}}}{\partial \boldsymbol{R}} \,.
 \end{equation}
 For a detailed derivation of \autoref{eqn:grad} and other methods, that unroll the optimization, to obtain the gradient we refer the interested reader to \cite{domke2012generic}.

\pseudoparagraph{Supervised Learning.}
 The goal of the paper is for a neural network $F$ learn the noise distributions of a keypoint detector. Given an image and a keypoint position, the network should predict the covariance of the noise $\boldsymbol{\Sigma}_{2\text{D},i} = F(\boldsymbol{I}, \boldsymbol{p}_{i})$. The gradient $d \mathcal{L} / d \boldsymbol{\Sigma}_{2\text{D}}$ allows for the network to learn the covariance matrices in an end-to-end manner by regression on the relative pose error. Given a dataset with know ground truth poses, we can use
 \begin{equation}
     \mathcal{L}_{\text{sup}} = e_{\text{rot}}
 \end{equation}
 as a training loss. This ensures, that learned covariances effectively minimize the rotational error. See \autoref{fig:architecture} for overview of the training process.
 \begin{figure}[t]
    \centering
    \begin{subfigure}[b]{0.235\textwidth}
        \centering
        {\includegraphics[trim={0.23cm 0.25cm 0.27cm 0.23cm},clip,width=\textwidth]{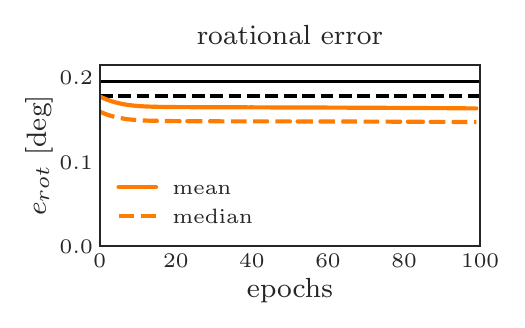}}
        \caption{$e_{\text{rot}}$ during training}
        \label{fig:synth_error}
    \end{subfigure}
    \vspace{0.0cm}
    \begin{subfigure}[b]{0.235\textwidth}
        \centering
        {\includegraphics[trim={0.23cm 0.25cm 0.27cm 0.21cm},clip,width=\textwidth]{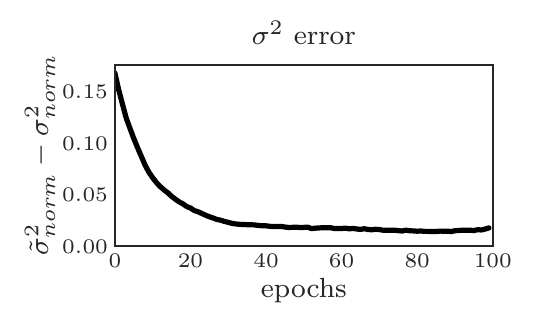}}
        \caption{$\sigma^2_{norm}$ error during training}
        \label{fig:synth_sigma}
    \end{subfigure}
    \vspace{-0.65cm}
    \caption{
        Rotational error (a) and differences between the true residual variance $\tilde{\sigma}^2$ and the learned variance $\sigma^2$ (b) over the training epochs. Starting from uniform covariances, our method adapts the covariances for each keypoint to minimize the rotational error. Simultaneously, this leads to a better estimate of $\sigma^2$.
    }
\label{fig:synth}
\vspace{-0.35cm}
\end{figure}

\begin{figure}[t]
    \centering
    \begin{subfigure}[b]{0.478\textwidth}
        \centering
        {\includegraphics[trim={0cm 0.2cm 0cm 0.2cm},clip,width=\textwidth]{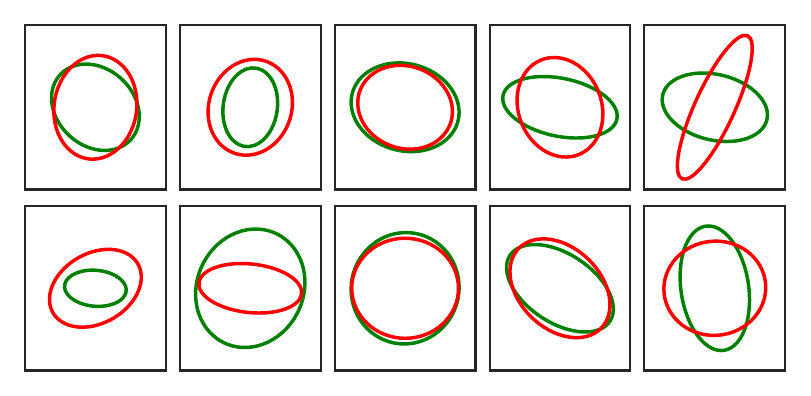}}
        \caption{covariance matrices in the 1st frame after 100 epochs training}
        \label{fig:synth_host_covs}
    \end{subfigure}
    \vspace{0.0cm}
    \begin{subfigure}[b]{0.478\textwidth}
        \centering
        {\includegraphics[trim={0cm 0.2cm 0cm 0.2cm},clip,width=\textwidth]{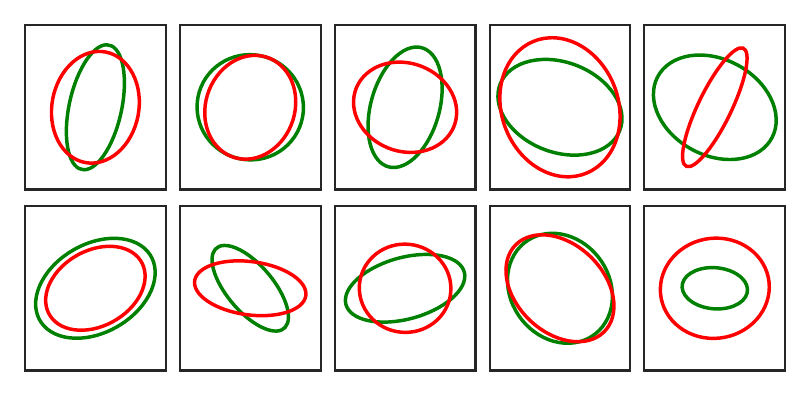}}
        \caption{covariance matrices in the 2nd frame after 100 epochs training}
        \label{fig:synth_target_covs}
    \end{subfigure}
    \vspace{-0.65cm}
    \caption{
        Estimated (red) covariance ellipses in the first (a) and the second (b) frame, learned from 128\,000 examples. Ground truth (green) covariances as comparison. Although the gradient minimizes the rotational error (see \autoref{fig:synth_error}), it is not capable of learning the correct covariance in the image plane. 
             }
\label{fig:synth_covs}
\vspace{-0.45cm}
\end{figure}

\pseudoparagraph{Self-Supervised Learning.}
 Finding a suitable annotated dataset for a specific task is often non-trivial. For our task, we need accurate ground truth poses that are difficult to acquire. But given a stream of images, like in VO, our method can be adapted to train a network in a self-supervised manner without the need for ground truth poses. For this, we follow the approach of \cite{ctcnet} to exploit the cycle-consistency between a tuple of images. The cycle-consistency loss for a triplet $\{\boldsymbol{I}_1, \boldsymbol{I}_2, \boldsymbol{I}_3\}$ of images is given by
 \begin{equation}
     \mathcal{L}_{\text{cycl}} = \angle \prod_{(i,j) \in \mathcal{P}} \boldsymbol{R}_{ij} \, ,
 \end{equation}
 where $\boldsymbol{R}_{ij}$ is the estimated rotation between images $I_i$ and $I_j$ and $\mathcal{P} = \{(1, 2), (2, 3), (3, 1)\}$ defines the cycle. As in \cite{ctcnet}, we also define an anchor loss
 \begin{equation}
     \mathcal{L}_{\text{anchor}} = \sum_{(i,j) \in \mathcal{P}} \angle \boldsymbol{R}_{ij} \boldsymbol{R}_{ij, \text{NEC}}^\top 
 \end{equation}
 with the \ac{nec} rotation estimate, as a regularising term. In contrast to \cite{ctcnet}, our method does not risk learning degenerate solutions from the cycle-consistency loss, since the rotation is estimated using independently detected keypoints. The final loss is then given by 
 \begin{equation}
     \mathcal{L}_{\text{self}} =\mathcal{L}_{\text{cycl}} + \lambda \mathcal{L}_{\text{anchor}}\, .
 \end{equation}

 \section{Experiments}
 We evaluate our method in both synthetic and real-world experiments. Over the synthetic data, we investigate the ability of the gradient to learn the underlying noise distribution correctly by overfitting covariance estimates directly. We also investigate if better noise estimation leads to a reduces rotational error. 

On real-world data, we use the gradient to train a network to predicts the noise distributions from images for different keypoint detectors. We explore fully supervised and self-supervised learning techniques for SuperPoint \cite{detone2018superpoint} and Basalt \cite{Basalt_Usenko2020} KLT-Tracks to verify that our method is agnostic to the type of feature descriptor used (classical vs learned). We evaluate the performance of the learned covariances in a visual odometry setting on the popular KITTI odometry and the EuRoC dataset. We also evaluate generalization capabilities from the KITTI to the EuRoC dataset.

For our experiments we implement \autoref{eqn:symm-pnec-cost} in both Theseus \cite{pineda2022theseus} and ceres \cite{ceres-solver}.
 We use the Theseus implementation to train our network, since it allows for batched optimization and provides the needed gradient (see \autoref{eqn:grad}). However, we use the ceres implementation for our evaluation. We found the Levenberg-Marquardt optimization of ceres to be faster and more stable than its theseus counterpart.
 \begin{figure}[t]
    \centering
    \begin{subfigure}[b]{0.235\textwidth}
        \centering
        {\includegraphics[trim={0.23cm 0.25cm 0.27cm 0.23cm},clip,width=\textwidth]{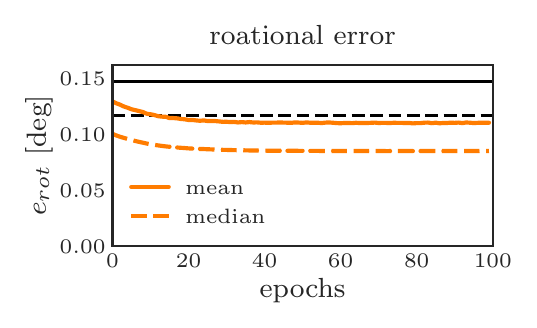}}
        \caption{$e_{\text{rot}}$ during training}
        \label{fig:synth_error2}
    \end{subfigure}
    \vspace{0.0cm}
    \begin{subfigure}[b]{0.235\textwidth}
        \centering
        {\includegraphics[trim={0.23cm 0.25cm 0.27cm 0.21cm},clip,width=\textwidth]{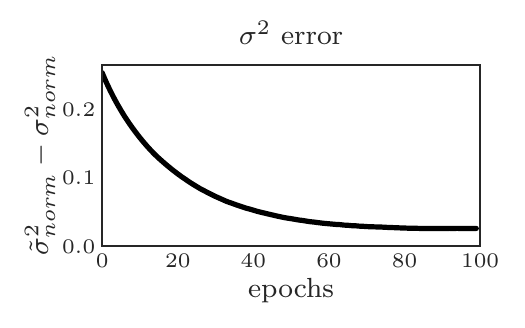}}
        \caption{$\sigma^2_{norm}$ error during training}
        \label{fig:synth_sigma2}
    \end{subfigure}
    \vspace{-0.65cm}
    \caption{
        Rotational error (a) and differences between the true residual variance $\tilde{\sigma}^2$ and the learned variance $\sigma^2$ (b) over the training epochs. As previously, our method learns to adapt the covariances for each keypoint to minimize rotational error. Minimizing the rotational error leads to a significantly better estimate of $\sigma^2$.
    }
\label{fig:synth2}
\vspace{-0.3cm}
\end{figure}

\begin{figure}[t]
    \centering
    \begin{subfigure}[b]{0.478\textwidth}
        \centering
        {\includegraphics[trim={0cm 0.2cm 0cm 0.2cm},clip,width=\textwidth]{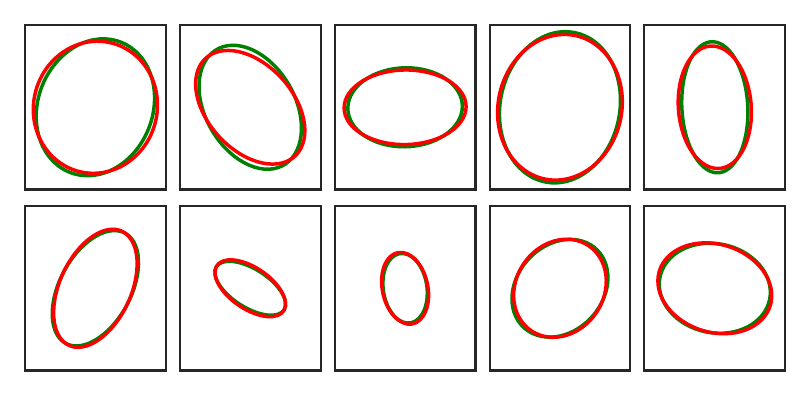}}
        \caption{covariance matrices in the 2nd frame after 100 epochs training}
        \label{fig:synth_target_covs2}
    \end{subfigure}
    \vspace{-0.7cm}
    \caption{
        Estimated (red) covariance ellipses in the second frame, learned from 128\,000 examples. Ground truth (green) covariances as comparison. Training data with enough variety gives a gradient that allows to correctly learn the covariances even in the image plane, overcoming the unobservabilities of the first experiment.
    }
\label{fig:synth_covs2}
\vspace{-0.6cm}
\end{figure}

\subsection{Simulated Experiments}
 In the simulated experiments we overfit covariance estimates for a single relative pose estimation problem using the gradient from \autoref{eqn:grad}. For this, We create a random relative pose estimation problem consisting of two camera-frames observing randomly generated points in 3D space. The points are projected into camera frames using a pinhole camera model. Each projected point is assigned a random gaussian noise distribution. From this 128\,000 random problems are sampled. We learn the noise distributions by initializing all covariance estimates as scaled identity matrices, solving the relative pose estimation problem using the \ac{pnec} and updating the parameters of the distribution using the gradient of \autoref{eqn:grad} directly. We train for 100 epochs with the ADAM \cite{Kingma2015AdamAM} optimizer with (0.9, 0.99) as parameters and a batch size of 12\,800 for a stable gradient. 
 \begin{figure}[t]
\begin{center}
    \includegraphics[trim={0.2cm 0.2cm 0.2cm 0.2cm},clip,width=0.40\textwidth]{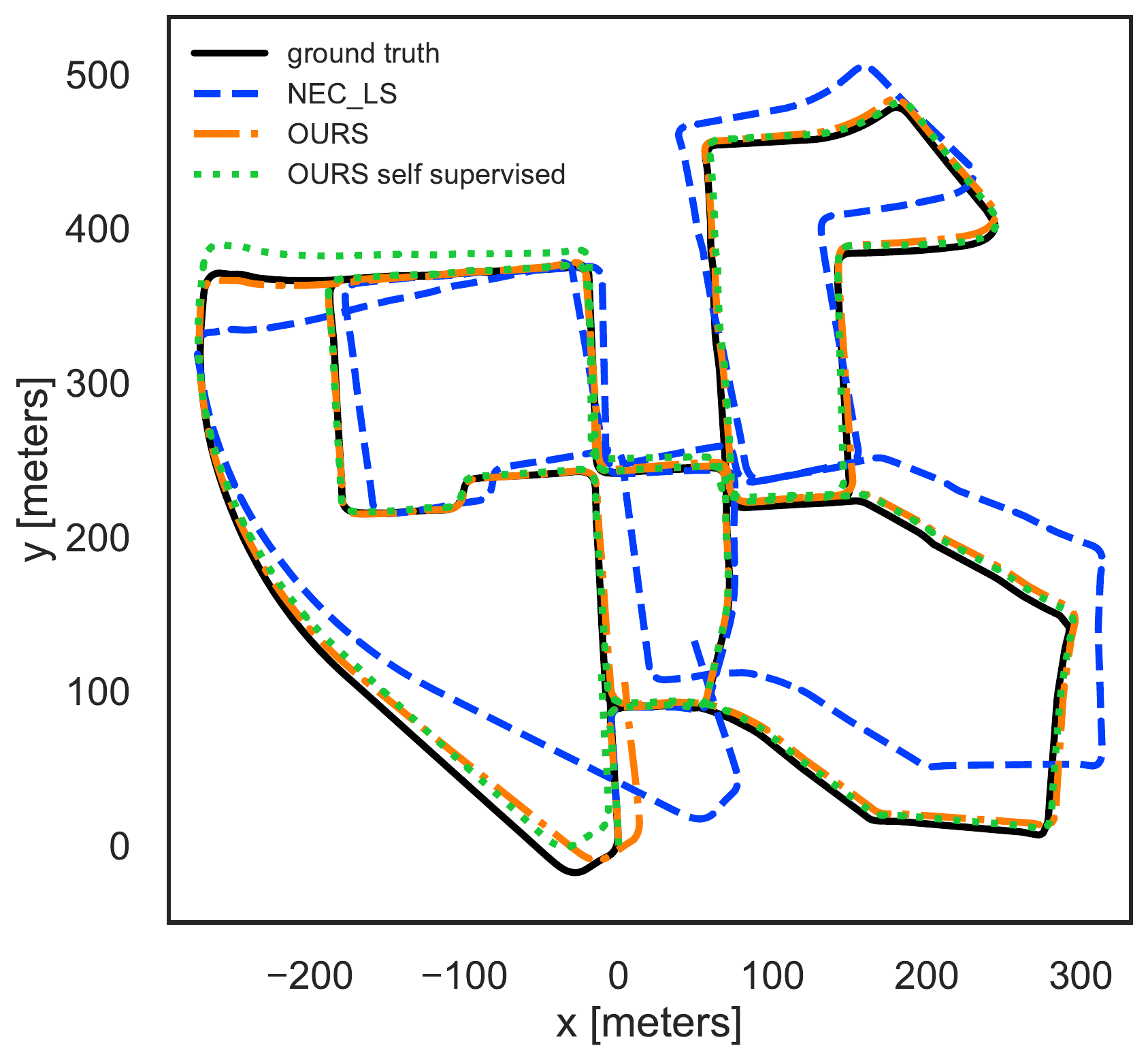}
    \vspace{-0.6cm}
\end{center}
    \caption{Qualitative trajectory comparison for KITTI seq.~00. Since we compare monocular methods, that cannot estimate the correct scale from a pair of images, we use the scale of the ground truth translations for visualization purposes. Both, our supervised and self-supervised approaches lead to significant improvements in the trajectory. There is little drift even without additional rotation averaging \cite{MRO_Chng2020} or loop closure \cite{ORB_SLAM2_Mur-Artal2017}.
         }
    \vspace{-0.75cm}
\label{fig:kitti_trajectory}
\end{figure}
 \begin{table*}[t]
\small
\centering
\sisetup{detect-weight,mode=text}
\renewrobustcmd{\bfseries}{\fontseries{b}\selectfont}
\renewrobustcmd{\boldmath}{}
\newrobustcmd{\B}{\bfseries}
\addtolength{\tabcolsep}{-3.5pt}

\begin{tabular} {p{0.6cm} r r r|r r r|r r r|r r r|r r r|r r r}
\toprule
& \multicolumn{3}{c}{\scshape Nist{\'e}r-5pt \cite{EM_Nister2003}}  
& \multicolumn{3}{c}{\scshape NEC \cite{EigenNEC_Kneip2013}} 
& \multicolumn{3}{c}{\scshape NEC-LS}  
& \multicolumn{3}{c}{\scshape WEIGHTED} 
& \multicolumn{3}{c}{\scshape OURS} 
& \multicolumn{3}{c}{\scshape OURS SELF-} 
\\
& \multicolumn{3}{c}{} 
& \multicolumn{3}{c}{} 
& \multicolumn{3}{c}{}
& \multicolumn{3}{c}{\scshape NEC-LS}  
& \multicolumn{3}{c}{\scshape SUPERVISED} 
& \multicolumn{3}{c}{\scshape SUPERVISED} 
\\
Seq.  &
  {\scshape $\text{RPE}_1$} & {\scshape $\text{RPE}_n$} & {\scshape $e_t$} &
  {\scshape $\text{RPE}_1$} & {\scshape $\text{RPE}_n$} & {\scshape $e_t$} &
  {\scshape $\text{RPE}_1$} & {\scshape $\text{RPE}_n$} & {\scshape $e_t$} &
  {\scshape $\text{RPE}_1$} & {\scshape $\text{RPE}_n$} & {\scshape $e_t$} &
  {\scshape $\text{RPE}_1$} & {\scshape $\text{RPE}_n$} & {\scshape $e_t$} &
  {\scshape $\text{RPE}_1$} & {\scshape $\text{RPE}_n$} & {\scshape $e_t$} \\
\midrule
08         &        0.195 &       17.020 &      4.24 &     0.081 &     8.284 &   3.66 &              0.056 &            7.004 &              2.50 &              0.054 &              6.059 &            2.50 &    \bfseries 0.050 &    \bfseries 4.067 &  \underline{2.46} &  \underline{0.050} &  \underline{4.118} &    \bfseries 2.46 \\
\rowcolor{Gray}
09         &        0.142 &        5.754 &      1.74 &     0.053 &     1.646 &   1.43 &              0.052 &            1.553 &              0.71 &              0.051 &              1.354 &  \bfseries 0.70 &  \underline{0.049} &  \underline{1.317} &              0.71 &    \bfseries 0.049 &    \bfseries 1.278 &  \underline{0.70} \\
10         &        0.295 &       16.678 &      6.57 &     0.167 &     9.264 &   4.43 &              0.064 &            4.787 &              1.79 &    \bfseries 0.063 &              4.389 &            1.76 &  \underline{0.063} &    \bfseries 3.513 &    \bfseries 1.64 &              0.065 &  \underline{3.821} &  \underline{1.65} \\
\midrule
\rowcolor{Gray}
train &        0.249 &       11.506 &      4.13 &     0.141 &    10.127 &   2.97 &              0.082 &            6.910 &              1.72 &              0.081 &              6.410 &            1.72 &  \underline{0.077} &    \bfseries 2.378 &    \bfseries 1.69 &    \bfseries 0.077 &  \underline{2.505} &  \underline{1.69} \\
test  &        0.200 &       14.349 &      4.07 &     0.089 &     6.917 &   3.28 &              0.056 &            5.353 &              1.96 &              0.055 &              4.676 &            1.95 &    \bfseries 0.052 &    \bfseries 3.333 &  \underline{1.91} &  \underline{0.053} &  \underline{3.408} &    \bfseries 1.91 \\

\bottomrule
\end{tabular}
\vspace{-0.25cm}
\caption{Quantitative comparison on the KITTI \cite{Geiger2012CVPR} dataset with SuperPoint \cite{detone2018superpoint} keypoints. We compare two rotation and one translation metric. The results are shown for each test sequence together with the mean results on the training and test set weighted by the sequence length. Both our training setups outperform the non-probablitic algorithms but also the weighted NEC-LS using SuperGlue confidences consistently across unseen data. The learned uncertainties are able to generalise well and improve the relative pose estimation significantly.}
\label{tab:kitti_superpoint}
\vspace{-0.35cm}
\end{table*}

 \begin{table*}[t]
\small
\centering
\sisetup{detect-weight,mode=text}
\renewrobustcmd{\bfseries}{\fontseries{b}\selectfont}
\renewrobustcmd{\boldmath}{}
\newrobustcmd{\B}{\bfseries}
\addtolength{\tabcolsep}{-3.5pt}
\begin{tabular} {p{0.6cm} r r r|r r r|r r r|r r r|r r r|r r r}
\toprule
& \multicolumn{3}{c}{\scshape Nist{\'e}r-5pt \cite{EM_Nister2003}}  & \multicolumn{3}{c}{\scshape NEC \cite{EigenNEC_Kneip2013}} & \multicolumn{3}{c}{\scshape NEC-LS} & \multicolumn{3}{c}{\scshape KLT-PNEC \cite{muhle2022pnec}} & \multicolumn{3}{c}{\scshape OURS} & \multicolumn{3}{c}{\scshape OURS SELF-} \\
& \multicolumn{3}{c}{} & \multicolumn{3}{c}{} & \multicolumn{3}{c}{} & \multicolumn{3}{c}{} & \multicolumn{3}{c}{\scshape SUPERVISED} & \multicolumn{3}{c}{\scshape SUPERVISED} \\
Seq.  &
  {\scshape $\text{RPE}_1$} & {\scshape $\text{RPE}_n$} & {\scshape $e_t$} &
  {\scshape $\text{RPE}_1$} & {\scshape $\text{RPE}_n$} & {\scshape $e_t$} &
  {\scshape $\text{RPE}_1$} & {\scshape $\text{RPE}_n$} & {\scshape $e_t$} &
  {\scshape $\text{RPE}_1$} & {\scshape $\text{RPE}_n$} & {\scshape $e_t$} &
  {\scshape $\text{RPE}_1$} & {\scshape $\text{RPE}_n$} & {\scshape $e_t$} &
  {\scshape $\text{RPE}_1$} & {\scshape $\text{RPE}_n$} & {\scshape $e_t$} \\
\midrule
08         &        0.126 &        6.929 &      3.44 &     0.088 &     3.902 &   8.91 &            0.053 &              2.908 &              2.49 &              0.054 &            2.524 &            2.42 &  \underline{0.048} &  \underline{2.373} &    \bfseries 2.36 &    \bfseries 0.047 &    \bfseries 1.706 &  \underline{2.36} \\
\rowcolor{Gray}
09         &        0.090 &        2.544 &      1.28 &     0.054 &     2.027 &   6.76 &            0.052 &              2.307 &              0.74 &              0.046 &  \bfseries 1.003 &            0.69 &  \underline{0.043} &              1.244 &    \bfseries 0.64 &    \bfseries 0.042 &  \underline{1.141} &  \underline{0.64} \\
10         &        0.188 &       11.554 &      4.43 &     0.119 &     8.302 &   8.53 &            0.066 &              4.576 &              1.78 &              0.063 &            4.480 &            1.71 &  \underline{0.058} &  \underline{3.789} &    \bfseries 1.58 &    \bfseries 0.056 &    \bfseries 3.623 &  \underline{1.60} \\
\midrule
\rowcolor{Gray}
train &        0.204 &        9.677 &      3.19 &     0.173 &     8.301 &   8.59 &            0.103 &              3.955 &              1.73 &              0.104 &            4.213 &            1.66 &    \bfseries 0.094 &  \underline{2.782} &    \bfseries 1.60 &  \underline{0.096} &    \bfseries 2.737 &  \underline{1.61} \\
test  &        0.129 &        6.722 &      3.11 &     0.085 &     4.237 &   8.34 &            0.055 &              3.060 &              1.96 &              0.054 &            2.514 &            1.90 &  \underline{0.048} &  \underline{2.359} &    \bfseries 1.82 &    \bfseries 0.048 &    \bfseries 1.910 &  \underline{1.83} \\

\bottomrule
\end{tabular}
\vspace{-0.25cm}
\caption{Quantitative comparison on the KITTI \cite{Geiger2012CVPR} dataset with \ac{klt} tracks \cite{Basalt_Usenko2020}. As in \autoref{tab:kitti_superpoint}, we show the results on the test set together with the mean on the train and test set weighted by the sequence lengths. As for SuperPoint, our methods improve all metrics consistently for unseen data. Our learned covariances are significantly better for relative pose estimation than the approximation used in \cite{muhle2022pnec}.  
}
\label{tab:kitti_klt}
\vspace{-0.5cm}
\end{table*}

\autoref{fig:synth_error} shows the decrease of the rotation error over the epochs. The learned covariances decrease the error by 
 $8\%$ and $16\%$ compared to unit covariances and the \ac{nec}, respectively. This validates the importance of good covariances for the \ac{pnec}, shown in \cite{muhle2022pnec}. \autoref{fig:synth_sigma} shows the average error for the normalized variance $\sigma^2_{norm}$, given by
 \begin{equation}
     \sigma^2_{i, norm} = \frac{N \cdot \sigma^2_{i}}{\sum_{j = 0}^N \sigma^2_{j}}
 \end{equation}
 over the training epochs, obtained at the ground truth relative pose. We compare the normalized error variance, as the scale of $\sigma^2$ is not observable from the gradient. The covariances that minimize the rotational error also approximate the residual uncertainty $\sigma^2$ very closely. However, while the residual uncertainty is approximated well, the learned 2D covariances in the image plane do not correspond to the correct covariances (see \autoref{fig:synth_covs}). This is due to two different reasons. First, due to $\sigma^2_i$ dependence on both $\boldsymbol{\Sigma}_{2\text{D},i}$ and $\boldsymbol{\Sigma}_{2\text{D},i}^\prime$, there is not a single unique solution. Secondly, the direction of the gradient is dependent on the translation between the images (see \autoref{sec:grad} for more details). In this experimental setup, the information flow to the images is limited and we can only learn the true distribution for $\sigma^2$ but not for the 2D images covariances.

To address the problems with limited information flow of the previous experiment, we propose a second experiment to negate the influence of these aforementioned factors. First, each individual problem has a randomly sampled relative pose, where the first frame stays fixed. This removes the influence of the translation on the gradient direction. The noise is still drawn from the same distributions as earlier. Second, we fix the noise in the first frame to be small, isotropic, and homogeneous in nature. Furthermore, we only learn the covariances in the second frame and provide the optimization with the ground truth noise in the first frame. \autoref{fig:synth2} and \autoref{fig:synth_covs2} show, that under these constraints, we are not only able the learn the distribution for $\sigma^2$ but also $\boldsymbol{\Sigma}^\prime_{2\text{D}}$. Together, both experiments show, that we can learn the correct distributions from noisy data by following the gradient that minimizes the rotational error.

\subsection{Real World Data}

We evaluate our method on the KITTI \cite{Geiger2012CVPR} and EuRoC \cite{Burri2016euroc} dataset. Since KITTI shows outdoor driving sequences and EuRoC shows indoor scenes captured with a drone, they exhibit different motion models as well as a variety of images. For KITTI we choose sequences 00-07 as the training set for both supervised and self-supervised training. Sequences 08-10 are used as the test set. We use a smaller UNet \cite{ronneberger2015u} architecture as our network to predict the covariances for the whole image. We chose this network since it gives us a good balance between batch size, training time and performance. The network predicts the parameters for the covariances directly. We choose
 \begin{equation}
     \boldsymbol{\Sigma}_{2\text{D}}(s, \alpha, \beta) = s  \boldsymbol{R}_{\alpha} \begin{pmatrix} \beta & 0 \\ 0 & 1 - \beta \end{pmatrix} \boldsymbol{R}_{\alpha}^\top
 \end{equation}
 as a parameterization \cite{pro_cov_Brooks2001}. To ensure that our network predicts valid covariances the network output is filtered with
 \begin{align}
     f_1(x) & = \left(1 + |x|\right)^{\text{sign(x)}} \\
     f_2(x) & = x\\
     f_3(x) & = \frac{1}{1 + e^{-x}}
 \end{align}
 for $s, \alpha, \beta$, respectively. Feature points that have subpixel accuracy use the nearest pixel covariance. See \autoref{sec:training} for more details on the training setup.

\begin{table*}[t]
\small
\centering
\sisetup{detect-weight,mode=text}
\renewrobustcmd{\bfseries}{\fontseries{b}\selectfont}
\renewrobustcmd{\boldmath}{}
\newrobustcmd{\B}{\bfseries}
\addtolength{\tabcolsep}{-4.0pt}
\begin{tabular} {p{0.8cm} r r r|r r r|r r r|r r r|r r r|r r r}
\toprule
& \multicolumn{3}{c}{\scshape Nist{\'e}r-5pt \cite{Nistr2004VisualO}}  
& \multicolumn{3}{c}{\scshape NEC \cite{EigenNEC_Kneip2013}} 
& \multicolumn{3}{c}{\scshape NEC-LS}  
& \multicolumn{3}{c}{\scshape WEIGHTED} 
& \multicolumn{3}{c}{\scshape OURS SELF-} 
& \multicolumn{3}{c}{\scshape OURS \autoref{tab:kitti_superpoint}} 
\\
& \multicolumn{3}{c}{} 
& \multicolumn{3}{c}{} 
& \multicolumn{3}{c}{}
& \multicolumn{3}{c}{\scshape NEC-LS}  
& \multicolumn{3}{c}{\scshape SUPERVISED} 
& \multicolumn{3}{c}{\scshape SUPERVISED} 
\\
Seq.  &
  {\scshape $\text{RPE}_1$} & {\scshape $\text{RPE}_n$} & {\scshape $e_t$} &
  {\scshape $\text{RPE}_1$} & {\scshape $\text{RPE}_n$} & {\scshape $e_t$} &
  {\scshape $\text{RPE}_1$} & {\scshape $\text{RPE}_n$} & {\scshape $e_t$} &
  {\scshape $\text{RPE}_1$} & {\scshape $\text{RPE}_n$} & {\scshape $e_t$} &
  {\scshape $\text{RPE}_1$} & {\scshape $\text{RPE}_n$} & {\scshape $e_t$} &
  {\scshape $\text{RPE}_1$} & {\scshape $\text{RPE}_n$} & {\scshape $e_t$} \\
\midrule
V1\_01      &        0.501 &              71.87 &    \bfseries 31.86 &    \bfseries 0.320 &              39.50 &  43.12 &              0.387 &        52.92 &     46.31 &              0.388 &              56.52 &           46.82 &  \underline{0.327} &  \bfseries 31.12 &              35.56 &              0.332 &  \underline{31.81} &  \underline{34.01} \\
\rowcolor{Gray}
V1\_02    &        0.541 &              32.01 &    \bfseries 20.36 &    \bfseries 0.389 &    \bfseries 28.11 &  26.95 &              0.540 &        70.08 &     28.94 &              0.542 &              68.35 &           29.81 &              0.444 &            30.39 &              21.98 &  \underline{0.436} &  \underline{29.07} &  \underline{21.29} \\
V1\_03 &        0.660 &  \underline{27.39} &              25.00 &    \bfseries 0.492 &    \bfseries 25.42 &  31.06 &              0.552 &        76.72 &     31.58 &              0.555 &              78.14 &           32.25 &  \underline{0.510} &            29.52 &  \underline{24.19} &              0.520 &              31.18 &    \bfseries 24.13 \\
\rowcolor{Gray}
V2\_01     &        0.515 &              61.45 &              33.51 &              0.316 &              31.95 &  39.79 &              0.310 &        35.84 &     39.00 &              0.314 &              38.62 &           39.62 &    \bfseries 0.285 &  \bfseries 17.61 &  \underline{32.40} &  \underline{0.295} &  \underline{22.41} &    \bfseries 30.58 \\
V2\_02    &        0.545 &              43.73 &              22.24 &              0.396 &              25.48 &  32.21 &  \underline{0.369} &        26.96 &     25.36 &    \bfseries 0.365 &  \underline{25.09} &           25.81 &              0.382 &            25.32 &  \underline{21.16} &              0.386 &    \bfseries 21.91 &    \bfseries 20.34 \\
\rowcolor{Gray}
V2\_03 &        1.123 &    \bfseries 36.71 &    \bfseries 28.77 &              0.976 &  \underline{48.26} &  37.60 &    \bfseries 0.939 &       107.11 &     36.74 &  \underline{0.941} &             100.73 &           36.71 &              0.942 &            52.72 &              31.13 &              0.991 &              55.41 &  \underline{30.40} \\
\midrule
mean      &        0.631 &              48.45 &  \underline{27.56} &  \underline{0.463} &              33.51 &  36.03 &              0.494 &        58.90 &     35.61 &              0.496 &              58.95 &           36.11 &    \bfseries 0.461 &  \bfseries 30.57 &              28.46 &              0.472 &  \underline{31.44} &    \bfseries 27.44 \\

\bottomrule
%
\end{tabular}
\vspace{-0.25cm}
\caption{Quantitative comparison on the Vicon sequences of the EuRoC dataset \cite{Burri2016euroc} with SuperPoint \cite{detone2018superpoint} keypoints. The dataset is more difficult than KITTI (see \autoref{tab:kitti_klt} and \autoref{tab:kitti_superpoint}) with SuperPoint and SuperGlue \cite{sarlin2020superglue} finding far fewer matches. As reported in \cite{muhle2022pnec} the least squares implementations struggle with bad initialization under these adverse conditions with NEC-LS performing especially poor. From all least squares optimizations, our learned covariances consistently perform the best, even outperforming the \ac{nec} most of the time.}
\label{tab:euroc}
\vspace{-0.3cm}
\end{table*}

\pseudoparagraph{Supervised Learning.}
 To show that our method generalizes to different keypoint detectors, we train two networks, one for SuperPoint \cite{detone2018superpoint} and one for \ac{klt} tracks obtained from \cite{Basalt_Usenko2020}. The SuperPoint keypoints are matched using SuperGlue \cite{sarlin2020superglue}. For training we use a batch size of 8 images pairs for SuperPoint and 16 images pairs for \ac{klt} tracks. We trained for 100 epochs for both SuperPoint and \ac{klt} tracks. More training details are provided in the supplementary material. To ensure our network does not overfit on specific keypoint locations, we randomly crop the images before finding correspondences during training time. During evaluation we use the uncropped images to obtain features. During training we randomly perturb the ground truth pose as a starting point. To increase robustness, we first use the eigenvalue based optimization of the \ac{nec} in a RANSAC scheme \cite{Kneip2014opengv} to filter outliers. This is followed by a custom least squares implementation of the \ac{nec} (NEC-LS), followed by optimizing \autoref{eqn:symm-pnec-cost}. As reported in \cite{muhle2022pnec} we found, that such a mutli-stage optimization provides the most robust and accurate results. We show examples of how the \ac{dnls}-learned covariances change the energy function landscape in the supplementary material.

\pseudoparagraph{Self-Supervised Learning.}
 We evaluate our self-supervised training setup on the same data as our supervised training. Due to needing image tuples instead of pairs, we reduce the batch size to 12 for \ac{klt} image triplets. This gives us 24 and 36 images pairs per batch, respectively. The training epochs are reduced to 50. More training details for the supervised and self-supervised training can be found in the supplementary material.

\pseudoparagraph{Results.} We evaluate the learned covariances in a VO setting. We compare the proposed \ac{dnls} approach to the popular 5pt algorithm \cite{Nistr2004VisualO} and the \ac{nec} \cite{EigenNEC_Kneip2013} as implemented in \cite{Kneip2014opengv}. To investigate the benefit of our learned covariances we include the NEC-LS implementation as well as the symmetric \ac{pnec} with the covariances from \cite{muhle2022pnec} in \autoref{tab:kitti_klt}. For \autoref{tab:kitti_superpoint} we additionally include a weighted version of our custom NEC-LS implementation with matching confidence from SuperGlue as weights. 
 All methods are given the same feature matches and use a constant motion model for initializing the optimizations.
 We evaluate on the rotational versions of the $\text{RPE}_1$ and $\text{RPE}_n$ and the cosine error $e_t$ for the translation as defined in \cite{muhle2022pnec, MRO_Chng2020}. 
 \autoref{tab:kitti_superpoint} and \autoref{tab:kitti_klt} show the average results on the test set over 5 runs for SuperPoint and \ac{klt} tracks on KITTI \cite{Geiger2012CVPR}, respectively. We show additional results in \autoref{sec:results}. Our methods consistently perform the best over all sequences, with the self-supervised being on par with our supervised training. Compared to its non-probabilistic counterpart NEC-LS, our method improves the $\text{RPE}_1$ by $7\%$ and $13\%$ and the $\text{RPE}_n$ by $37\%$ and $23\%$ for different keypoint detectors on unseen data. It also improves upon weighted methods, like weighted NEC-LS and the non-learned covariances for the \ac{pnec} \cite{muhle2022pnec}. This demonstrates the importance of correctly modeling the feature correspondence quality. We show an example trajectory in \autoref{fig:kitti_trajectory}. 

\autoref{tab:euroc} shows the results on the EuRoC dataset for SuperPoint. Pose estimation is significantly more difficult compared to KITTI, often having few correspondences between images. However, our method generalizes to different datasets, with the network trained on KITTI and our self-supervised approach, outperforming the others most of the time. Especially a direct comparison with NEC-LS, the closest non-probabilistic method, shows significant improvements of $7\%$ for $\text{RPE}_1$ and $48\%$ for the $\text{RPE}_n$. 

\section{Discussion and Limitations}
 Our experiments demonstrate the capability of our framework to to correctly learn positional uncertainty, leading to improved results for relative pose estimation for VO. Our approach generalizes to different feature extractors and to different datasets, providing a unified approach to estimate the noise distribution of keypoint detectors.
 However, our method requires more computational resources than the original uncertainty estimation for the PNEC.

We evaluate our learned covariances in a visual odometry setting, showing that they lead to reduced errors and especially less drift in the trajectory. However, this does not guarantee that the covariances are \emph{calibrated}. Our framework inherits the ambiguity of the \ac{pnec} with regard to the noise scale. The true scale of the noise is not observable from relative pose estimation alone and only the relative scale between covariances can be learned. For the purposes of VO, this scale ambiguity is negligible. 

As our synthetic experiments show, diverse data is needed to correctly identify the $2\text{D}$ noise distribution. However, obtaining the noise distribution is difficult for keypoint detectors -- hence learning it from pose regression.
 Further limitations are addressed in \autoref{sec:limitations}.

 \section{Conclusion}
 We present a novel \ac{dnls} framework for estimating positional uncertainty. Our framework can be combined with any feature extraction algorithm, making it extremely versatile. Regressing the noise distribution from relative pose estimation, ensures that learned covariance matrices are suitable for visual odometry tasks. In synthetic experiments, our framework is capable to learn the correct noise distribution from noisy data. We showed the practical application of our framework on real-world data for different feature extractors. Our learned uncertainty consistently outperforms a variety of non-probabilistic relative pose estimation algorithms as well as other uncertainty estimation methods.

\footnotesize{
 \paragraph{Acknowledgements.}
 This work was supported by the ERC Advanced Grant SIMULACRON, by the Munich Center for Machine Learning and by the EPSRC Programme Grant VisualAI EP/T028572/1.
 }

\newpage
 \appendix
 \twocolumn[\centering \section*{Learning Correspondence Uncertainty
 via Differentiable Nonlinear Least Squares\\[1.3ex]Supplementary Material}]
 \section{Overview}
 This supplementary material presents additional insight into learning positional uncertainty using \ac{dnls}. We start by addressing limitations of our framework in \autoref{sec:limitations}. \autoref{sec:approx} gives a derivation of the residual variance $\sigma^2_s$ for the symmetric \ac{pnec}. We investigate the unobservabilities of the gradient in \autoref{sec:grad}. The training and evaluation details are given in \autoref{sec:training}. We show further quantitative evaluations in \autoref{sec:minimum} and \autoref{sec:results}. This includes examples of how the learned covariances move the minimum around the ground truth and the results on the sequences 00-07 of the KITTI \cite{Geiger2012CVPR} dataset. We compare our learned covariances against error estimates from reprojection using ground truth poses.

\section{Limitations} \label{sec:limitations}
 In this section, we will address limitations of our method, not mentioned in the main paper due to constrained space. We learn to estimate the noise distribution of keypoint detectors, using regression on the pose error. The gradient we use for learning the distribution is restricted to points in the image that are detected as keypoints. This restrict our method to learn only on regions of the image with a high chance of producing keypoints. While we don't need uncertainty information for regions without keypoints, this sparse information flow might reduce generalization capabilities to different datasets. Sparsity if further enhanced by using RANSAC to filter outliers, removing points that are too far off. However, we choose to include RANSAC for our training to obtain better pose estimates for gradients not dominated by outliers. We tried to mitigate the effect of overfitting on keypoint positions by cropping the images, leading to different keypoint positions. Furthermore, our experiments showed that generalization between KITTI and EuRoC are possible. 

\begin{figure}[t]
    \centering
    {\includegraphics[trim={0.25cm 0.3cm 0.25cm 0.2cm},clip,width=0.4\textwidth]{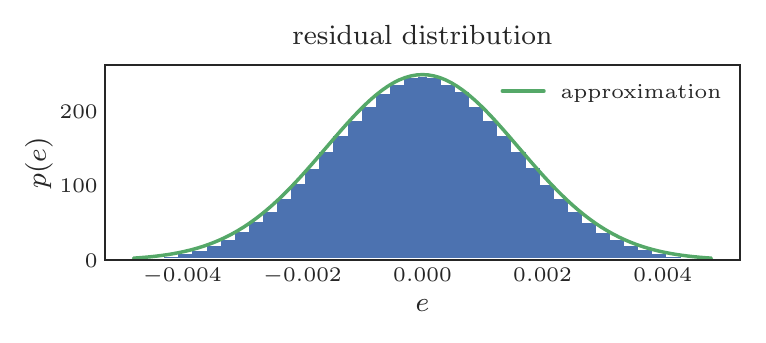}}
    \vspace{-0.35cm}
    \caption{
        Approximation of the residual variances. The analytical approximation given in the main paper accurately models the true distribution of the residual given a similar setup to the KITTI dataset.
    }
\label{fig:sigma}
\vspace{-0.45cm}
\end{figure}
 \begin{figure}[t]
    \centering
    {\includegraphics[trim={0.25cm 0.25cm 0.25cm 0.2cm},clip,width=0.4\textwidth]{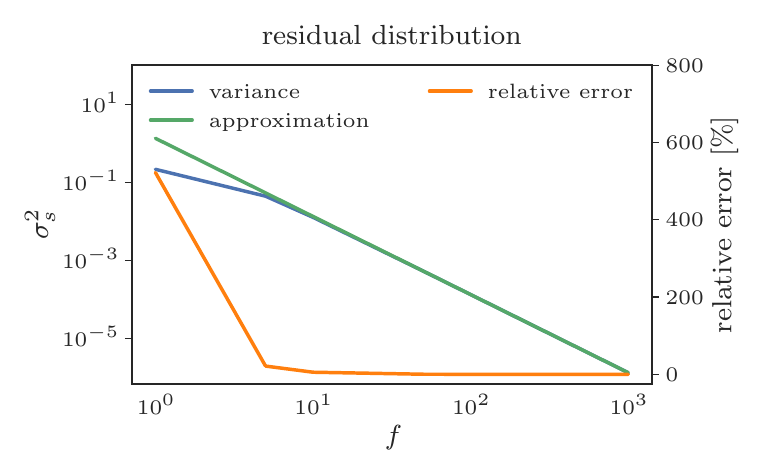}}
    \vspace{-0.35cm}
    \caption{
        Approximation of the residual variances over different focal lengths. The scale of the variance is correlated with the focal length. Our approximation is better, the smaller the variance is. For a focal length similar to the one found in the KITTI dataset, the relative error is $0.015\%$. 
    }
\label{fig:sigma_focal}
\vspace{-0.55cm}
\end{figure}

\begin{figure*}[t]
    \centering
    \begin{subfigure}[b]{0.82\textwidth}
        \centering
        {\includegraphics[trim={1.0cm 0cm 1.0cm 0cm},clip,width=\textwidth]{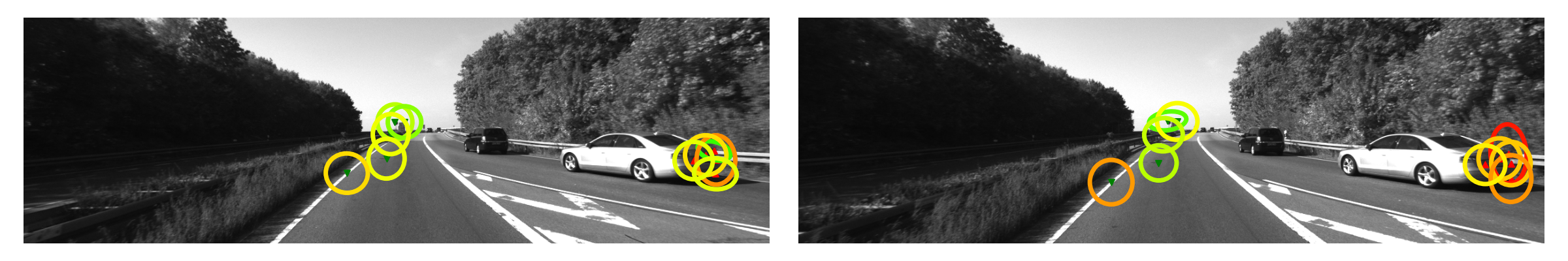}}
        \caption{selected keypoints with covariances}
        \label{fig:bad1kp}
    \end{subfigure}
    \vspace{0.0cm}
    \begin{subfigure}[b]{0.16\textwidth}
        \centering
        {\includegraphics[trim={0.25cm 0.5cm 0.25cm 0.5cm},clip,width=\textwidth]{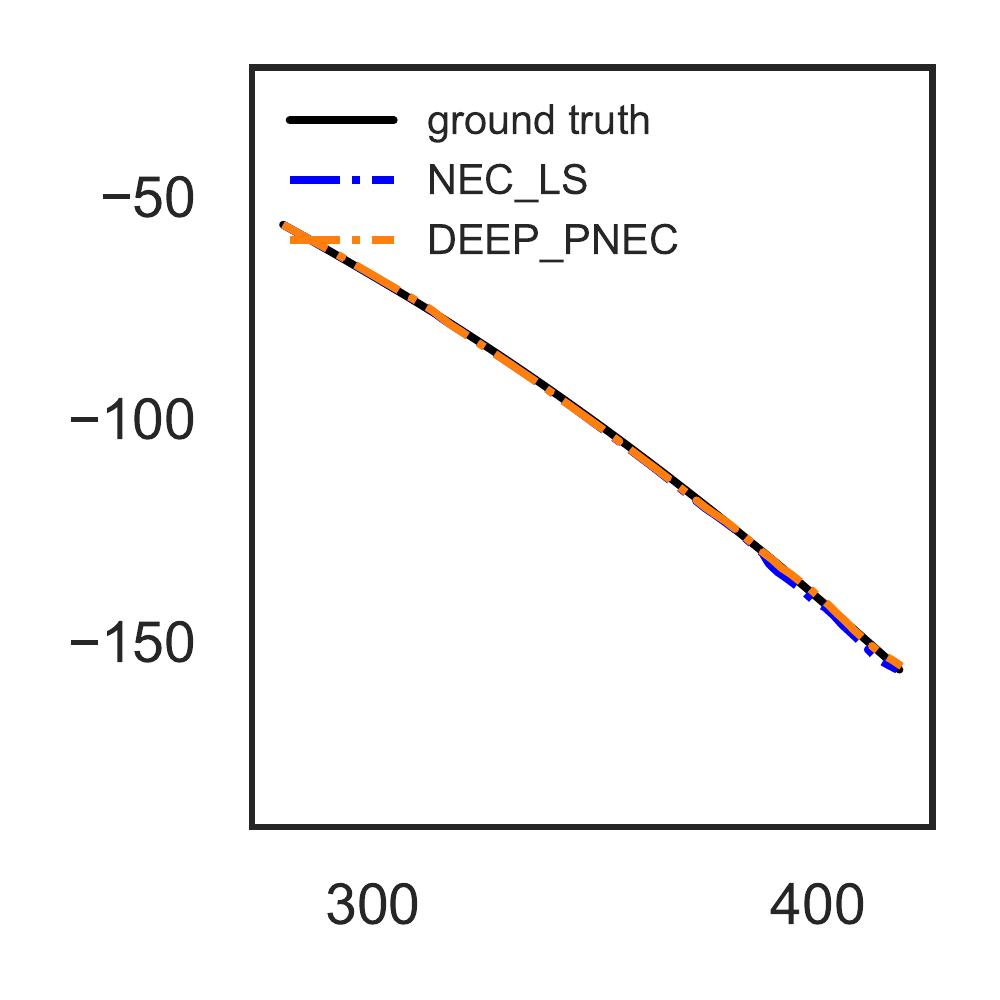}}
        \caption{trajectory}
        \label{fig:bad1traj}
    \end{subfigure}
    \vspace{-0.1cm}
    \begin{subfigure}[b]{0.82\textwidth}
        \centering
        {\includegraphics[trim={1.0cm 0cm 1.0cm 0cm},clip,width=\textwidth]{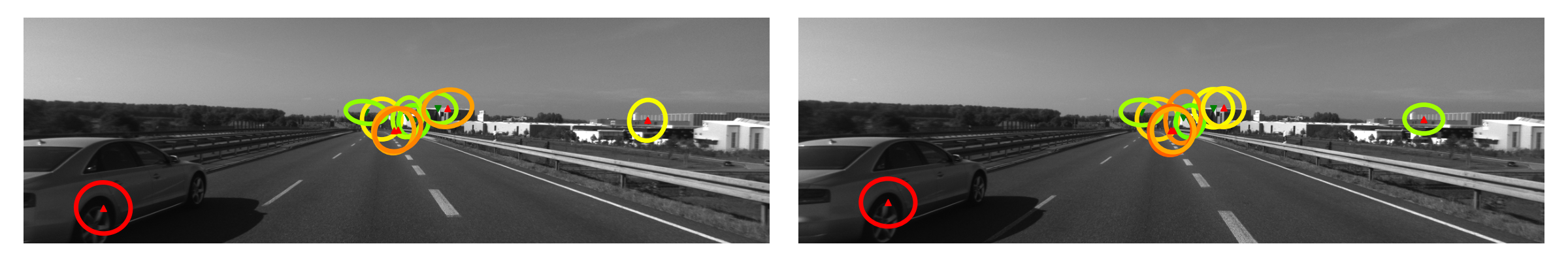}}
        \caption{selected keypoints with covariances}
        \label{fig:bad2kp}
    \end{subfigure}
    \vspace{0.0cm}
    \begin{subfigure}[b]{0.16\textwidth}
        \centering
        {\includegraphics[trim={0.25cm 0.5cm 0.25cm 0.5cm},clip,width=\textwidth]{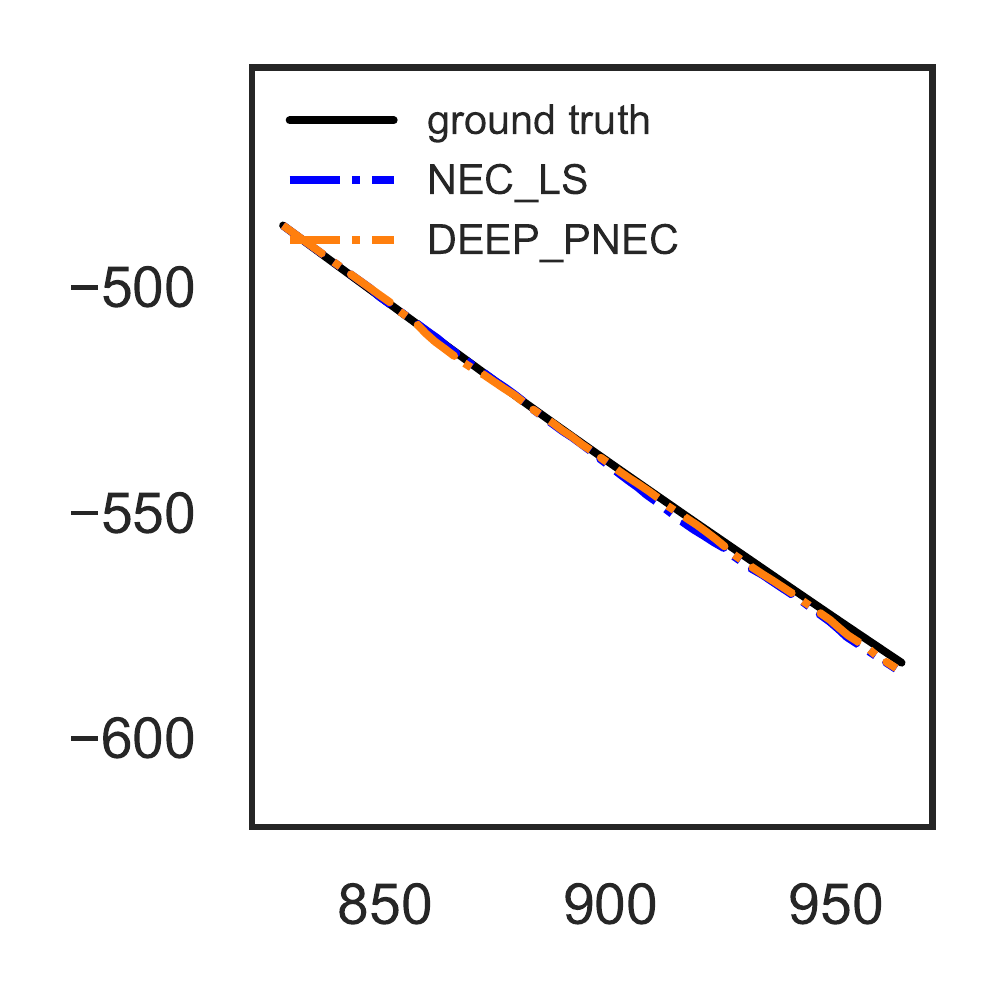}}
        \caption{trajectory}
        \label{fig:bad2traj}
    \end{subfigure}
    \vspace{-0.1cm}
    \begin{subfigure}[b]{0.82\textwidth}
        \centering
        {\includegraphics[trim={1.0cm 0cm 1.0cm 0cm},clip,width=\textwidth]{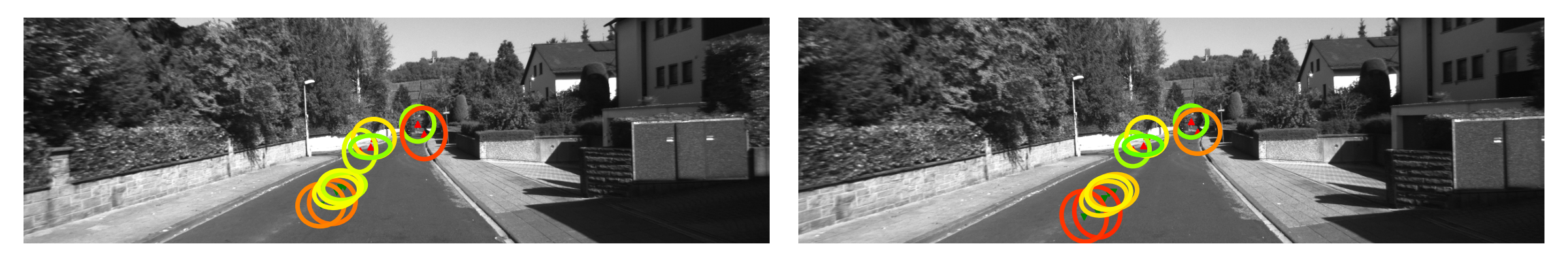}}
        \caption{selected keypoints with covariances}
        \label{fig:bad3kp}
    \end{subfigure}
    \vspace{0.0cm}
    \begin{subfigure}[b]{0.16\textwidth}
        \centering
        {\includegraphics[trim={0.25cm 0.5cm 0.25cm 0.5cm},clip,width=\textwidth]{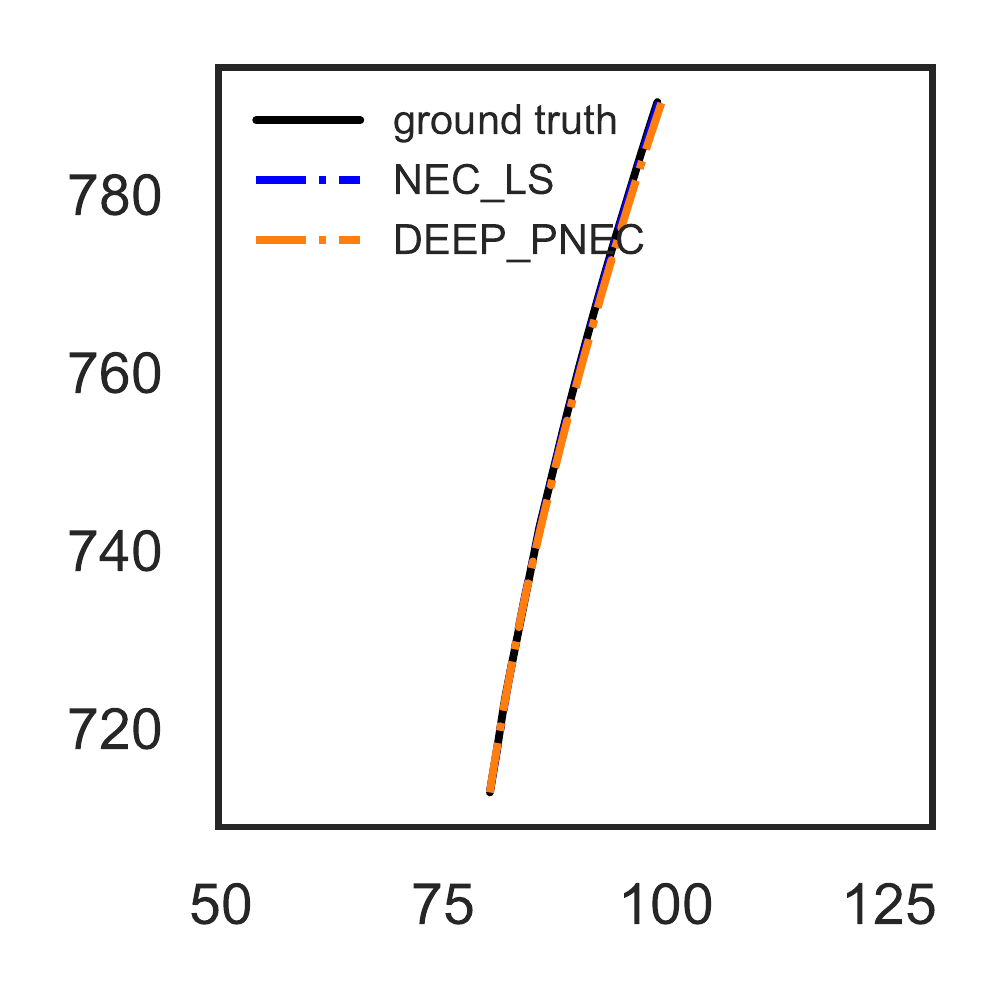}}
        \caption{trajectory}
        \label{fig:bad3traj}
    \end{subfigure}
    \vspace{-0.1cm}
    \caption{
        Left: estimated keypoints with covariances (color-coded ellipses) for examples where our method performs worse than NEC-LS. Good ($\color{green}\blacktriangledown$) and bad correspondences ($\color{red}\blacktriangle$) based on the reprojection error. Right: corresponding sections of the trajectory. (a) and (c) show examples with keypoints on dynamic objects. Although their estimated covariances is somewhat lower (especially in (c)) this is not enough to compensate the error. (e) shows an example where points with a higher reprojection error get assigned a covariances on a similar level or slightly better than good correspondences.
    }
\label{fig:bad}
\end{figure*}

\begin{figure*}[t]
    \centering
    \begin{subfigure}[b]{0.82\textwidth}
        \centering
        {\includegraphics[trim={1.0cm 0cm 1.0cm 0cm},clip,width=\textwidth]{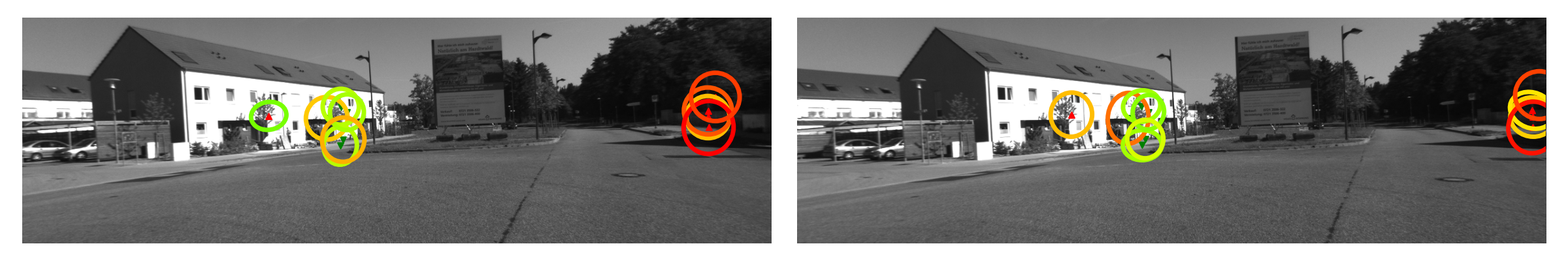}}
        \caption{selected keypoints with covariances}
        \label{fig:good1kp}
    \end{subfigure}
    \vspace{0.0cm}
    \begin{subfigure}[b]{0.16\textwidth}
        \centering
        {\includegraphics[trim={0.25cm 0.5cm 0.25cm 0.5cm},clip,width=\textwidth]{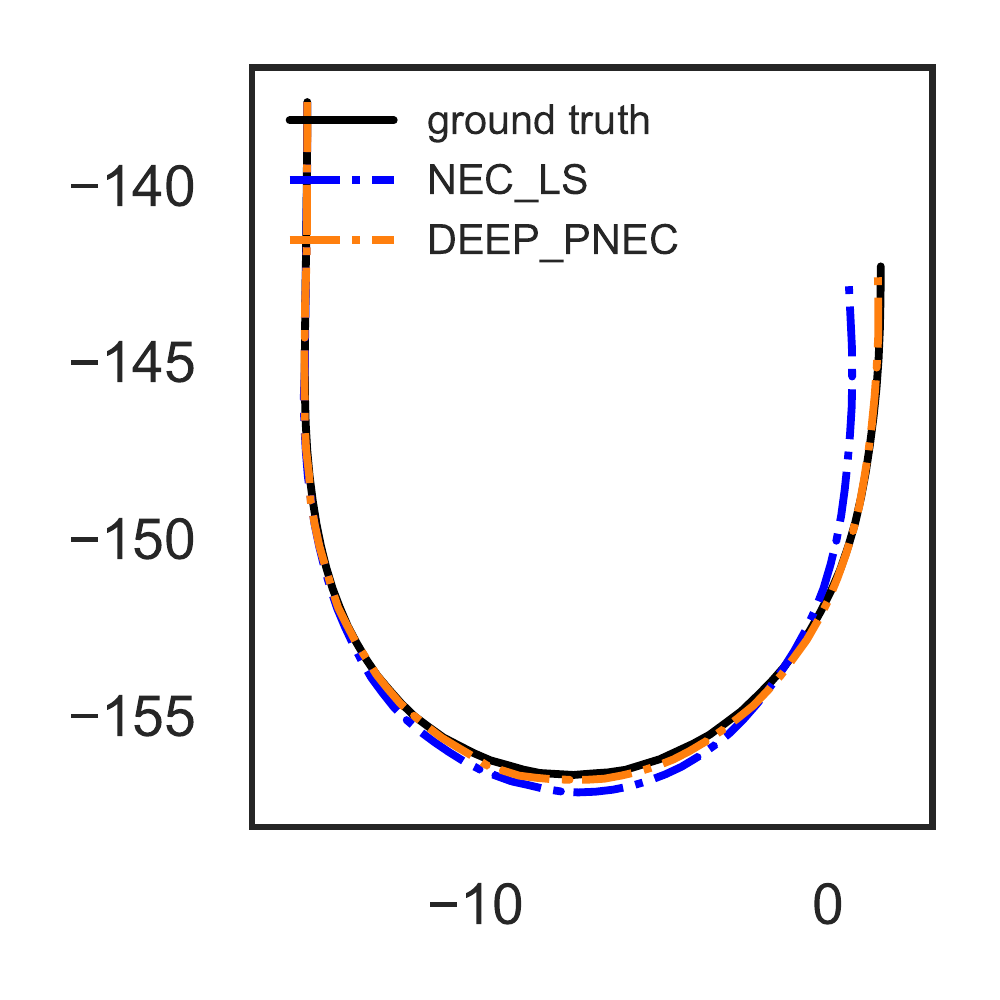}}
        \caption{trajectory}
        \label{fig:good1traj}
    \end{subfigure}
    \vspace{-0.1cm}
    \begin{subfigure}[b]{0.82\textwidth}
        \centering
        {\includegraphics[trim={1.0cm 0cm 1.0cm 0cm},clip,width=\textwidth]{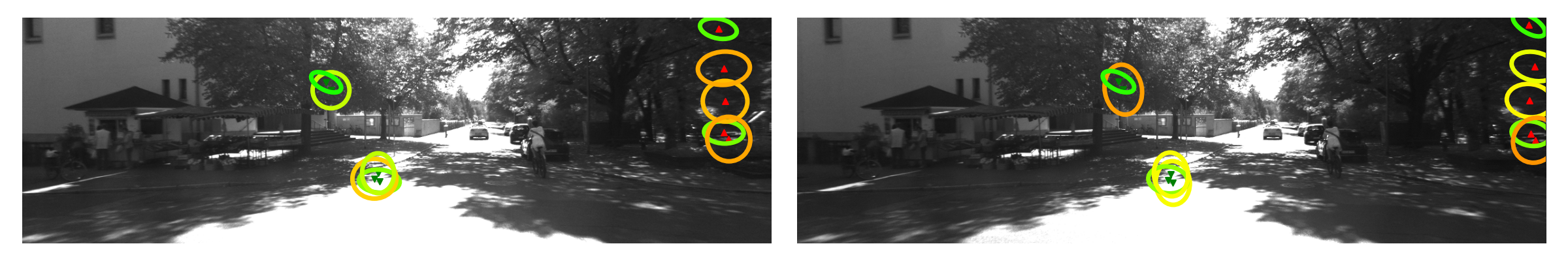}}
        \caption{selected keypoints with covariances}
        \label{fig:good2kp}
    \end{subfigure}
    \vspace{0.0cm}
    \begin{subfigure}[b]{0.16\textwidth}
        \centering
        {\includegraphics[trim={0.25cm 0.5cm 0.25cm 0.5cm},clip,width=\textwidth]{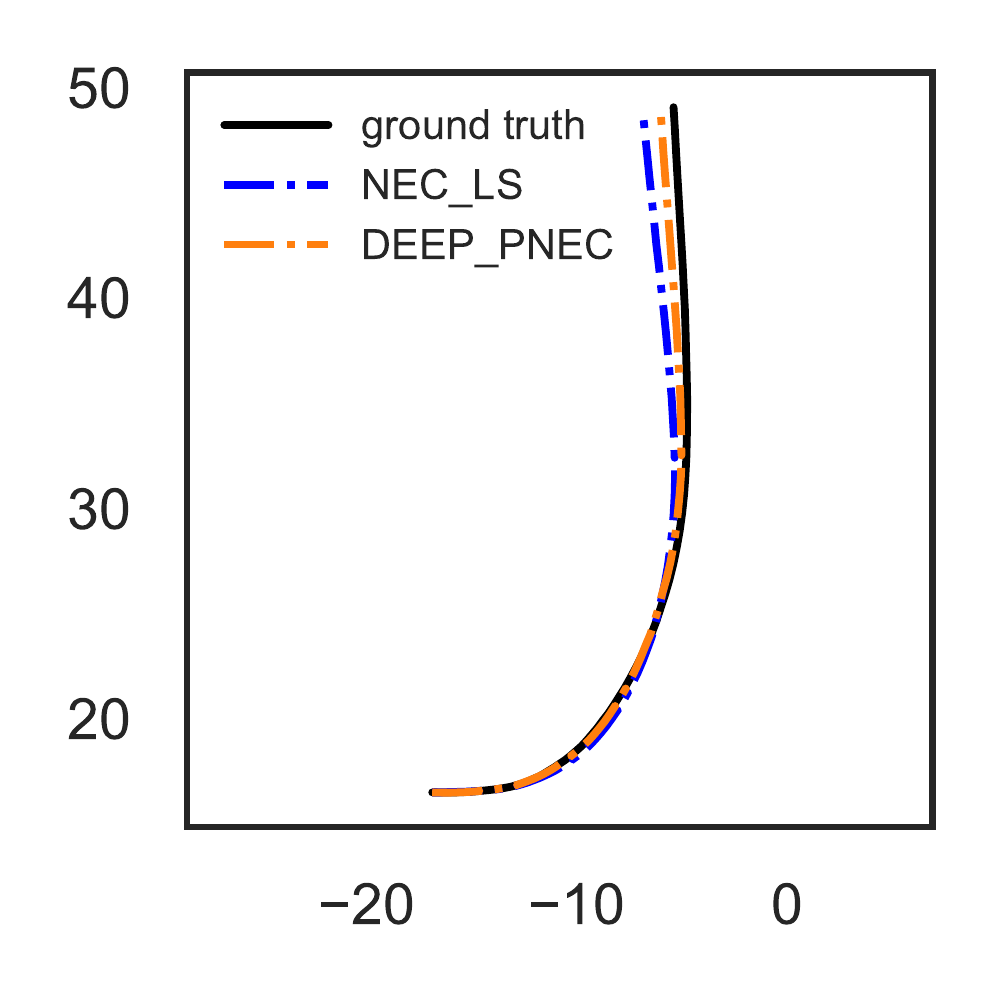}}
        \caption{trajectory}
        \label{fig:good2traj}
    \end{subfigure}
    \vspace{-0.1cm}
    \begin{subfigure}[b]{0.82\textwidth}
        \centering
        {\includegraphics[trim={1.0cm 0cm 1.0cm 0cm},clip,width=\textwidth]{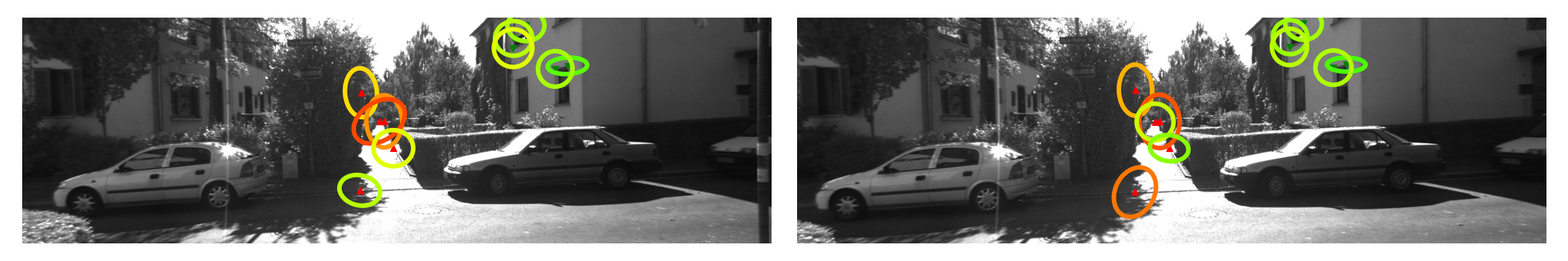}}
        \caption{selected keypoints with covariances}
        \label{fig:good3kp}
    \end{subfigure}
    \vspace{0.0cm}
    \begin{subfigure}[b]{0.16\textwidth}
        \centering
        {\includegraphics[trim={0.25cm 0.5cm 0.25cm 0.5cm},clip,width=\textwidth]{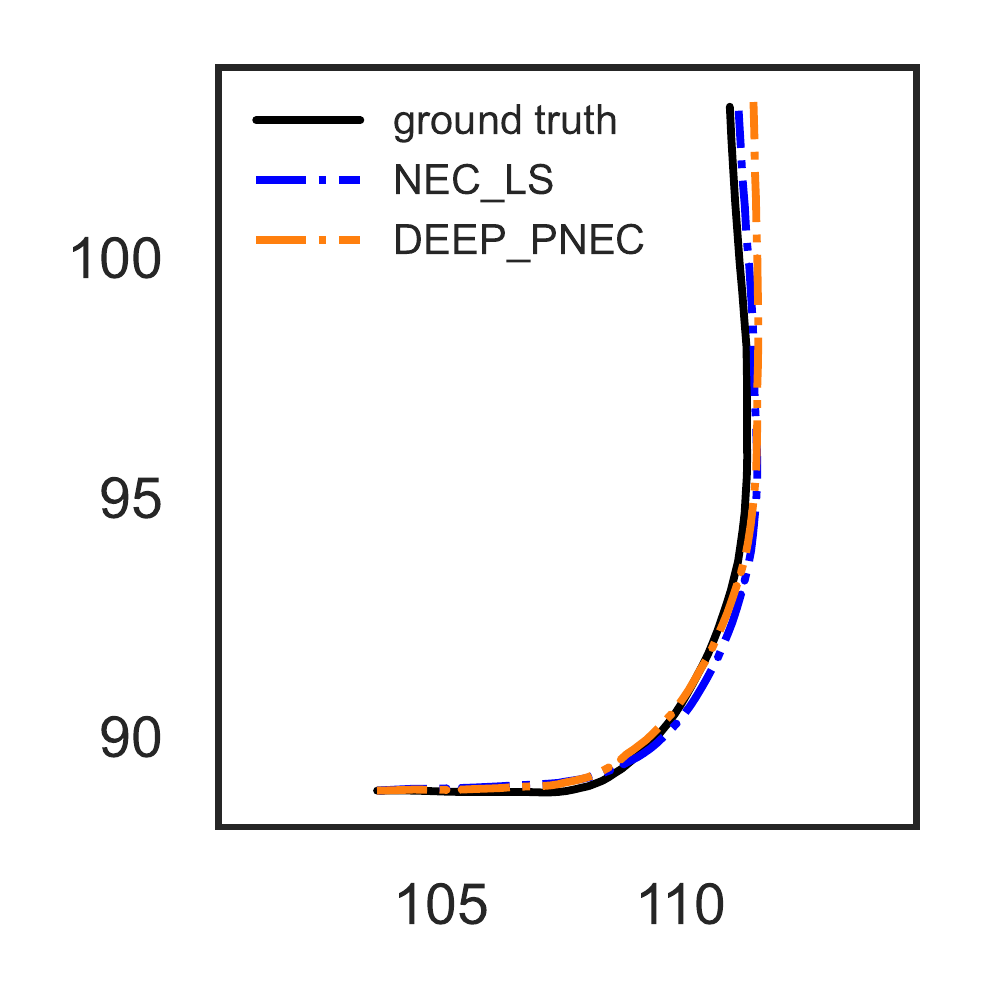}}
        \caption{trajectory}
        \label{fig:good3traj}
    \end{subfigure}
    \vspace{-0.1cm}
    \caption{
        Left: estimated keypoints with covariances (color-coded ellipses) for examples where our method performs better than NEC-LS. Good ($\color{green}\blacktriangledown$) and bad correspondences ($\color{red}\blacktriangle$) based on the reprojection error. Right: corresponding sections of the trajectory. Covariances for bad correspondences are estimated to be higher in these examples. They are down-weighted in the optimization leading to better pose estimates.
    }
\label{fig:good}
\vspace{-0.45cm}
\end{figure*}

\autoref{fig:bad} and \autoref{fig:good} show examples where our method performs worse and better than the NEC-LS optimization based on the estimated covariances. We investigate the keypoints with the highest and lowest reprojection error. As \autoref{fig:bad} shows, our method is not always able to compensate keypoints on dynamic objects leading to a large rotational error. The trajectories in \autoref{fig:good} show the improvements our method is able to achieve compared to NEC-LS.


\section{Approximating $\sigma^2_s$} \label{sec:approx}
 This section show derives the residual variance from the bearing vector covariances in both images. Given both bearing vectors $\boldsymbol{f}$ and $\boldsymbol{f}\prime$ are noisy, we can write them as
 \begin{align}
     \boldsymbol{f} & = \boldsymbol{\mu} + \boldsymbol{\eta}, \quad \boldsymbol{\eta} \sim \mathcal{N}(\boldsymbol{0}, \boldsymbol{\Sigma})\,, \\
     \boldsymbol{f} & = \boldsymbol{\mu}^\prime + \boldsymbol{\eta}^\prime, \quad \boldsymbol{\eta}^\prime \sim \mathcal{N}(\boldsymbol{0}, \boldsymbol{\Sigma}^\prime)\,,
 \end{align}
 with a constant and a noise term. We then get the new normal vector as
 \begin{align}
     \boldsymbol{n}_s & = (\boldsymbol{\mu} + \boldsymbol{\eta}) \times \boldsymbol{R} (\boldsymbol{\mu}^\prime + \boldsymbol{\eta}^\prime) \\
     & = \hat{\boldsymbol{\mu}} \boldsymbol{R} \boldsymbol{\mu}^\prime + \hat{\boldsymbol{\mu}} \boldsymbol{R} \boldsymbol{\eta}^\prime + \hat{\boldsymbol{\eta}} \boldsymbol{R} \boldsymbol{\mu}^\prime + \hat{\boldsymbol{\eta}} \boldsymbol{R} \boldsymbol{\eta}^\prime \,, \nonumber
 \end{align}
 with a constant term $\boldsymbol{\mu}_n =  \hat{\boldsymbol{\mu}} \boldsymbol{R} \boldsymbol{\mu}^\prime$ and a noise term $\boldsymbol{\eta}_n = \boldsymbol{R} \boldsymbol{\eta}^\prime + \hat{\boldsymbol{\eta}} \boldsymbol{R} \boldsymbol{\mu}^\prime + \hat{\boldsymbol{\eta}} \boldsymbol{R} \boldsymbol{\eta}^\prime$. The noise term is zero centered and has a variance of
 \begin{equation}
     \boldsymbol{\Sigma}_n = \hat{(\boldsymbol{R} \boldsymbol{\mu}^\prime_i)} \boldsymbol{\Sigma}_i \hat{(\boldsymbol{R} \boldsymbol{\mu}^\prime_i)}{}^\top + \hat{\boldsymbol{\mu}_i} \boldsymbol{R} \boldsymbol{\Sigma}^\prime_i \boldsymbol{R}^\top \hat{\boldsymbol{\mu}_i}{}^\top + \tilde{\boldsymbol{\Sigma}}\,,
 \end{equation}
 where $\tilde{\boldsymbol{\Sigma}}$ is constructed from the columns of $\boldsymbol{\Sigma}$ and $\boldsymbol{\Sigma}^\prime_R = \boldsymbol{R} \boldsymbol{\Sigma}^\prime \boldsymbol{R}^\top$ as
 \begin{equation}
     \tilde{\boldsymbol{\Sigma}} = \begin{bmatrix} (\boldsymbol{\Sigma}_2 \times \boldsymbol{\Sigma}^\prime_{R,3} + \boldsymbol{\Sigma}_3 \times \boldsymbol{\Sigma}^\prime_{R,2})^\top \\
     (\boldsymbol{\Sigma}_3 \times \boldsymbol{\Sigma}^\prime_{R,1} + \boldsymbol{\Sigma}_1 \times \boldsymbol{\Sigma}^\prime_{R,3})^\top \\
     (\boldsymbol{\Sigma}_1 \times \boldsymbol{\Sigma}^\prime_{R,2} + \boldsymbol{\Sigma}_2 \times \boldsymbol{\Sigma}^\prime_{R,1})^\top
     \end{bmatrix} \,.
 \end{equation}
 As stated in the main paper, we use an approximation of the noise distribution. Since $\tilde{\boldsymbol{\Sigma}}$ is order of magnitudes smaller than the other terms, we can approximate $\boldsymbol{\Sigma}_n$ as
 \begin{equation}
     \boldsymbol{\Sigma}_n \approx \hat{(\boldsymbol{R} \boldsymbol{\mu}^\prime_i)} \boldsymbol{\Sigma}_i \hat{(\boldsymbol{R} \boldsymbol{\mu}^\prime_i)}{}^\top + \hat{\boldsymbol{\mu}_i} \boldsymbol{R} \boldsymbol{\Sigma}^\prime_i \boldsymbol{R}^\top \hat{\boldsymbol{\mu}_i}{}^\top \,.
 \end{equation}

The final residual variance is given by
 \begin{equation}
     \sigma^2_s = \boldsymbol{t}^\top \boldsymbol{\Sigma}_n \boldsymbol{t}\,.
 \end{equation}

\autoref{fig:sigma} shows a comparison between our approximation and a the true residual distribution, given noisy image points. Do to the unprojection of the image points to bearing vectors, the trace of the bearing vector covariances is small for a focal length $f$ of ca. 720 pixels on the KITTI dataset, since $\text{tr}(\boldsymbol{\Sigma}) \sim 1 / f^2$. Given the small covariances, $\tilde{\boldsymbol{\Sigma}}$ is several magnitudes smaller than the other terms, making the approximation accurate. \autoref{fig:sigma_focal} shows the correlation between the variance and the focal length. 

\section{Gradient} \label{sec:grad}
 In this section, we show that the gradient $\partial \mathcal{L} / \partial \boldsymbol{\Sigma}_{2\text{D}}$ is restricted by the problem geometry. We state the components needed to obtain $\partial \mathcal{L} / \partial \boldsymbol{\Sigma}_{2\text{D}}$ and show, how the geometry restricts their direction. Therefore, given a constant geometry the overall gradient direction only moves little throughout the training. 

We start by rewriting the residual $e_s$ of symmetric \ac{pnec} energy function as
 \begin{equation}
     e_s = \frac{n}{\sigma_s} = \frac{n}{\sqrt{d_{\boldsymbol{\Sigma}} + d_{\boldsymbol{\Sigma}^\prime}}}\,,
 \end{equation}
 for easier differentiation, with the components
 \begin{align}
     n & = \boldsymbol{t}^\top \hat{\boldsymbol{f}} \exp{\hat{\boldsymbol{x}}} \boldsymbol{R} \boldsymbol{f}^\prime \boldsymbol{f}^\prime{}^\top \boldsymbol{R}^\top \exp{\hat{\boldsymbol{x}}}^\top \hat{\boldsymbol{f}}{}^\top \boldsymbol{t}\,, \\
     d_{\boldsymbol{\Sigma}} & = \left(\left(\exp{\hat{\boldsymbol{x}}} \boldsymbol{R} \boldsymbol{f}^\prime \right) \times \boldsymbol{t} 
     \right){}^\top
     \boldsymbol{\Sigma} \left(\left(\exp{\hat{\boldsymbol{x}}} \boldsymbol{R} \boldsymbol{f}^\prime \right) \times \boldsymbol{t} 
     \right)\,, \\
     d_{\boldsymbol{\Sigma}^\prime} & = \boldsymbol{t}^\top \hat{\boldsymbol{f}} \exp{\hat{\boldsymbol{x}}} \boldsymbol{R} \boldsymbol{\Sigma}^\prime  \boldsymbol{R}^\top \exp{\hat{\boldsymbol{x}}}^\top \hat{\boldsymbol{f}}{}^\top \boldsymbol{t}\,.
 \end{align}
 Since we are working with rotations in $\text{SO}(3)$ we differentiate with regard to $\boldsymbol{x} \in \text{so}(3)$ around the identity rotation. This gives us the following gradients
 \begin{align}
     \frac{\partial n}{\partial \boldsymbol{x}} & = 2 \left( ( \boldsymbol{R} \boldsymbol{f}^\prime \boldsymbol{f}^\prime{}^\top \boldsymbol{R}^\top \exp{\hat{\boldsymbol{x}}}^\top \hat{\boldsymbol{f}}{}^\top \boldsymbol{t}) \times (\hat{\boldsymbol{f}} \boldsymbol{t}) \right)^\top \,,  \\
     \frac{\partial d_{\boldsymbol{\Sigma}}}{\partial \boldsymbol{x}} & = 2 \left( ( \boldsymbol{R} \boldsymbol{f}^\prime ) \times (\hat{\boldsymbol{t}} \boldsymbol{\Sigma} \hat{\boldsymbol{t}}^\top \exp{\hat{\boldsymbol{x}}} \boldsymbol{R} \boldsymbol{f}^\prime) \right)^\top \,,  \\
     \frac{\partial d_{\boldsymbol{\Sigma}^\prime}}{\partial \boldsymbol{x}} & = 2 \left(( \boldsymbol{R} \boldsymbol{\Sigma}^\prime \boldsymbol{R}^\top \exp{\hat{\boldsymbol{x}}}^\top \hat{\boldsymbol{f}}{}^\top \boldsymbol{t}) \times (\hat{\boldsymbol{f}} \boldsymbol{t} ) \right)^\top \,,
 \end{align}
 with regard to the rotation. The direction of each gradient is restricted by the cross product. The gradient for the residual is given by
 \begin{equation}
     \frac{\partial e_s}{\partial \boldsymbol{x}} = \frac{1}{\sigma_s}\frac{\partial n}{\partial \boldsymbol{x}} - \frac{n}{2 \sigma_s^3} \left(\frac{\partial d_{\boldsymbol{\Sigma}}}{\partial \boldsymbol{x}} + \frac{\partial d_{\boldsymbol{\Sigma}^\prime}}{\partial \boldsymbol{x}} \right)\,.
 \end{equation}
 The gradients with regard to the bearing vector covariances are solely dependent on the geometry as they are given by
 \begin{align}
     \frac{\partial d_{\boldsymbol{\Sigma}}}{\partial \boldsymbol{\Sigma}} & = \left(\boldsymbol{t} \times (\exp{\hat{\boldsymbol{x}}} \boldsymbol{R} \boldsymbol{f}^\prime)\right) \left(\boldsymbol{t} \times (\exp{\hat{\boldsymbol{x}}} \boldsymbol{R} \boldsymbol{f}^\prime)\right)^\top \,, \\
     \frac{\partial d_{\boldsymbol{\Sigma}^\prime}}{\partial \boldsymbol{\Sigma}^\prime} & = \left( \boldsymbol{R}^\top \exp{\hat{\boldsymbol{x}}}^\top \hat{\boldsymbol{f}}{}^\top \boldsymbol{t} \right) \left( \boldsymbol{R}^\top \exp{\hat{\boldsymbol{x}}}^\top \hat{\boldsymbol{f}}{}^\top \boldsymbol{t} \right)^\top \,.
 \end{align}
 The gradients of the residual are given by
 \begin{align}
     \frac{\partial e_s}{\partial \boldsymbol{\Sigma}} & = - \frac{n}{2 \sigma_s^3} \frac{\partial d_{\boldsymbol{\Sigma}}}{\partial \boldsymbol{\Sigma}} \,, \\
     \frac{\partial e_s}{\partial \boldsymbol{\Sigma}^\prime} & = - \frac{n}{2 \sigma_s^3} \frac{\partial d_{\boldsymbol{\Sigma}^\prime}}{\partial \boldsymbol{\Sigma}^\prime} \,.
 \end{align}
 Since all components are restricted by the geometry of the problem, the overall gradient is somewhat restricted as well. We show this empirically in the following.

\begin{figure*}[t]
    \centering
    \begin{subfigure}[b]{0.3\textwidth}
        \centering
        {\includegraphics[trim={0.25cm 0.3cm 0.25cm 0.2cm},clip,width=\textwidth]{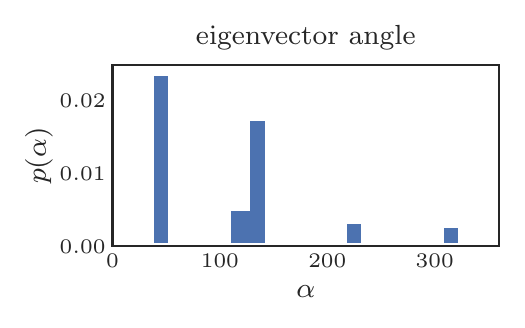}}
        \caption{10 epochs}
        \label{fig:hist10}
    \end{subfigure}
    \vspace{0.0cm}
    \begin{subfigure}[b]{0.3\textwidth}
        \centering
        {\includegraphics[trim={0.25cm 0.3cm 0.25cm 0.2cm},clip,width=\textwidth]{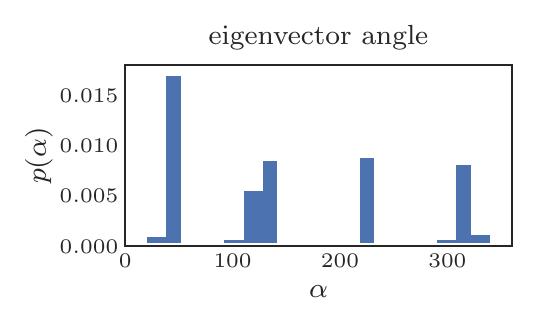}}
        \caption{50 epochs}
        \label{fig:hist50}
    \end{subfigure}
    \begin{subfigure}[b]{0.3\textwidth}
        \centering
        {\includegraphics[trim={0.25cm 0.3cm 0.25cm 0.2cm},clip,width=\textwidth]{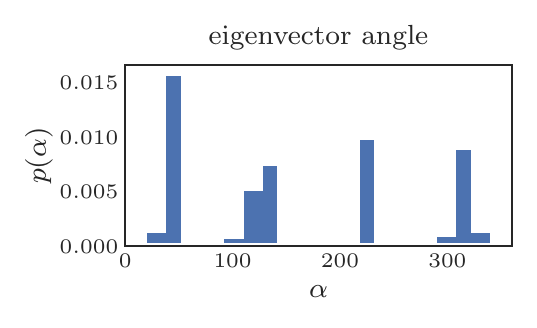}}
        \caption{100 epochs}
        \label{fig:hist100}
    \end{subfigure}
    \vspace{-0.35cm}
    \caption{
        Histogram of eigenvector angles for the gradient $\partial \mathcal{L} / \partial \boldsymbol{\Sigma}_{2\text{D}}^\prime$ after 10, 50, and 100 epochs. The histogram shows 4 distinct peaks, with only a few points in between. This shows the limited direction that the gradients have, making it difficult to learn the true distribution of the covariances with little diversity in the training data.
    }
\label{fig:hist}
\vspace{-0.45cm}
\end{figure*}

\begin{figure*}[t]
    \centering
    \begin{subfigure}[b]{0.28\textwidth}
        \centering
        {\includegraphics[trim={0.25cm 0.3cm 0.25cm 0.2cm},clip,width=\textwidth]{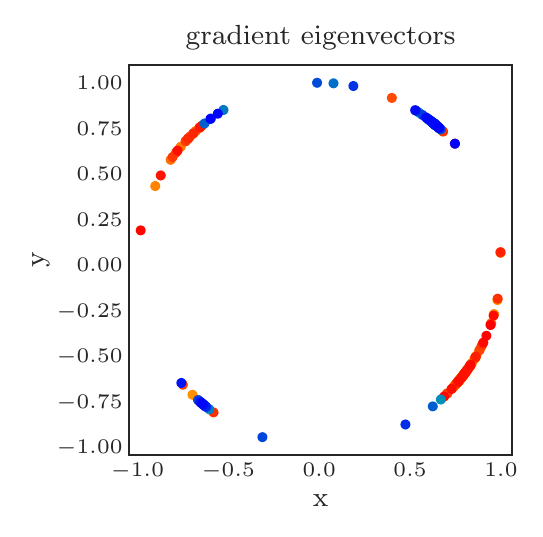}}
        \caption{10 epochs}
        \label{fig:circ10}
    \end{subfigure}
    \vspace{0.0cm}
    \begin{subfigure}[b]{0.28\textwidth}
        \centering
        {\includegraphics[trim={0.25cm 0.3cm 0.25cm 0.2cm},clip,width=\textwidth]{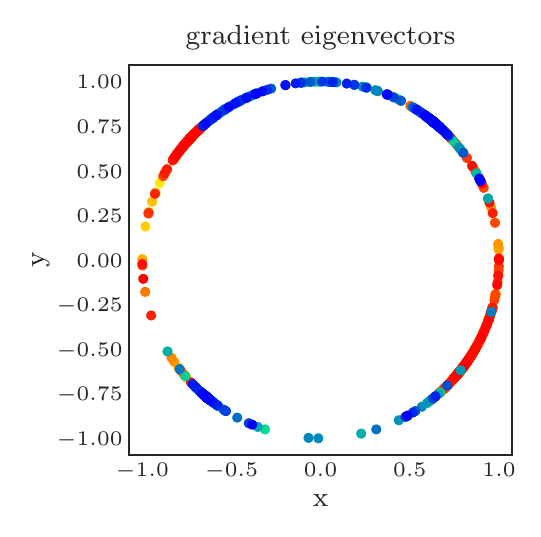}}
        \caption{50 epochs}
        \label{fig:circ50}
    \end{subfigure}
    \begin{subfigure}[b]{0.405\textwidth}
        \centering
        {\includegraphics[trim={0.25cm 0.3cm 0.25cm 0.2cm},clip,width=\textwidth]{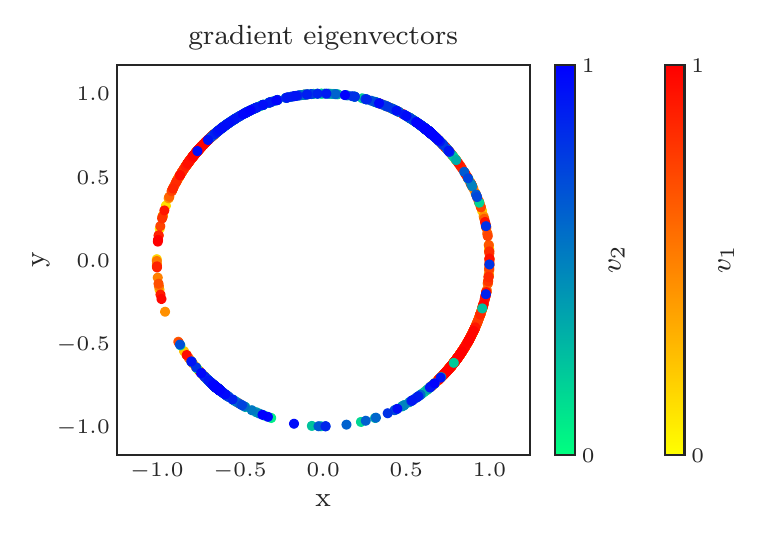}}
        \caption{100 epochs}
        \label{fig:circ100}
    \end{subfigure}
    \vspace{-0.35cm}
    \caption{
        Distribution of eigenvectors of the gradient $\partial \mathcal{L} / \partial \boldsymbol{\Sigma}_{2\text{D}}^\prime$ after 10, 50, and 100 epochs. Eigenvectors are color coded (green to blue and yellow to red) depending, whether there are the 1st or 2nd eigenvector and their epoch. While after 100 epochs most of the circle is covered, the eigenvectors aggregate at certain positions. Especially after 10 epochs, the eigenvectors are sparsely distributed. This shows a limited range of directions for the gradient.
    }
\label{fig:circ}
\vspace{-0.45cm}
\end{figure*}

 \autoref{fig:hist} and \autoref{fig:circ} give the distribution of the gradient for the first experiment on synthetic data, where all individual problems share the same geometric setup. \autoref{fig:circ} shows the eigenvectors of $\partial \mathcal{L} / \partial \boldsymbol{\Sigma}_{2\text{D}}$ for one covariance in the image plane. After 10 epochs of training, the eigenvectors are mainly located at 4 distinct regions, showing the restriction of the gradient direction. Even after 100 epochs of training certain regions show only few eigenvectors. The angular distribution of the eigenvectors in \autoref{fig:hist} show 4 distinct peaks, with almost no eigenvectors in between.

\begin{figure*}[t]
    \centering
    \begin{subfigure}[b]{0.3\textwidth}
        \centering
        {\includegraphics[trim={0.25cm 0.3cm 0.25cm 0.2cm},clip,width=\textwidth]{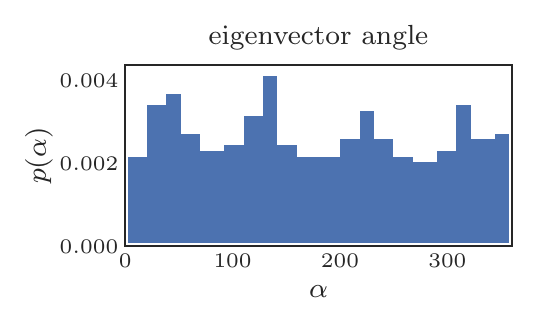}}
        \caption{10 epochs}
        \label{fig:adv_hist10}
    \end{subfigure}
    \vspace{0.0cm}
    \begin{subfigure}[b]{0.3\textwidth}
        \centering
        {\includegraphics[trim={0.25cm 0.3cm 0.25cm 0.2cm},clip,width=\textwidth]{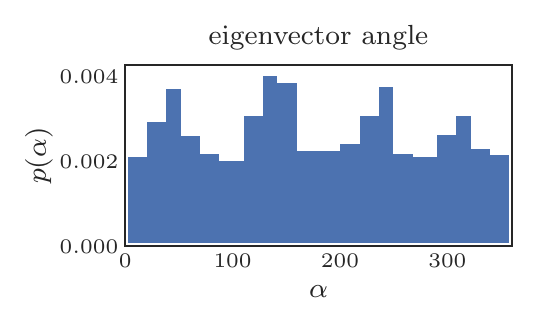}}
        \caption{50 epochs}
        \label{fig:adv_hist50}
    \end{subfigure}
    \begin{subfigure}[b]{0.3\textwidth}
        \centering
        {\includegraphics[trim={0.25cm 0.3cm 0.25cm 0.2cm},clip,width=\textwidth]{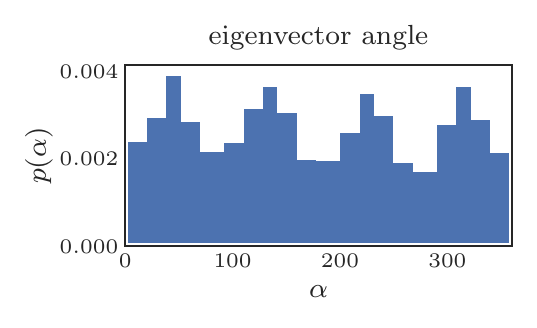}}
        \caption{100 epochs}
        \label{fig:adv_hist100}
    \end{subfigure}
    \vspace{-0.35cm}
    \caption{
        Histogram of eigenvector angles for the gradient $\partial \mathcal{L} / \partial \boldsymbol{\Sigma}_{2\text{D}}^\prime$ after 10, 50, and 100 epochs. While it shows 4 distinct peaks, event after only 10 epochs many points lie in between. The direction of the gradient is not limited, allowing for a better fit to the ground truth noise distribution.
    }
\label{fig:adv_hist}
\vspace{-0.45cm}
\end{figure*}

\begin{figure*}[t]
    \centering
    \begin{subfigure}[b]{0.28\textwidth}
        \centering
        {\includegraphics[trim={0.25cm 0.3cm 0.25cm 0.2cm},clip,width=\textwidth]{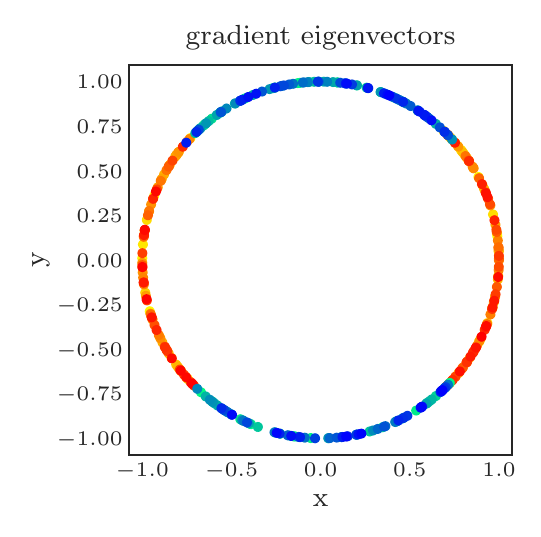}}
        \caption{10 epochs}
        \label{fig:adv_circ10}
    \end{subfigure}
    \vspace{0.0cm}
    \begin{subfigure}[b]{0.28\textwidth}
        \centering
        {\includegraphics[trim={0.25cm 0.3cm 0.25cm 0.2cm},clip,width=\textwidth]{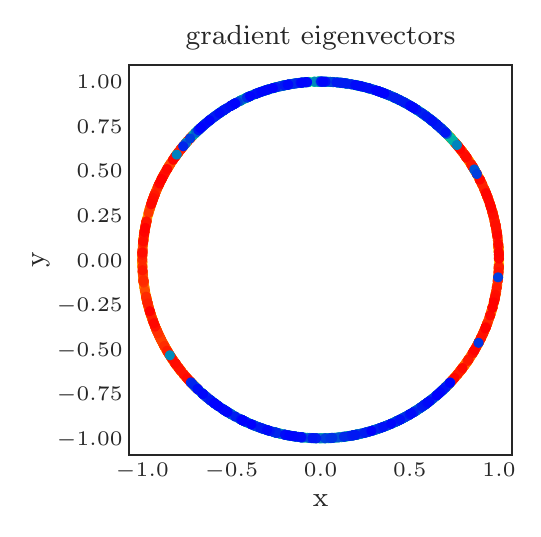}}
        \caption{50 epochs}
        \label{fig:adv_circ50}
    \end{subfigure}
    \begin{subfigure}[b]{0.405\textwidth}
        \centering
        {\includegraphics[trim={0.25cm 0.3cm 0.25cm 0.2cm},clip,width=\textwidth]{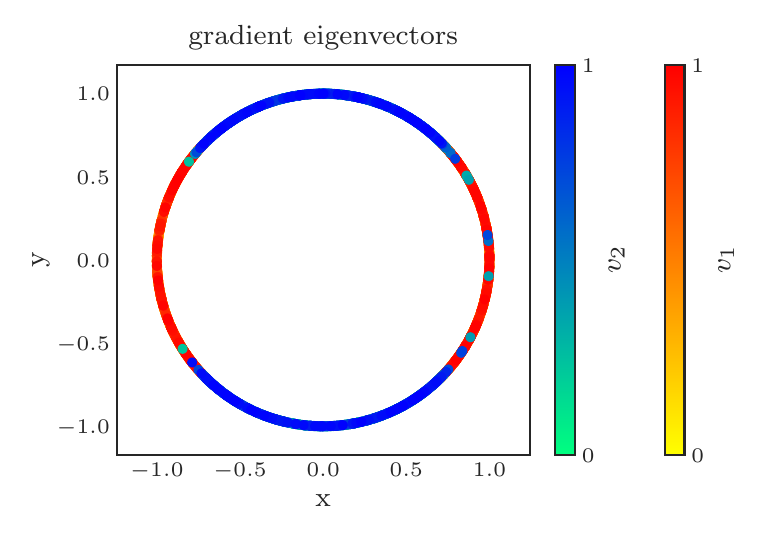}}
        \caption{100 epochs}
        \label{fig:adv_circ100}
    \end{subfigure}
    \vspace{-0.35cm}
    \caption{
        Distribution of eigenvectors of the gradient $\partial \mathcal{L} / \partial \boldsymbol{\Sigma}_{2\text{D}}^\prime$ after 10, 50, and 100 epochs. Eigenvectors are color coded (green to blue and yellow to red) depending, whether there are the 1st or 2nd eigenvector and their epoch. Even after 10 epochs, the eigenvectors are evenly distributed. This show, that the gradient has no limit for its direction, allowing for a better fit to the noise distribution even in the image plane.
    }
\label{fig:adv_circ}
\vspace{-0.45cm}
\end{figure*}

 \autoref{fig:adv_hist} and \autoref{fig:adv_circ} show the distribution of the gradient for the second experiment on synthetic data, with more diverse data. Given the diverse data, there are eigenvectors in all directions, even after 10 epochs. \autoref{fig:adv_hist} still shows 4 distinct peaks, however there is no sparsity in the distribution.

 The sparse distribution of the gradient direction prohibit learning the correct noise distribution for the first experiment. Only the residual variance is correctly estimated. However, the introduction of diverse data with different geometries removes this restriction, leading better covariance estimates.

 \begin{table}[t]
\small
\centering
\sisetup{detect-weight,mode=text}
\renewrobustcmd{\bfseries}{\fontseries{b}\selectfont}
\renewrobustcmd{\boldmath}{}
\newrobustcmd{\B}{\bfseries}
\addtolength{\tabcolsep}{-3.5pt}

\begin{tabular} {p{3.6cm} c|c}
\toprule
Hyperparameter & KITTI & EuRoC \\
\midrule
optimizer & ADAM & ADAM \\
$\beta_1$ & $0.9$ & $0.9$ \\ 
$\beta_2$ & $0.999$ & $0.999$ \\
learning rate & $5\cdot 10^{-4}$ & $5\cdot 10^{-4}$ \\
\midrule
PNEC and theseus & & \\
\midrule
regularization & $10^{-13}$ & $10^{-13}$ \\
damping & $10^7$ & $10^7$ \\
iterations & $100$ & $100$ \\
\midrule
RANSAC & & \\
\midrule
iterations & $5000$ & $5000$ \\
threshold & $10^{-6}$ & $8\cdot 10^{-7}$ \\
\bottomrule
\end{tabular}
\vspace{-0.3cm}
\caption{Parameters used for training and evaluation.}
\label{tab:kitti_parameters}
\vspace{-0.5cm}
\end{table}

 \vspace{-0.2cm}
 \section{Hyperparameters} \label{sec:training}
 This section details the training and evaluation parameters for our \ac{dnls} framework for estimating noise distributions of keypoints. All models are trained on two RTX 5000 GPUs with 16GB of memory for around 3 days. We use a UNet architecture with 3 output channels for predicting the uncertainty parameters. The UNet has 4 down convolutions and 4 up convolutions with $32, 64, 128, 256$ and $128, 64, 32, 16$ channels, respectively. \autoref{tab:SuperPoint_parameters} gives the SuperPoint and SuperGlue hyperparameters for training and evaluation. For our supervised training, we train on consecutive image pairs of the training sequences. For our self-supervised training we create the training tuples from 3 consecutive images. When training with SuperPoint, we crop the images to size (1200, 300), whereas for KLT-Tracks, we crop it to (1200, 320). We found that reducing the height too much for KLT-tracks leads to not enough tracks. For evaluating with KLT-tracks on KITTI we change the following to \cite{muhle2022pnec}: instead of tracking keypoints over multiple images, we start with fresh keypoints for each image pair. To account for the symmetric \ac{pnec}, we slightly modify the uncertainty extraction. We use \cite[suppl., Eqn. (8)]{muhle2022pnec} as the uncertainty measure for the tracks in both frames. We found, that these changes already give better results than the ones stated \cite{muhle2022pnec}. \autoref{tab:kitti_parameters} gives the training parameter for optimizer, theseus and the \ac{pnec} energy function not stated in the main paper.
 \begin{table}[t]
\small
\centering
\sisetup{detect-weight,mode=text}
\renewrobustcmd{\bfseries}{\fontseries{b}\selectfont}
\renewrobustcmd{\boldmath}{}
\newrobustcmd{\B}{\bfseries}
\addtolength{\tabcolsep}{-3.5pt}

\begin{tabular} {p{3.6cm} c|c c}
\toprule
Hyperparameter & training & KITTI & EuRoC \\
\midrule
max keypoints & 256 & 2048 & 1024 \\
keypoint threshold & 0.005 & 0.005 & 0.0005 \\
nms radius & 3 & 3 & 3\\
\midrule
weights & outdoor & outdoor & indoor \\
sinkhorn iterations & 20 & 20 & 20 \\
match threshold & 0.5 & 0.5 & 0.01 \\
\bottomrule
\end{tabular}
\vspace{-0.3cm}
\caption{Hyperparameters for SuperPoint and SuperGlue during training and evaluation on the KITTI and EuRoC dataset.}
\label{tab:SuperPoint_parameters}
\vspace{-0.5cm}
\end{table}

 \section{Moving the Minimum} \label{sec:minimum}
\begin{figure*}[t]
    \centering
    \vspace{0.0cm}
    \begin{subfigure}[b]{0.32\textwidth}
        \centering
        {\includegraphics[trim={0.25cm 0.25cm 0.25cm 0.2cm},clip,width=\textwidth]{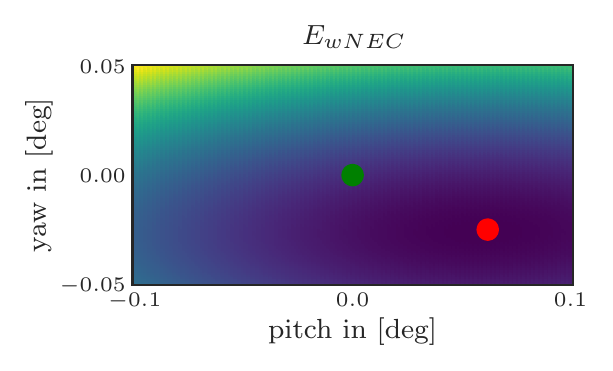}}
        \caption{weighted NEC-LS}
        \label{fig:minimum_sp_1}
    \end{subfigure}
    \begin{subfigure}[b]{0.32\textwidth}
        \centering
        {\includegraphics[trim={0.25cm 0.25cm 0.25cm 0.2cm},clip,width=\textwidth]{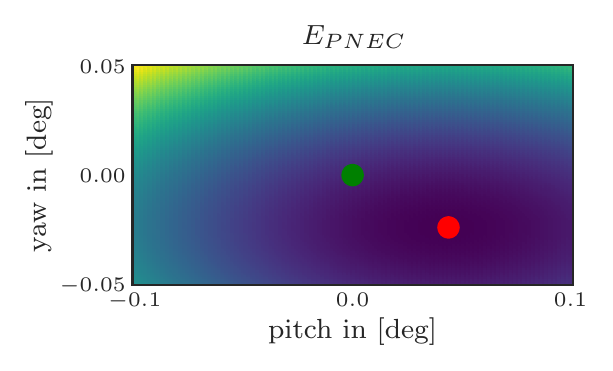}}
        \caption{Ours, supervised}
        \label{fig:minimum_sp_s_1}
    \end{subfigure}
    \begin{subfigure}[b]{0.32\textwidth}
        \centering
        {\includegraphics[trim={0.25cm 0.25cm 0.25cm 0.2cm},clip,width=\textwidth]{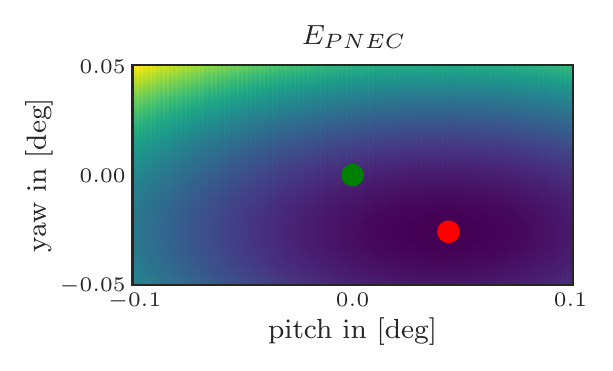}}
        \caption{Ours, self-supervised}
        \label{fig:minimum_sp_ss_1}
    \end{subfigure}
    \begin{subfigure}[b]{0.32\textwidth}
        \centering
        {\includegraphics[trim={0.25cm 0.25cm 0.25cm 0.2cm},clip,width=\textwidth]{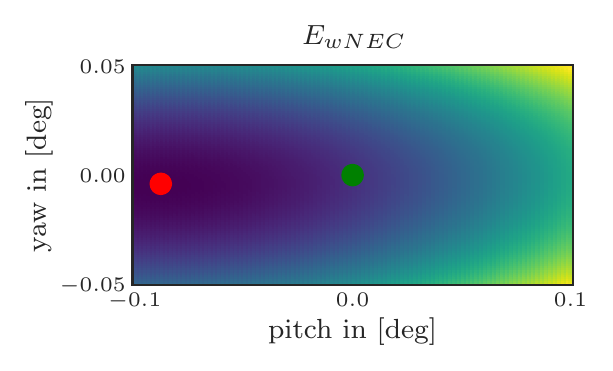}}
        \caption{weighted NEC-LS}
        \label{fig:minimum_sp_2}
    \end{subfigure}
    \begin{subfigure}[b]{0.32\textwidth}
        \centering
        {\includegraphics[trim={0.25cm 0.25cm 0.25cm 0.2cm},clip,width=\textwidth]{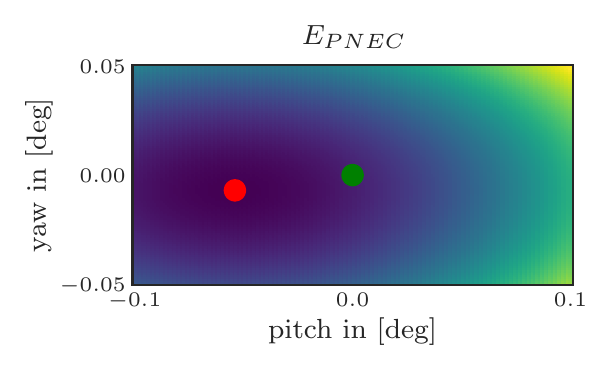}}
        \caption{Ours, supervised}
        \label{fig:minimum_sp_s_2}
    \end{subfigure}
    \begin{subfigure}[b]{0.32\textwidth}
        \centering
        {\includegraphics[trim={0.25cm 0.25cm 0.25cm 0.2cm},clip,width=\textwidth]{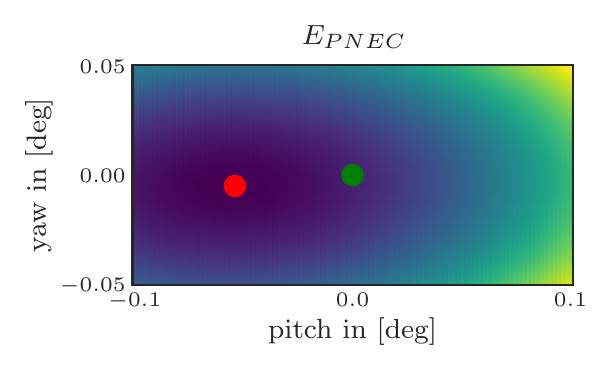}}
        \caption{Ours, self-supervised}
        \label{fig:minimum_sp_ss_2}
    \end{subfigure}
    \vspace{-0.25cm}
    \caption{
        Energy functions evaluated for rotations around the ground truth pose (green). Minimum of the cost function is marked in red. The energy function is evaluated for SuperPoint keypoint for two pose estimation problems on the KITTI dataset, filtered with RANSAC at the ground truth pose. We compare the weighted NEC-LS energy function to the \ac{pnec} energy function with our supervised and self-supervised covariances. While the overall shape of the energy function stays the same, our learned covariances move the minimum closer to the ground truth.
    }
\label{fig:minimum_sp}
\vspace{-0.45cm}
\end{figure*}

\begin{figure*}[t]
    \centering
    \vspace{0.0cm}
    \begin{subfigure}[b]{0.32\textwidth}
        \centering
        {\includegraphics[trim={0.25cm 0.25cm 0.25cm 0.2cm},clip,width=\textwidth]{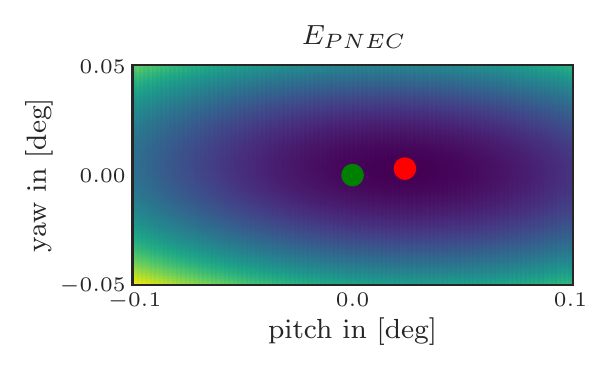}}
        \caption{KLT-PNEC}
        \label{fig:minimum_klt_1}
    \end{subfigure}
    \begin{subfigure}[b]{0.32\textwidth}
        \centering
        {\includegraphics[trim={0.25cm 0.25cm 0.25cm 0.2cm},clip,width=\textwidth]{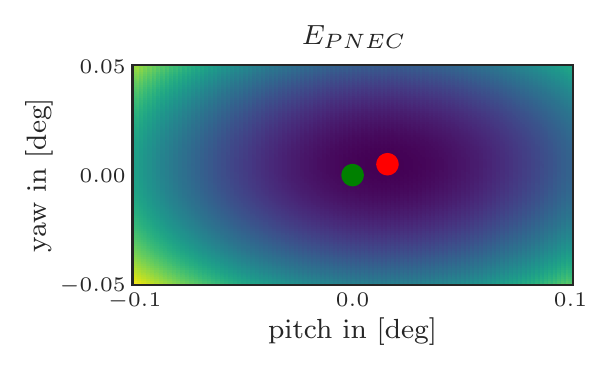}}
        \caption{Ours, supervised}
        \label{fig:minimum_klt_s_1}
    \end{subfigure}
    \begin{subfigure}[b]{0.32\textwidth}
        \centering
        {\includegraphics[trim={0.25cm 0.25cm 0.25cm 0.2cm},clip,width=\textwidth]{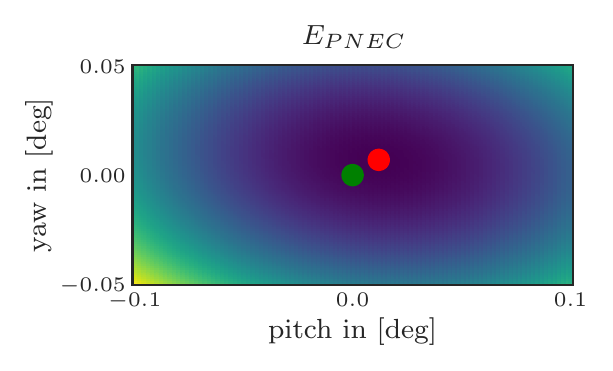}}
        \caption{Ours, self-supervised}
        \label{fig:minimum_klt_ss_1}
    \end{subfigure}
    \begin{subfigure}[b]{0.32\textwidth}
        \centering
        {\includegraphics[trim={0.25cm 0.25cm 0.25cm 0.2cm},clip,width=\textwidth]{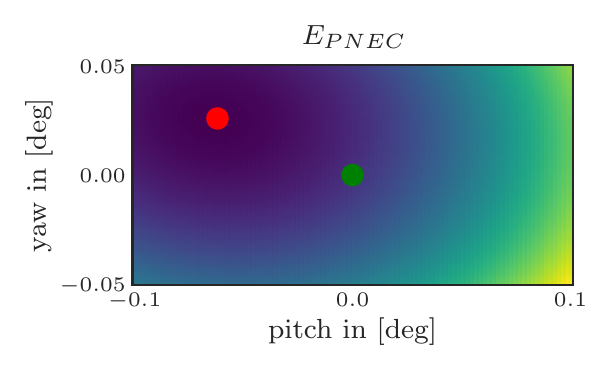}}
        \caption{KLT-PNEC}
        \label{fig:minimum_klt_2}
    \end{subfigure}
    \begin{subfigure}[b]{0.32\textwidth}
        \centering
        {\includegraphics[trim={0.25cm 0.25cm 0.25cm 0.2cm},clip,width=\textwidth]{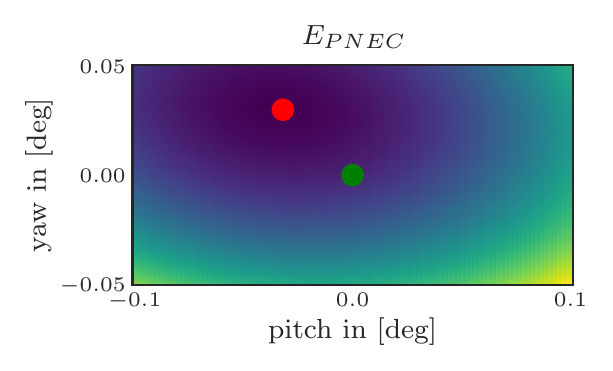}}
        \caption{Ours, supervised}
        \label{fig:minimum_klt_s_2}
    \end{subfigure}
    \begin{subfigure}[b]{0.32\textwidth}
        \centering
        {\includegraphics[trim={0.25cm 0.25cm 0.25cm 0.2cm},clip,width=\textwidth]{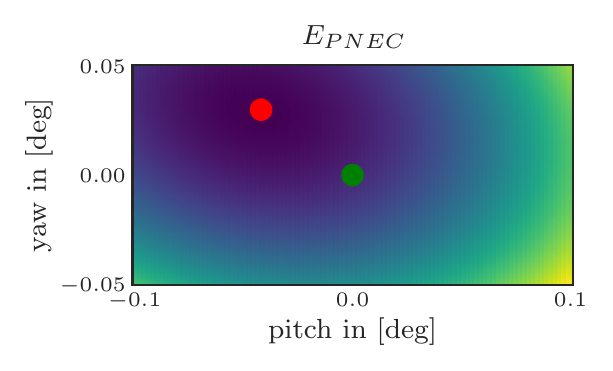}}
        \caption{Ours, self-supervised}
        \label{fig:minimum_klt_ss_2}
    \end{subfigure}
    \vspace{-0.25cm}
    \caption{
        Energy functions evaluated for rotations around the ground truth pose (green). Minimum of the cost function is marked in red. The energy function is evaluated for KLT-tracks for two pose estimation problems on the KITTI dataset, filtered with RANSAC at the ground truth pose. We compare the \ac{pnec} energy function using the KLT-covariances with our supervised and self-supervised covariances. While the overall shape of the energy function stays the same, our learned covariances move the minimum closer to the ground truth.
    }
\label{fig:minimum_klt}
\vspace{-0.45cm}
\end{figure*}

 \autoref{fig:minimum_klt} and \autoref{fig:minimum_sp} show examples for energy functions around the ground truth pose on the KITTI dataset. The energy functions are evaluated with keypoints filtered using the reprojection error also used in the RANSAC scheme of \cite{Kneip2014opengv} to remove outliers. We show the energy functions evaluated for rotations around the ground truth for yaw and pitch. While the overall shape of the energy function stays the same, our methods moves the minimum closer to the ground truth pose by learning the covariances.

 \section{Further Results} \label{sec:results}
 In this section we present additional results on the KITTI dataset, not presented in the main paper due to constrained space. We give the evaluation results for all sequences, training and test set. To present more comparisons with baseline methods, we replace the Nist\'er-5pt \cite{Nistr2004VisualO} with the 8pt \cite{longuet1987readings} algorithm. Furthermore, we replace the weighted NEC-LS and the KLT-PNEC. Instead, we add another \ac{pnec} method, where we approximate the error distribution using a reprojection error. Following \cite{Kneip2014opengv}, we triangulate a 3D point using the feature correspondence $\boldsymbol{p}_i, \boldsymbol{p}_i^\prime$ and the ground truth pose. We reproject the point into the images as $\tilde{\boldsymbol{p}}_i, \tilde{\boldsymbol{p}}_i^\prime$ and approximate the the error distribution as scaled isotropic covariances
 \begin{align}
     \boldsymbol{\Sigma}_{2\text{D}, i} & = \| \tilde{\boldsymbol{p}}_i -  \boldsymbol{p}_i\|^2 \boldsymbol{I}_2\,, \\
     \boldsymbol{\Sigma}^\prime_{2\text{D}, i} & = \| \tilde{\boldsymbol{p}}_i^\prime -  \boldsymbol{p}_i^\prime \|^2 \boldsymbol{I}_2\,.
 \end{align}
We clip the scale of the covariances at $0.01$ and $4.0$. \autoref{tab:kitti_superpoint_full} shows the results for the training and test set on KITTI with SuperPoint. While the reprojection method achieves the best results for the $\text{RPE}_1$ and $e_t$, our methods are often not far behind. This shows, that our network is capable and not too far off, when it comes to pose estimation. \autoref{tab:kitti_klt_full} shows the results for KITTI with \acs{klt}-tracks. 

\begin{figure*}[t]
    \centering
    \vspace{0.0cm}
    \begin{subfigure}[b]{0.3\textwidth}
        \centering
        {\includegraphics[trim={0.25cm 0.25cm 0.25cm 0.2cm},clip,width=\textwidth]{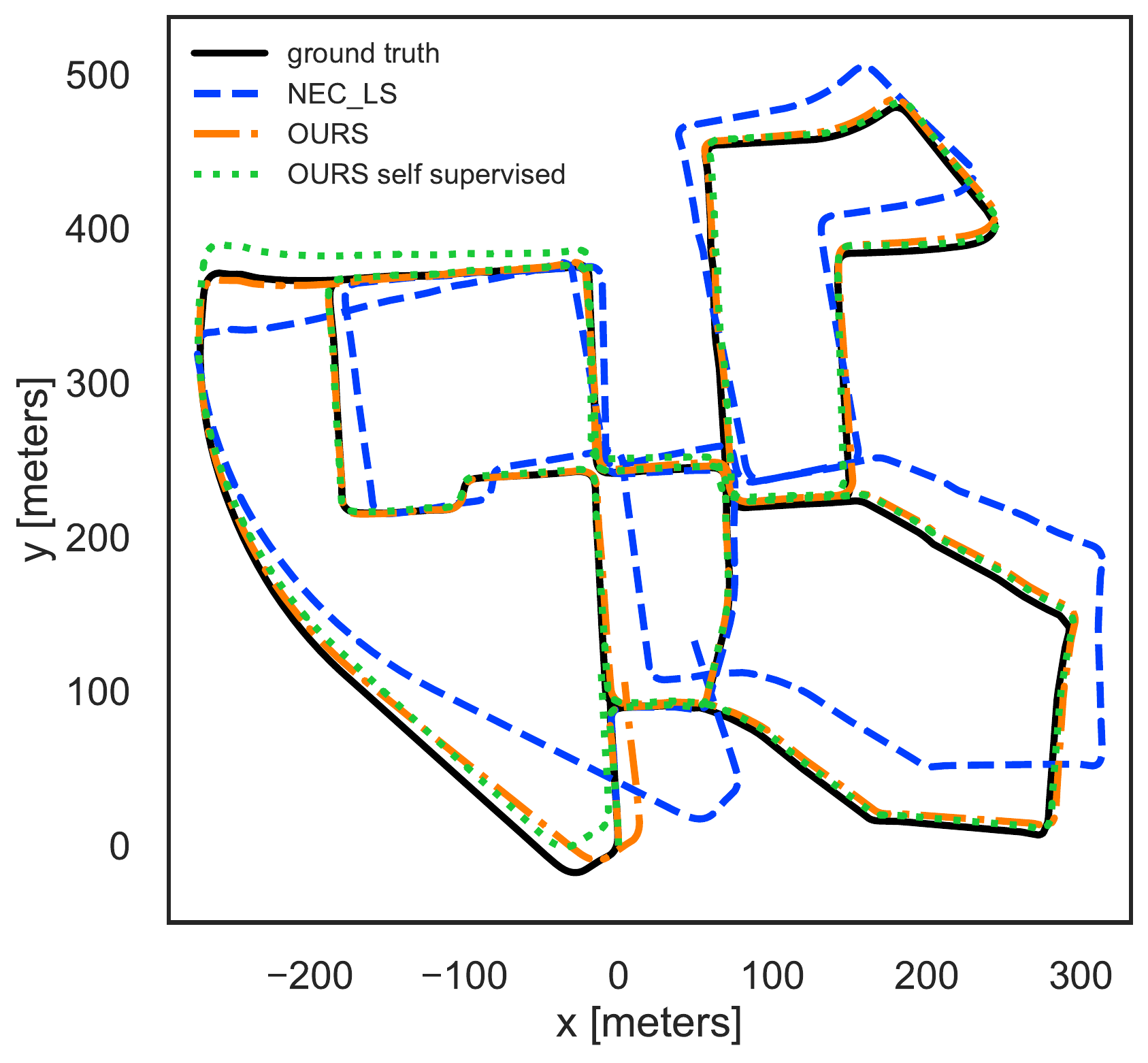}}
        \caption{sequence 00}
        \label{fig:traj_sp_00}
    \end{subfigure}
    \begin{subfigure}[b]{0.3\textwidth}
        \centering
        {\includegraphics[trim={0.25cm 0.25cm 0.25cm 0.2cm},clip,width=\textwidth]{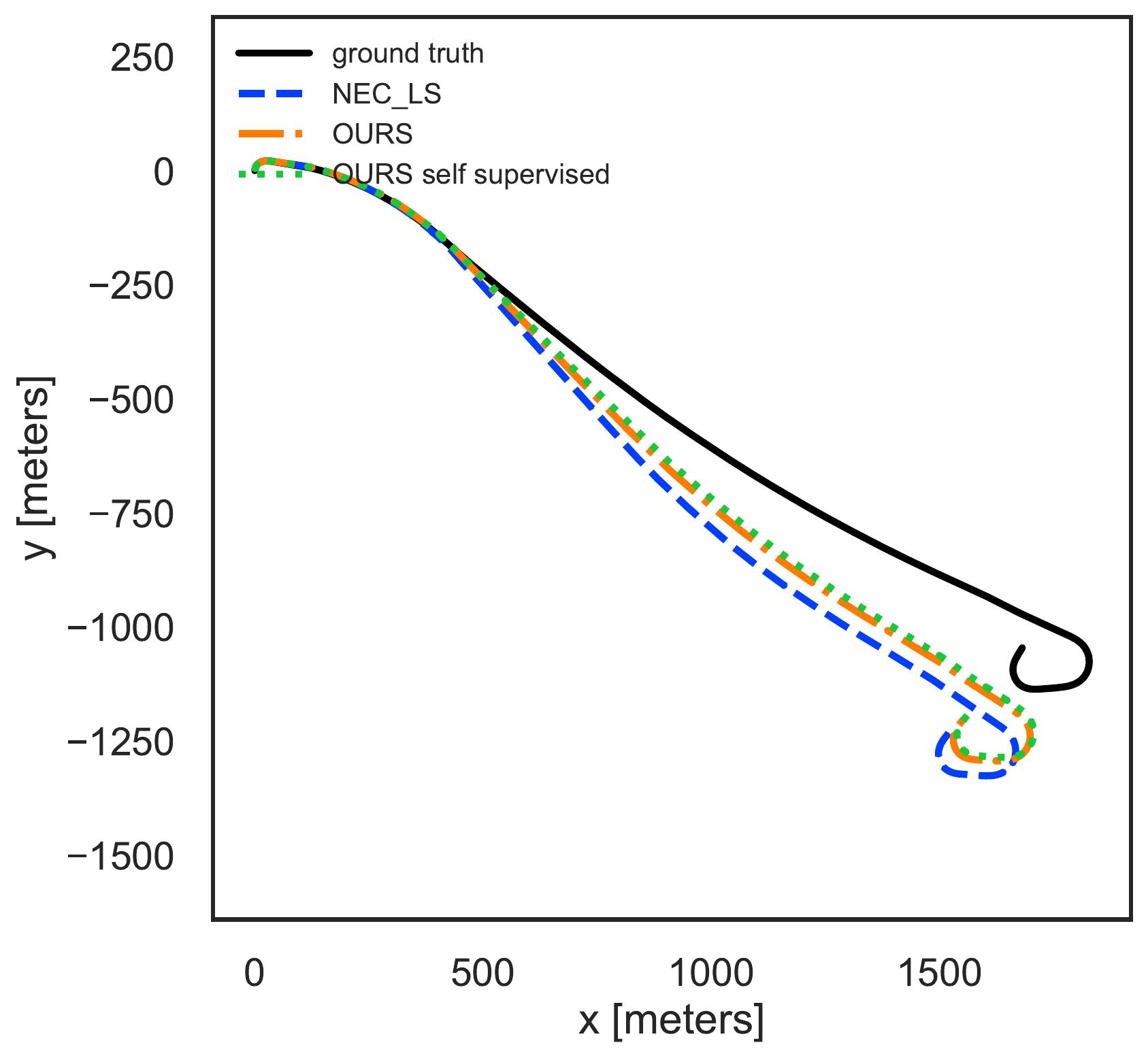}}
        \caption{sequence 01}
        \label{fig:traj_sp_01}
    \end{subfigure}
    \begin{subfigure}[b]{0.3\textwidth}
        \centering
        {\includegraphics[trim={0.25cm 0.25cm 0.25cm 0.2cm},clip,width=\textwidth]{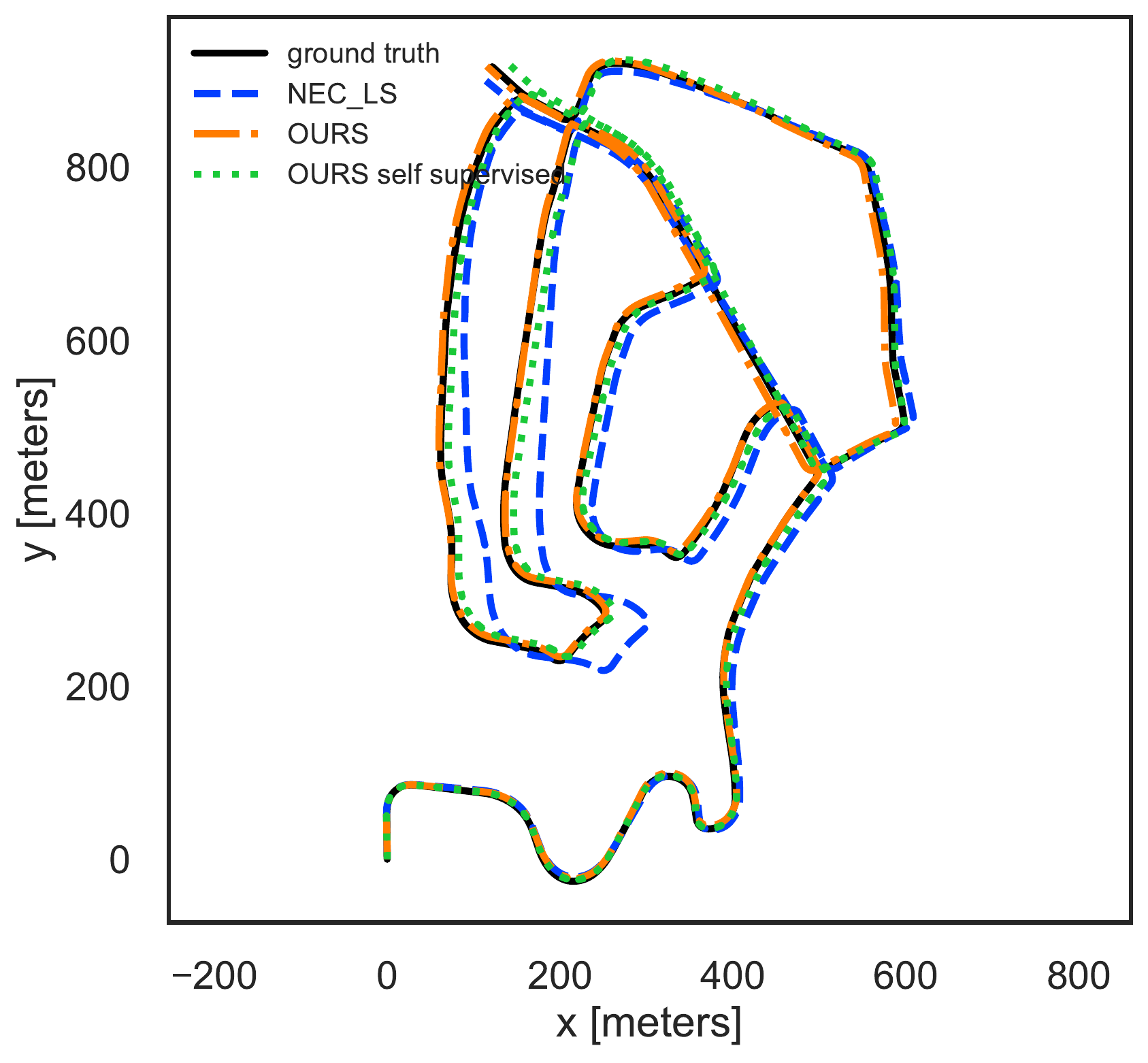}}
        \caption{sequence 02}
        \label{fig:traj_sp_02}
    \end{subfigure}
    \begin{subfigure}[b]{0.3\textwidth}
        \centering
        {\includegraphics[trim={0.25cm 0.25cm 0.25cm 0.2cm},clip,width=\textwidth]{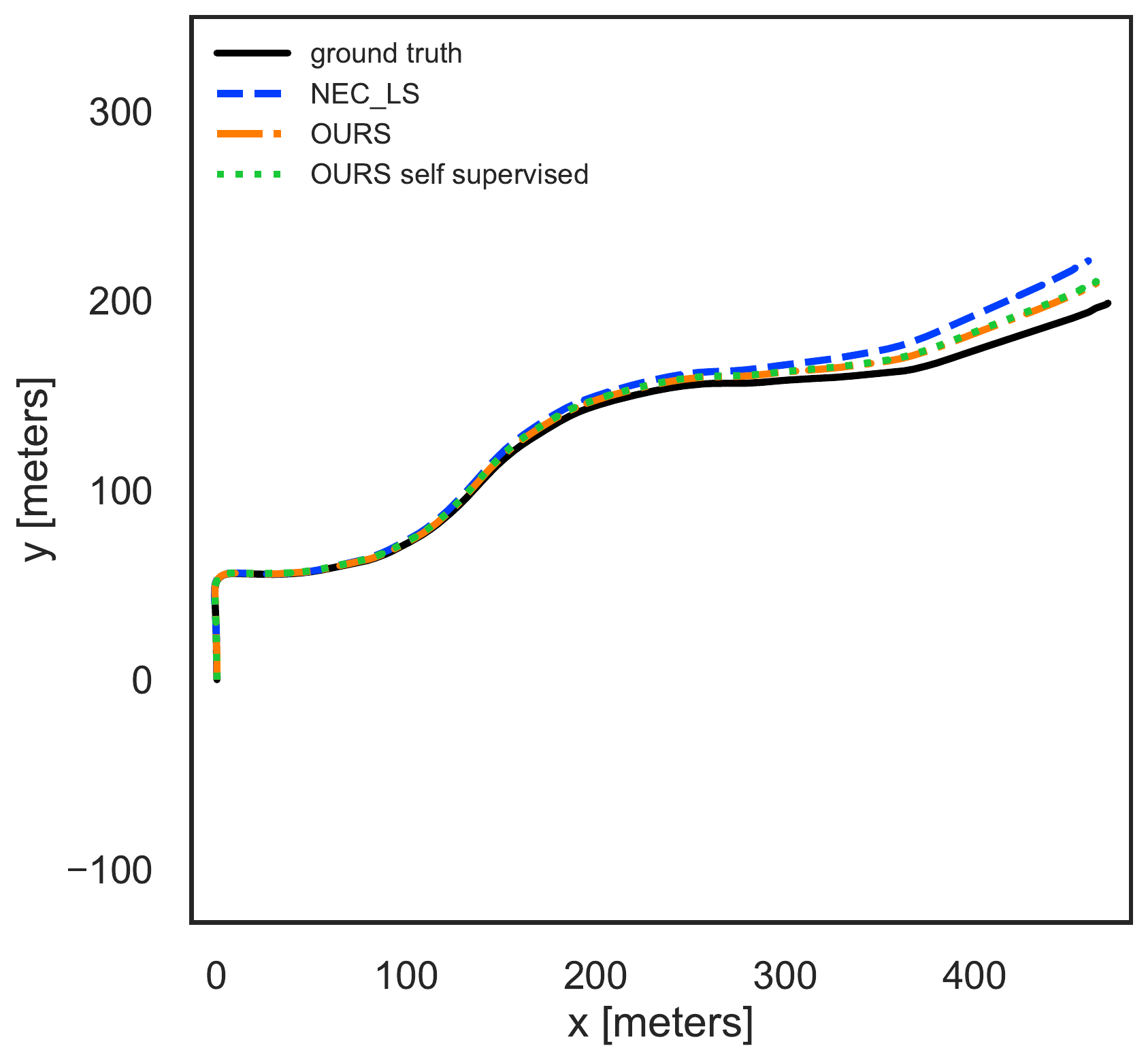}}
        \caption{sequence 03}
        \label{fig:traj_sp_03}
    \end{subfigure}
    \begin{subfigure}[b]{0.3\textwidth}
        \centering
        {\includegraphics[trim={0.25cm 0.25cm 0.25cm 0.2cm},clip,width=\textwidth]{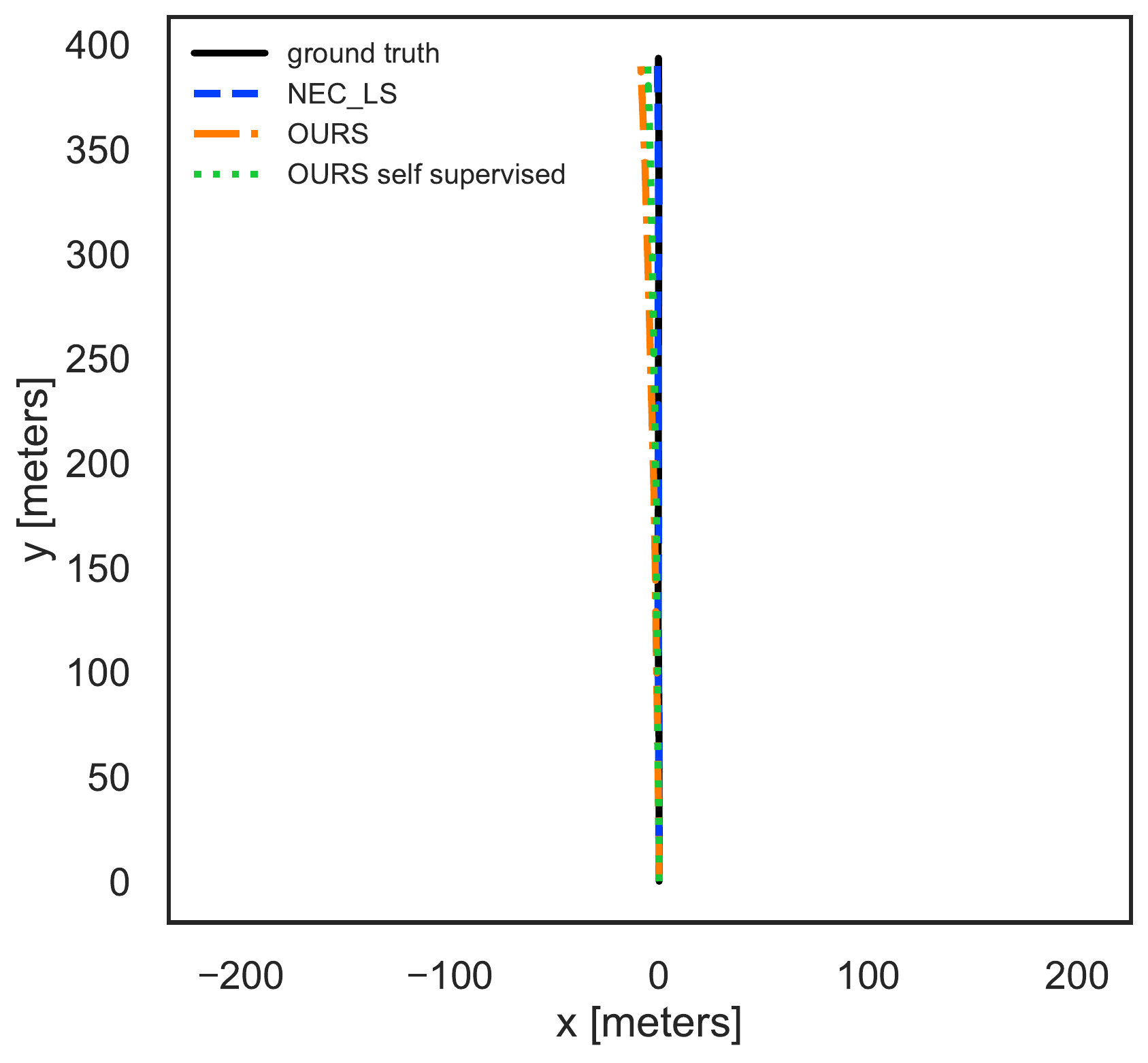}}
        \caption{sequence 04}
        \label{fig:traj_sp_04}
    \end{subfigure}
    \begin{subfigure}[b]{0.3\textwidth}
        \centering
        {\includegraphics[trim={0.25cm 0.25cm 0.25cm 0.2cm},clip,width=\textwidth]{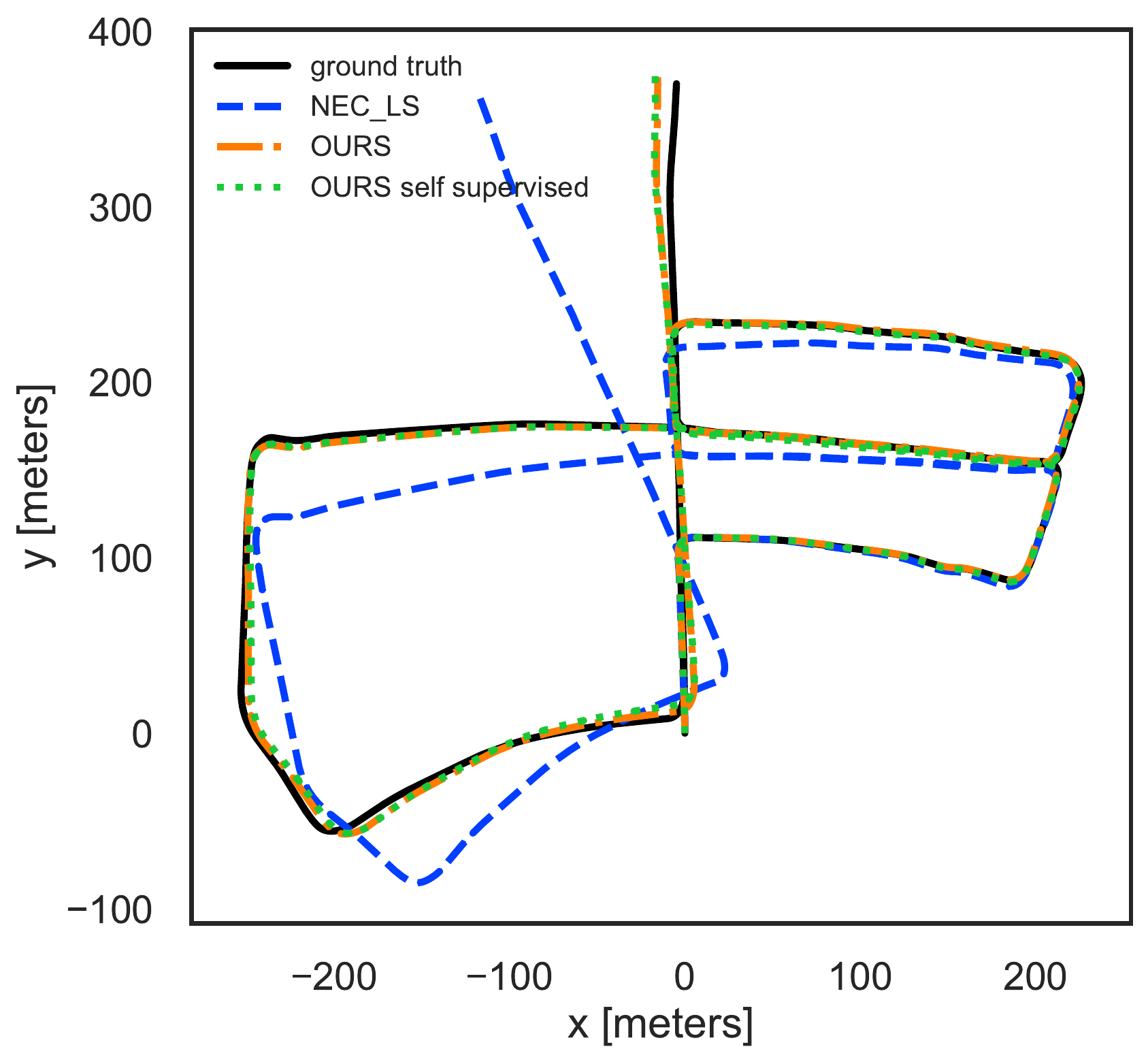}}
        \caption{sequence 05}
        \label{fig:traj_sp_05}
    \end{subfigure}
    \begin{subfigure}[b]{0.3\textwidth}
        \centering
        {\includegraphics[trim={0.25cm 0.25cm 0.25cm 0.2cm},clip,width=\textwidth]{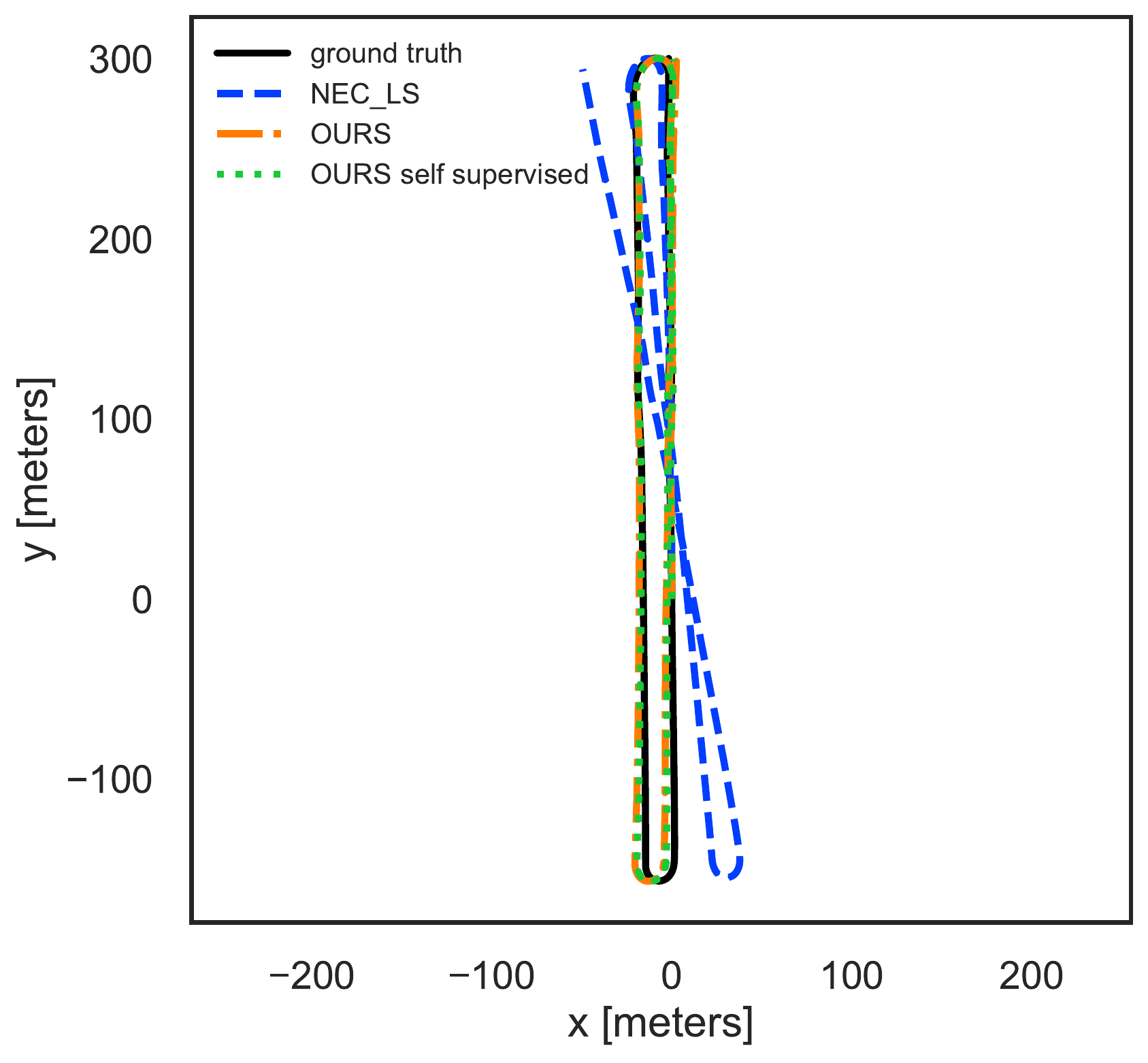}}
        \caption{sequence 06}
        \label{fig:traj_sp_06}
    \end{subfigure}
    \begin{subfigure}[b]{0.3\textwidth}
        \centering
        {\includegraphics[trim={0.25cm 0.25cm 0.25cm 0.2cm},clip,width=\textwidth]{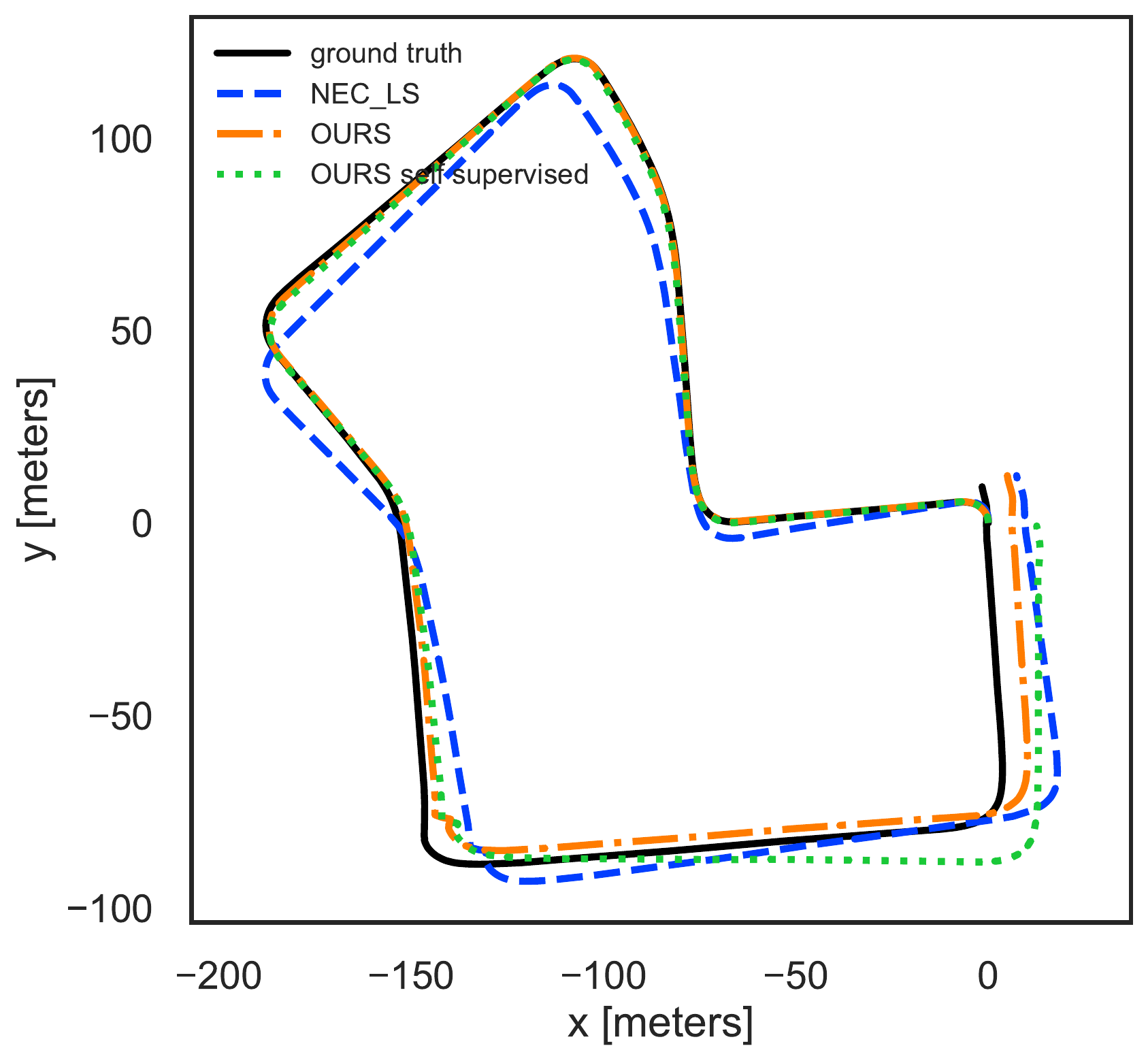}}
        \caption{sequence 07}
        \label{fig:traj_sp_07}
    \end{subfigure}
    \begin{subfigure}[b]{0.3\textwidth}
        \centering
        {\includegraphics[trim={0.25cm 0.25cm 0.25cm 0.2cm},clip,width=\textwidth]{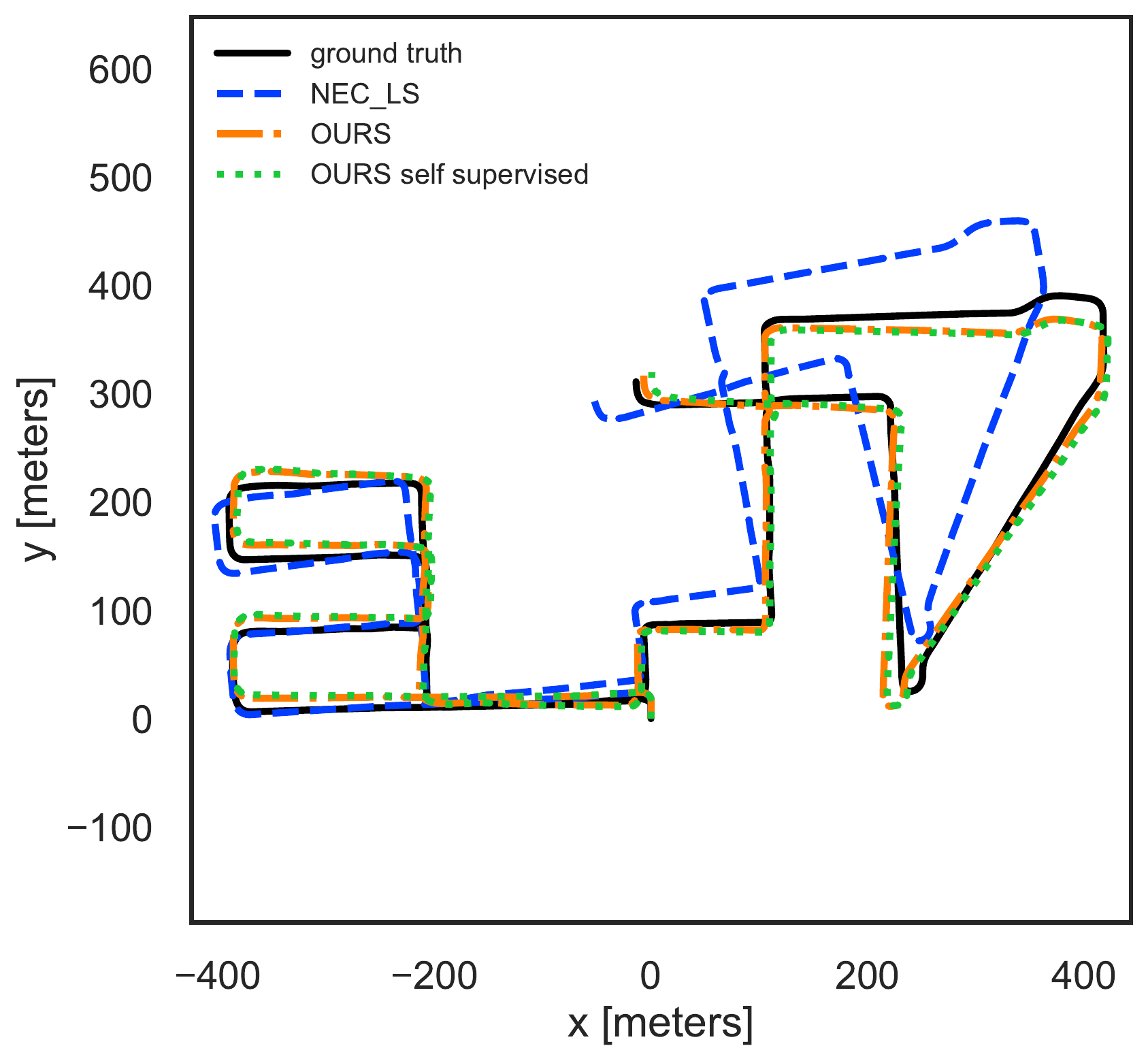}}
        \caption{sequence 08}
        \label{fig:traj_sp_08}
    \end{subfigure}
    \begin{subfigure}[b]{0.3\textwidth}
        \centering
        {\includegraphics[trim={0.25cm 0.25cm 0.25cm 0.2cm},clip,width=\textwidth]{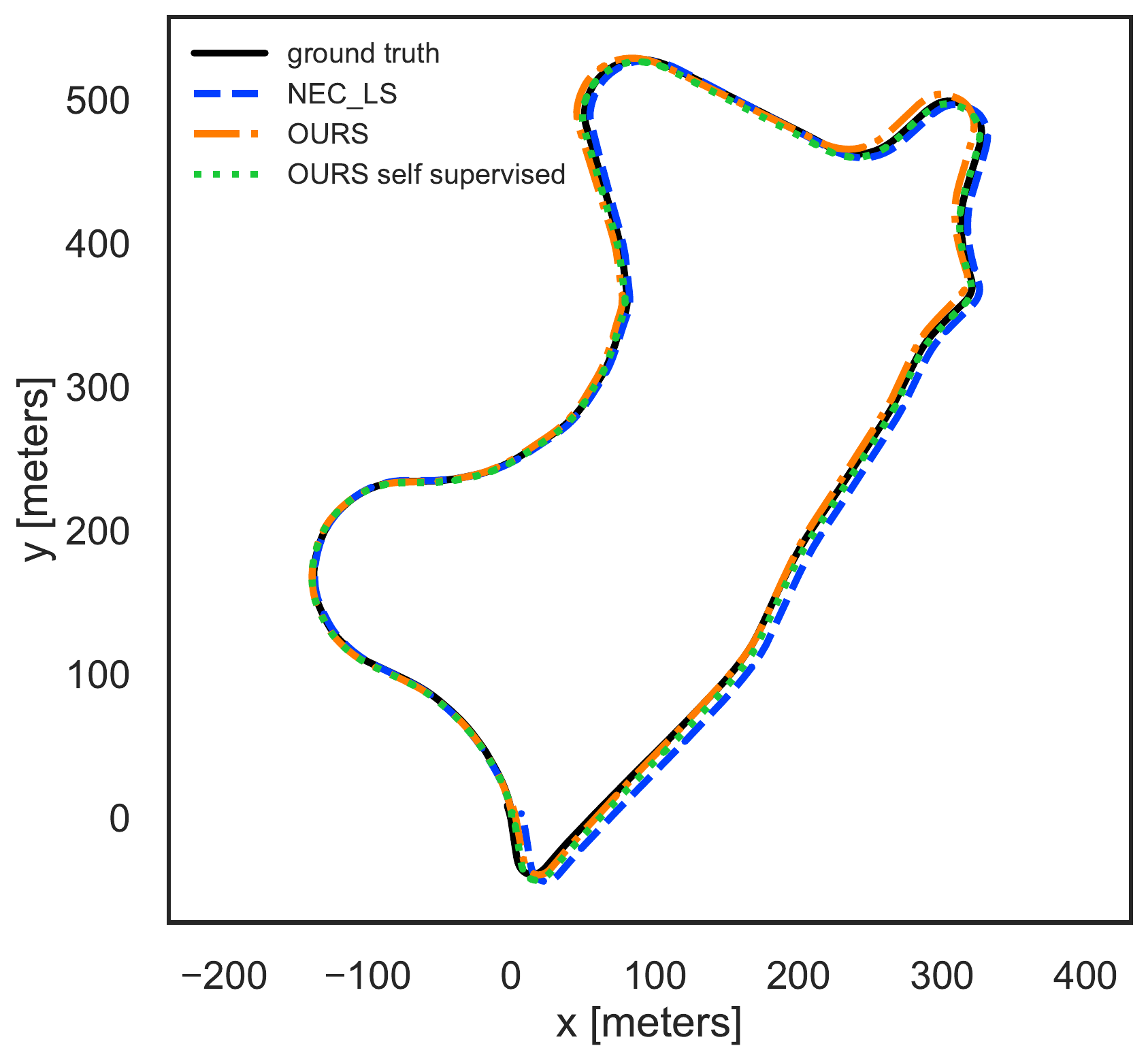}}
        \caption{sequence 09}
        \label{fig:traj_sp_09}
    \end{subfigure}
    \begin{subfigure}[b]{0.3\textwidth}
        \centering
        {\includegraphics[trim={0.25cm 0.25cm 0.25cm 0.2cm},clip,width=\textwidth]{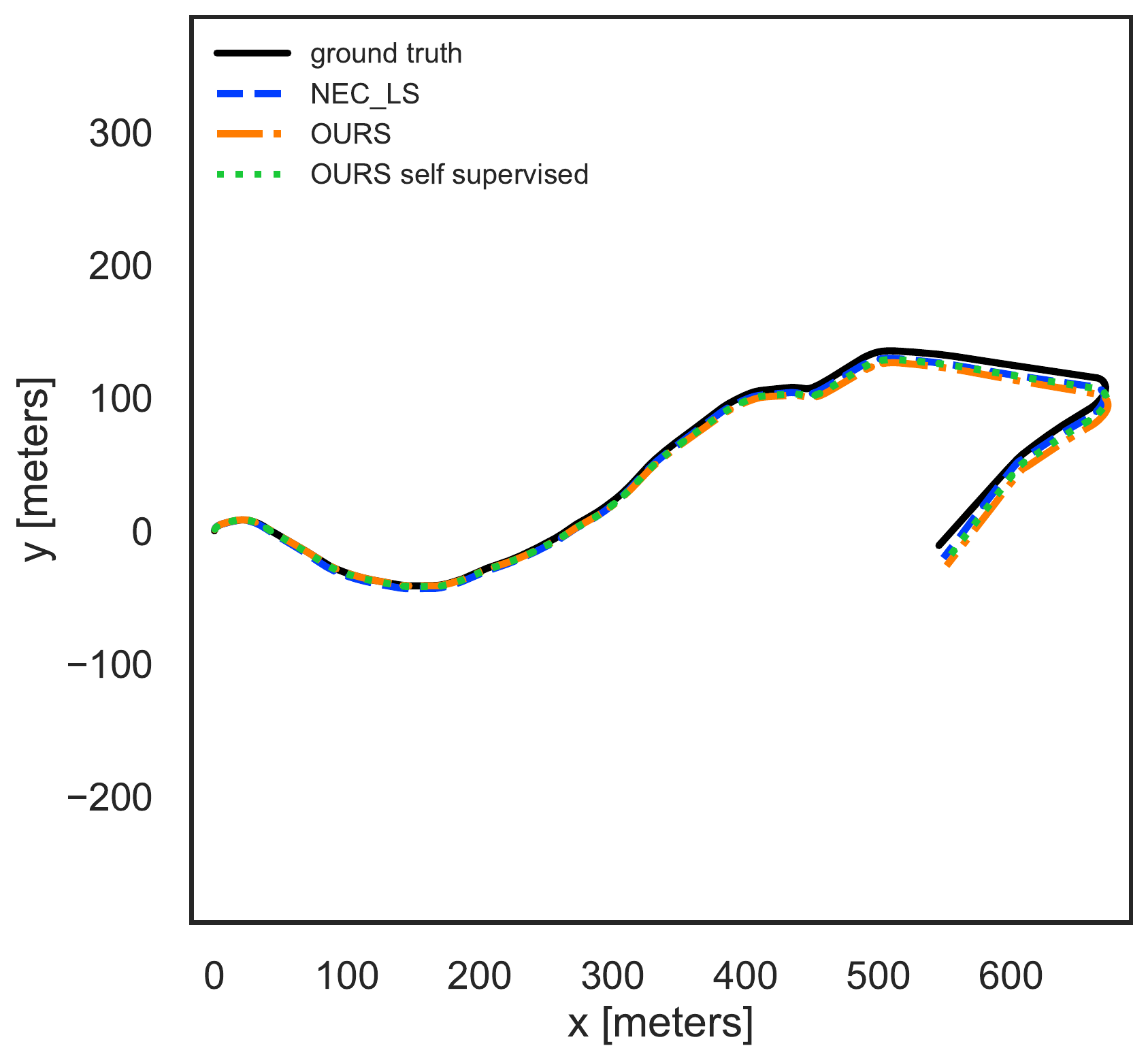}}
        \caption{sequence 10}
        \label{fig:traj_sp_10}
    \end{subfigure}
    \vspace{-0.25cm}
    \caption{
        Trajectory comparison for the KITTI visual odometry sequences for SuperPoint keypoints. Since we compare monocular methods, that cannot estimate the correct scale from a pair of images, we use the scale of the ground truth translations for visualization purposes.
    }
\label{fig:traj_sp}
\vspace{-0.45cm}
\end{figure*}

\begin{figure*}[t]
    \centering
    \vspace{0.0cm}
    \begin{subfigure}[b]{0.3\textwidth}
        \centering
        {\includegraphics[trim={0.25cm 0.25cm 0.25cm 0.2cm},clip,width=\textwidth]{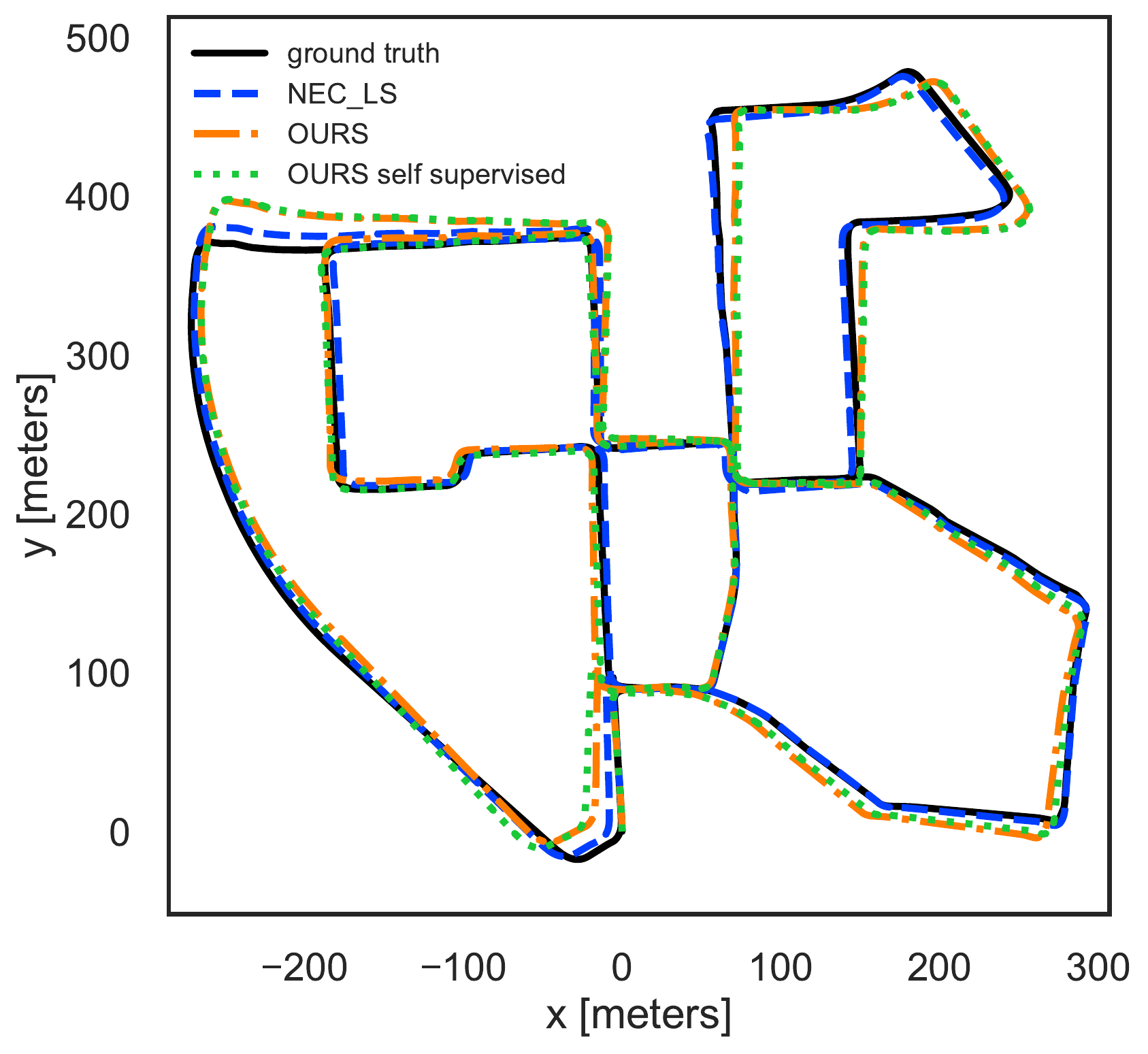}}
        \caption{sequence 00}
        \label{fig:traj_klt_00}
    \end{subfigure}
    \begin{subfigure}[b]{0.3\textwidth}
        \centering
        {\includegraphics[trim={0.25cm 0.25cm 0.25cm 0.2cm},clip,width=\textwidth]{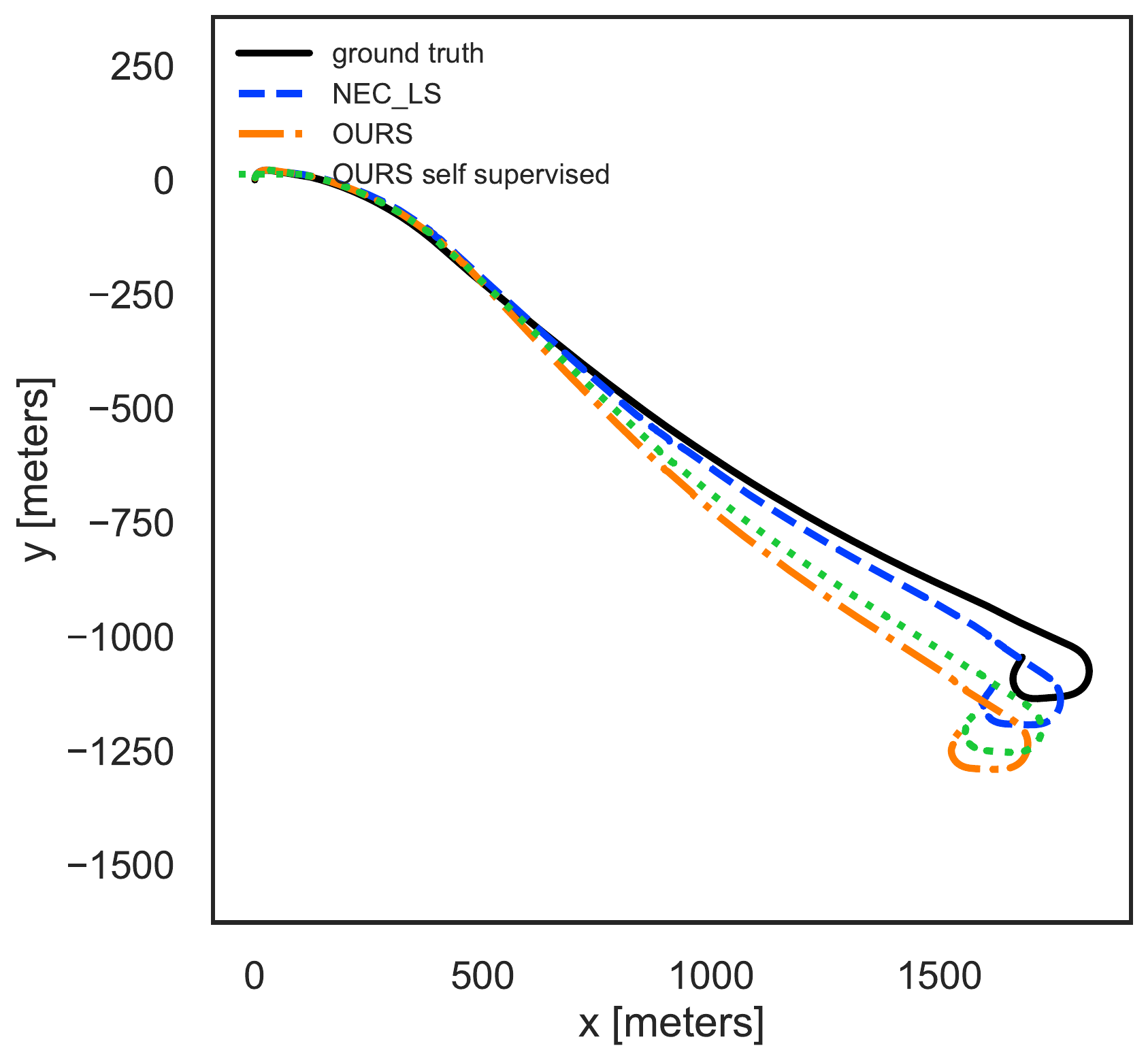}}
        \caption{sequence 01}
        \label{fig:traj_klt_01}
    \end{subfigure}
    \begin{subfigure}[b]{0.3\textwidth}
        \centering
        {\includegraphics[trim={0.25cm 0.25cm 0.25cm 0.2cm},clip,width=\textwidth]{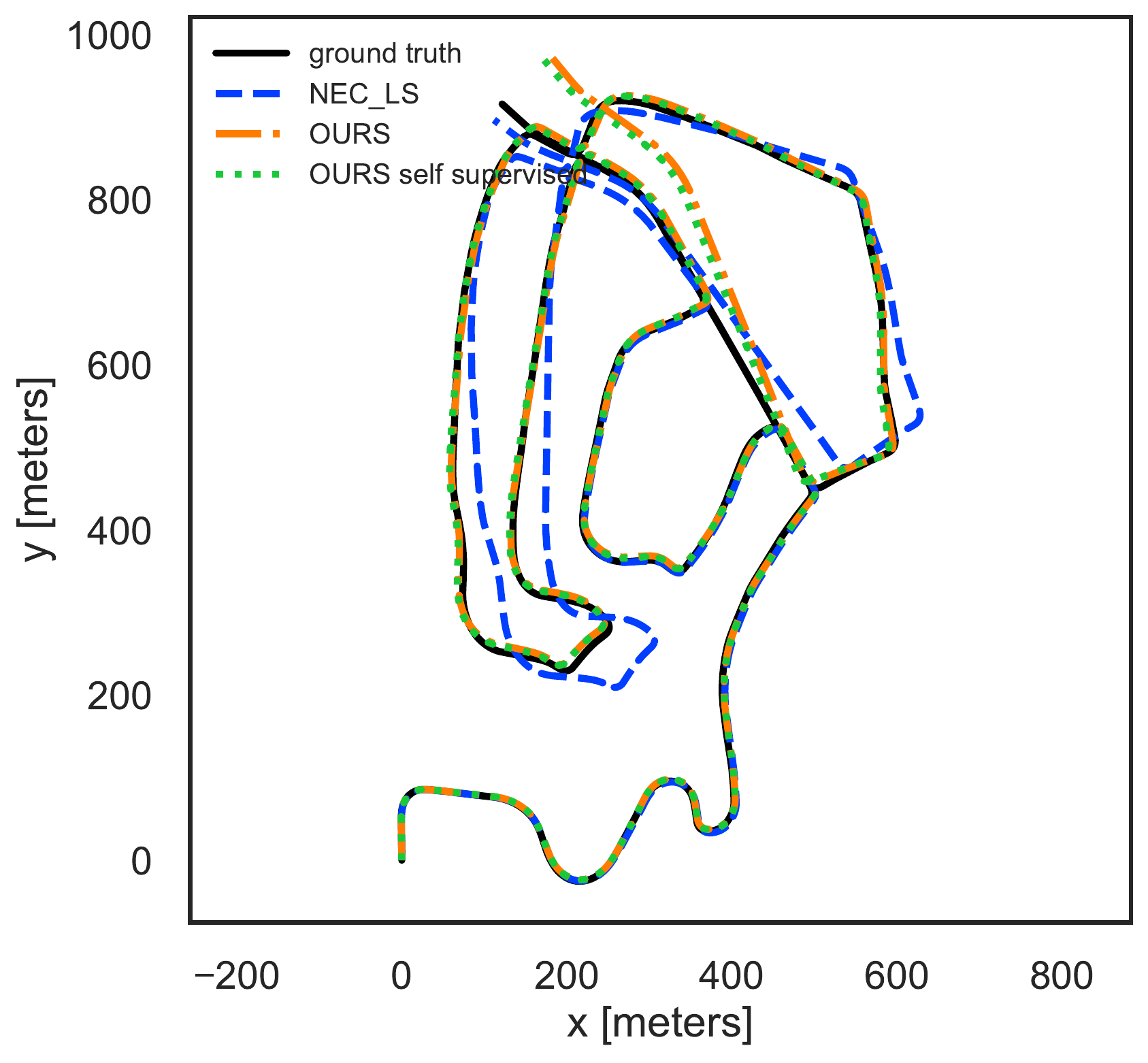}}
        \caption{sequence 02}
        \label{fig:traj_klt_02}
    \end{subfigure}
    \begin{subfigure}[b]{0.3\textwidth}
        \centering
        {\includegraphics[trim={0.25cm 0.25cm 0.25cm 0.2cm},clip,width=\textwidth]{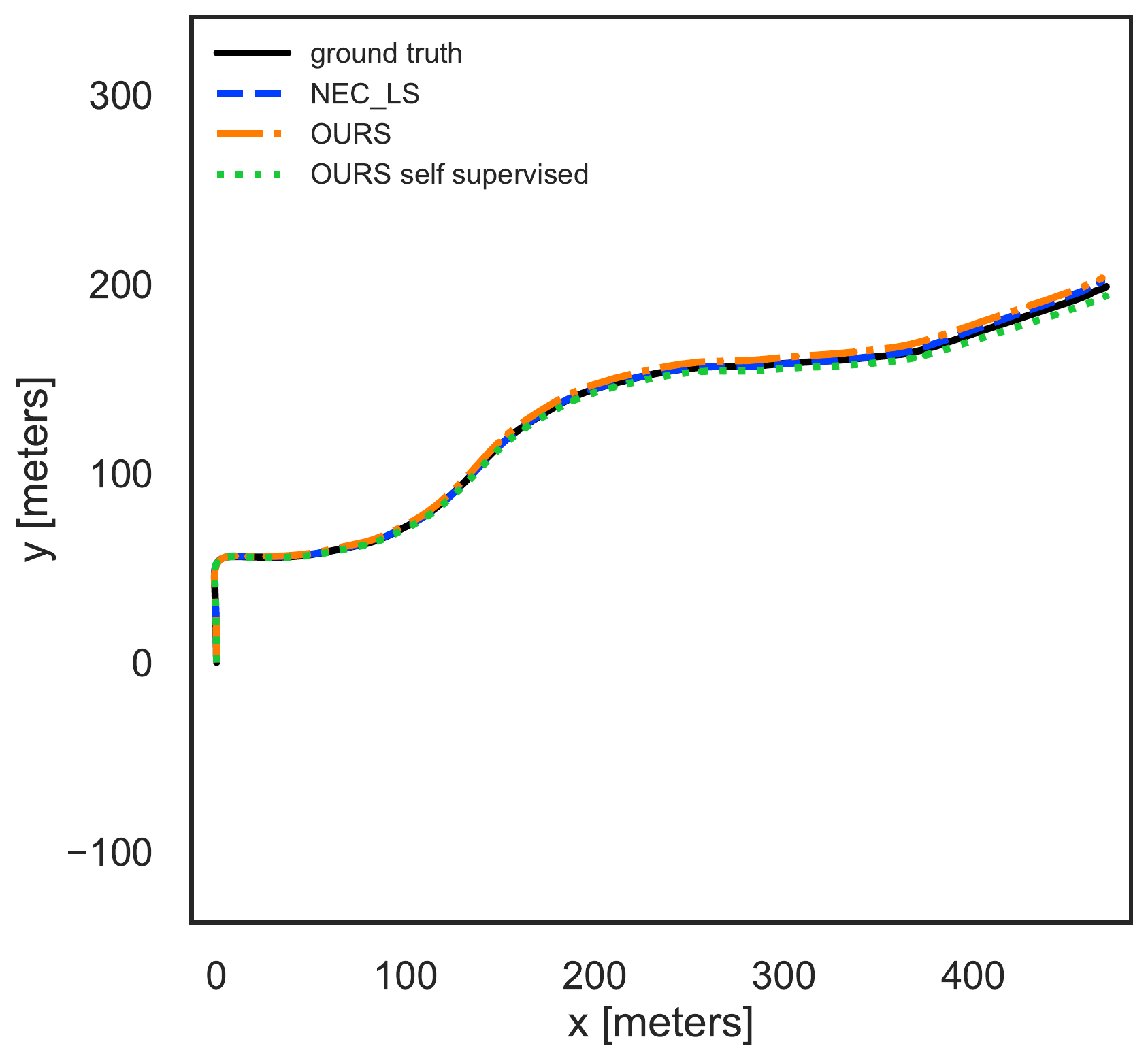}}
        \caption{sequence 03}
        \label{fig:traj_klt_03}
    \end{subfigure}
    \begin{subfigure}[b]{0.3\textwidth}
        \centering
        {\includegraphics[trim={0.25cm 0.25cm 0.25cm 0.2cm},clip,width=\textwidth]{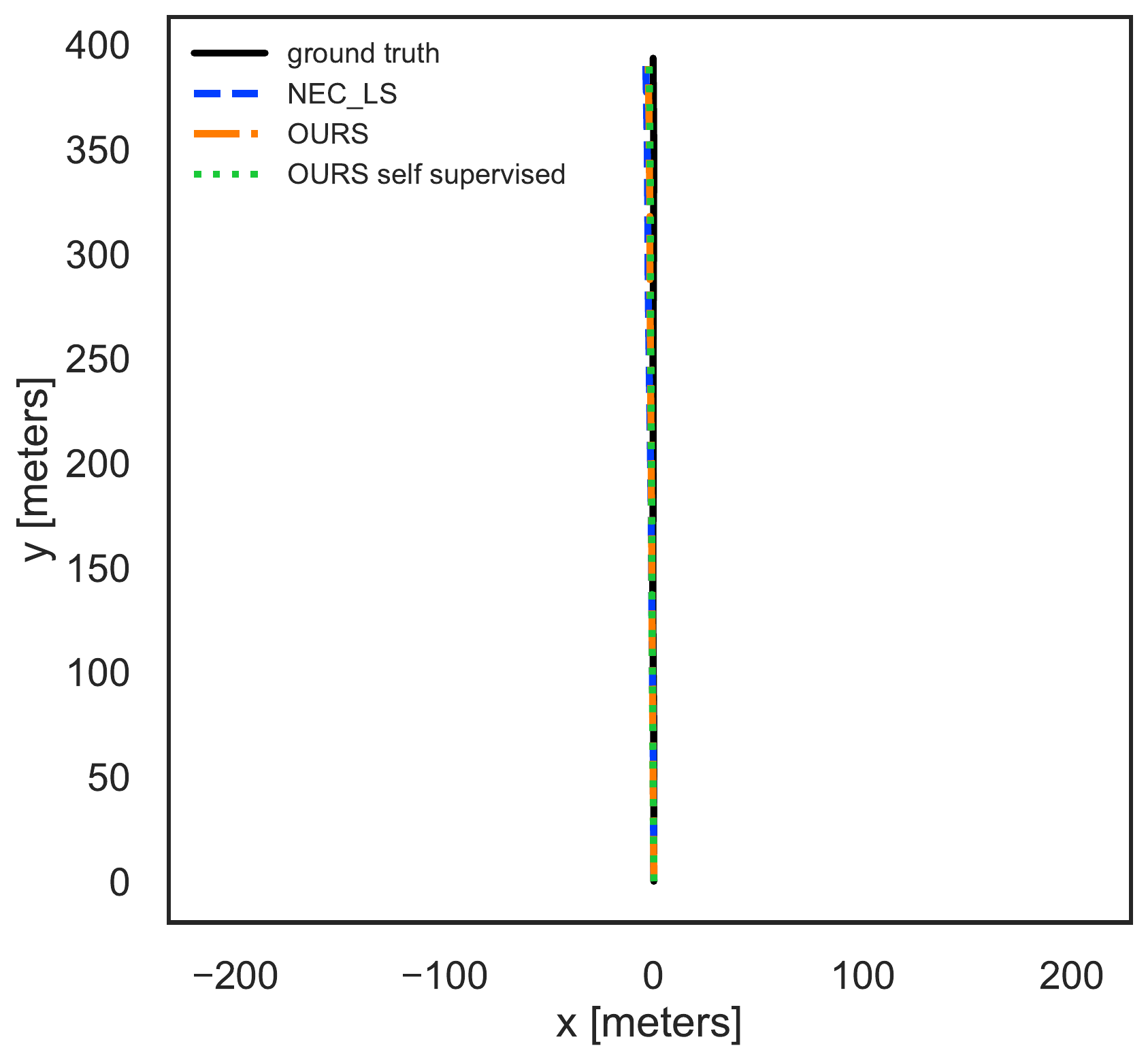}}
        \caption{sequence 04}
        \label{fig:traj_klt_04}
    \end{subfigure}
    \begin{subfigure}[b]{0.3\textwidth}
        \centering
        {\includegraphics[trim={0.25cm 0.25cm 0.25cm 0.2cm},clip,width=\textwidth]{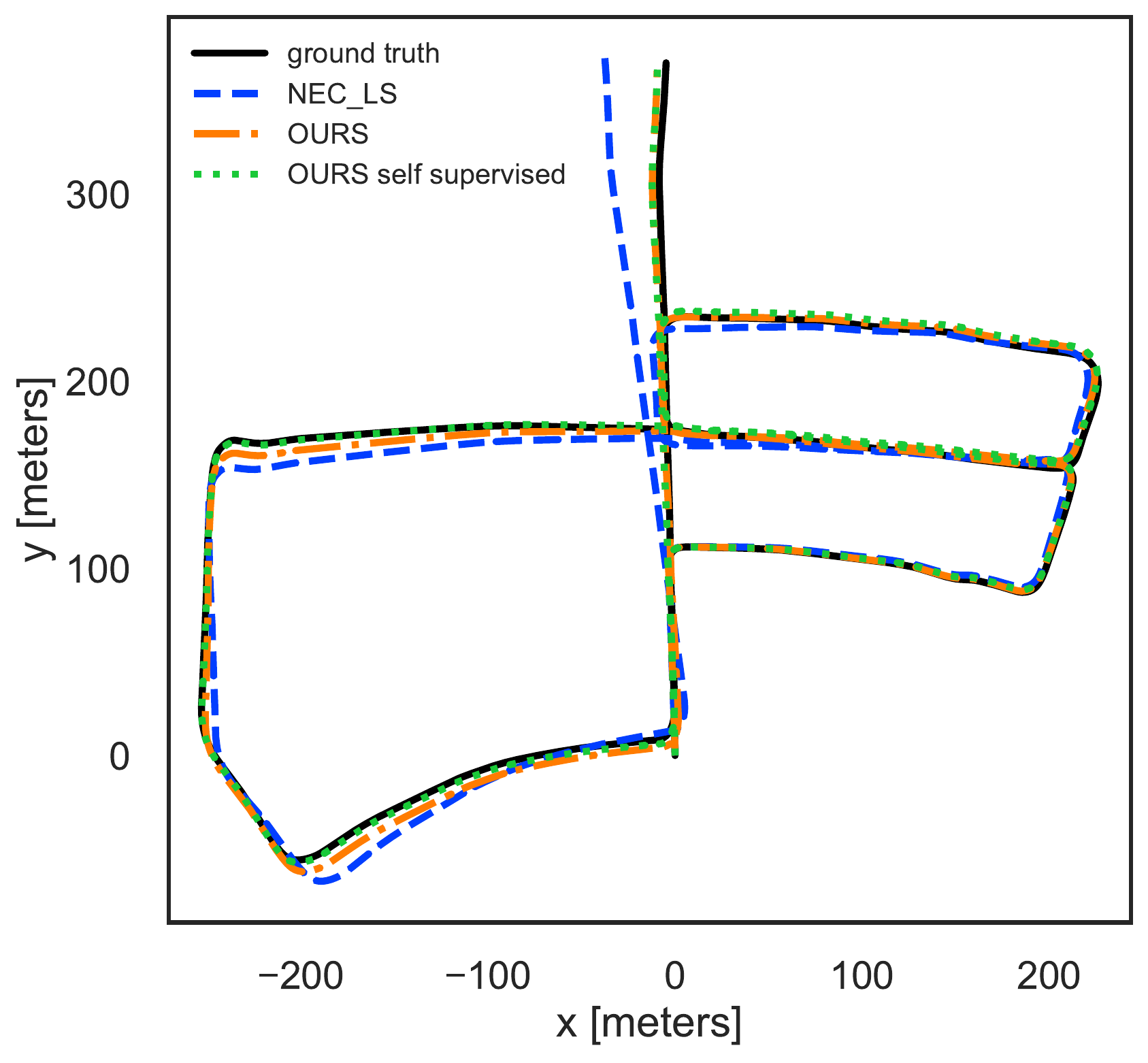}}
        \caption{sequence 05}
        \label{fig:traj_klt_05}
    \end{subfigure}
    \begin{subfigure}[b]{0.3\textwidth}
        \centering
        {\includegraphics[trim={0.25cm 0.25cm 0.25cm 0.2cm},clip,width=\textwidth]{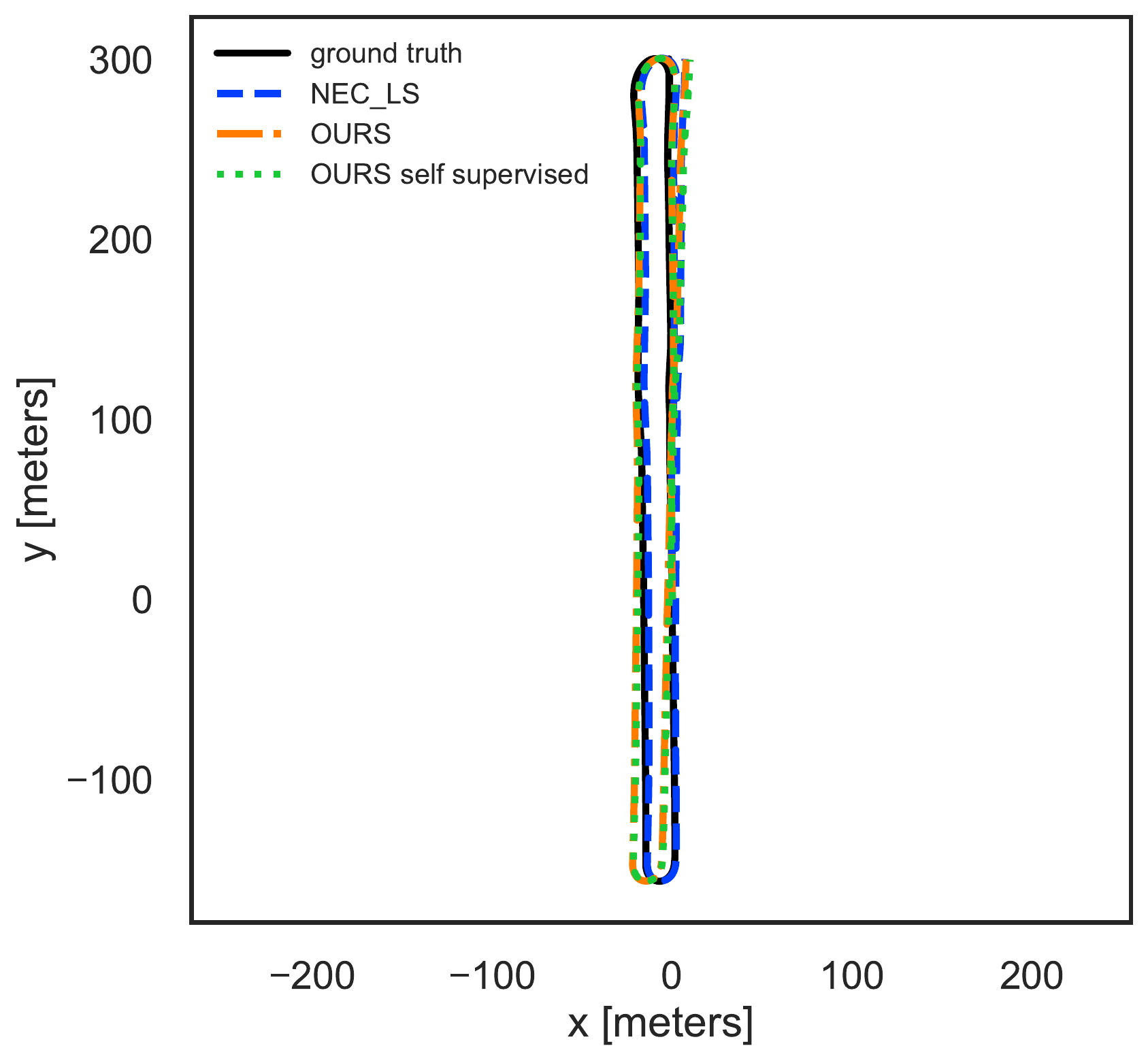}}
        \caption{sequence 06}
        \label{fig:traj_klt_06}
    \end{subfigure}
    \begin{subfigure}[b]{0.3\textwidth}
        \centering
        {\includegraphics[trim={0.25cm 0.25cm 0.25cm 0.2cm},clip,width=\textwidth]{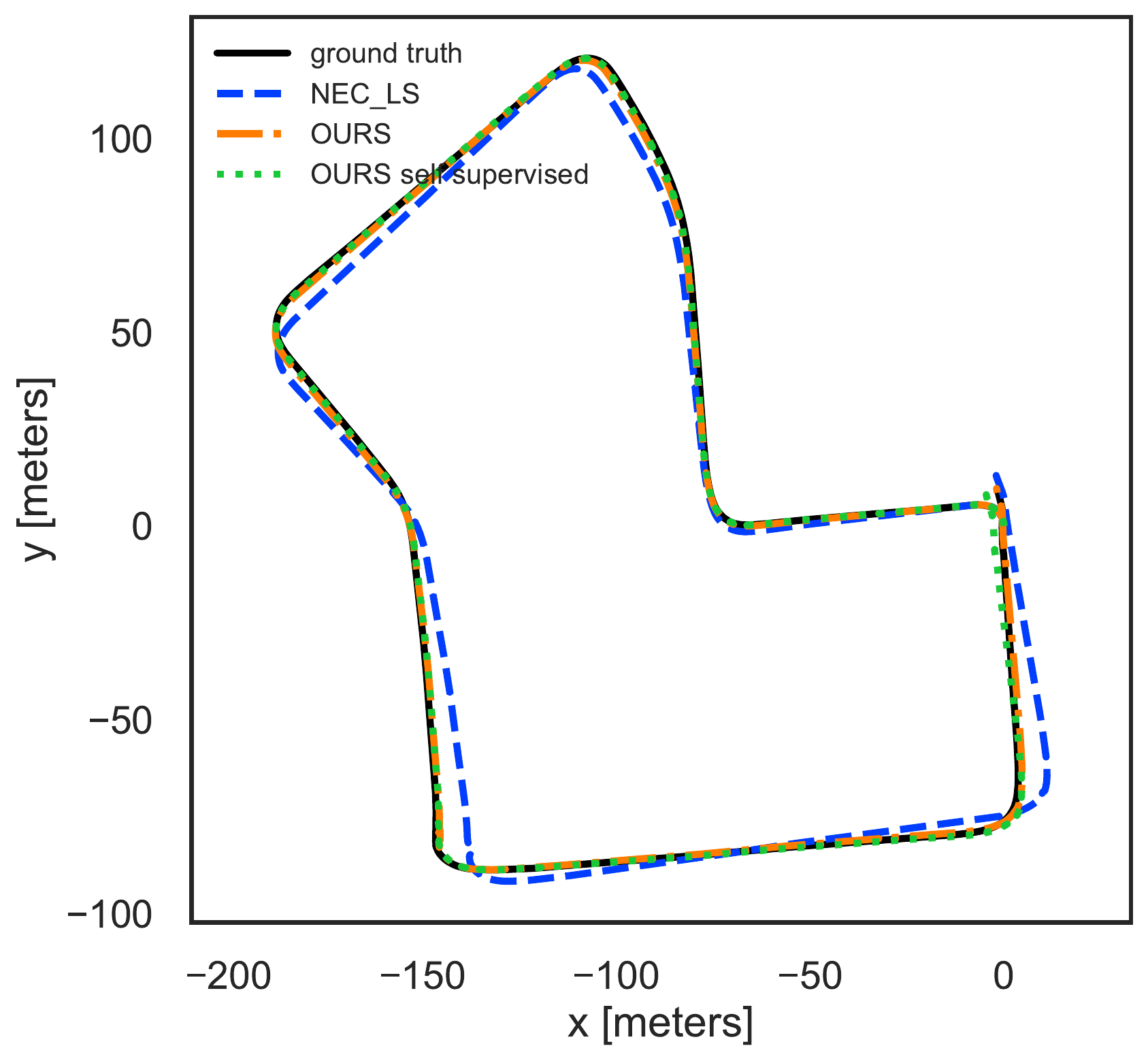}}
        \caption{sequence 07}
        \label{fig:traj_klt_07}
    \end{subfigure}
    \begin{subfigure}[b]{0.3\textwidth}
        \centering
        {\includegraphics[trim={0.25cm 0.25cm 0.25cm 0.2cm},clip,width=\textwidth]{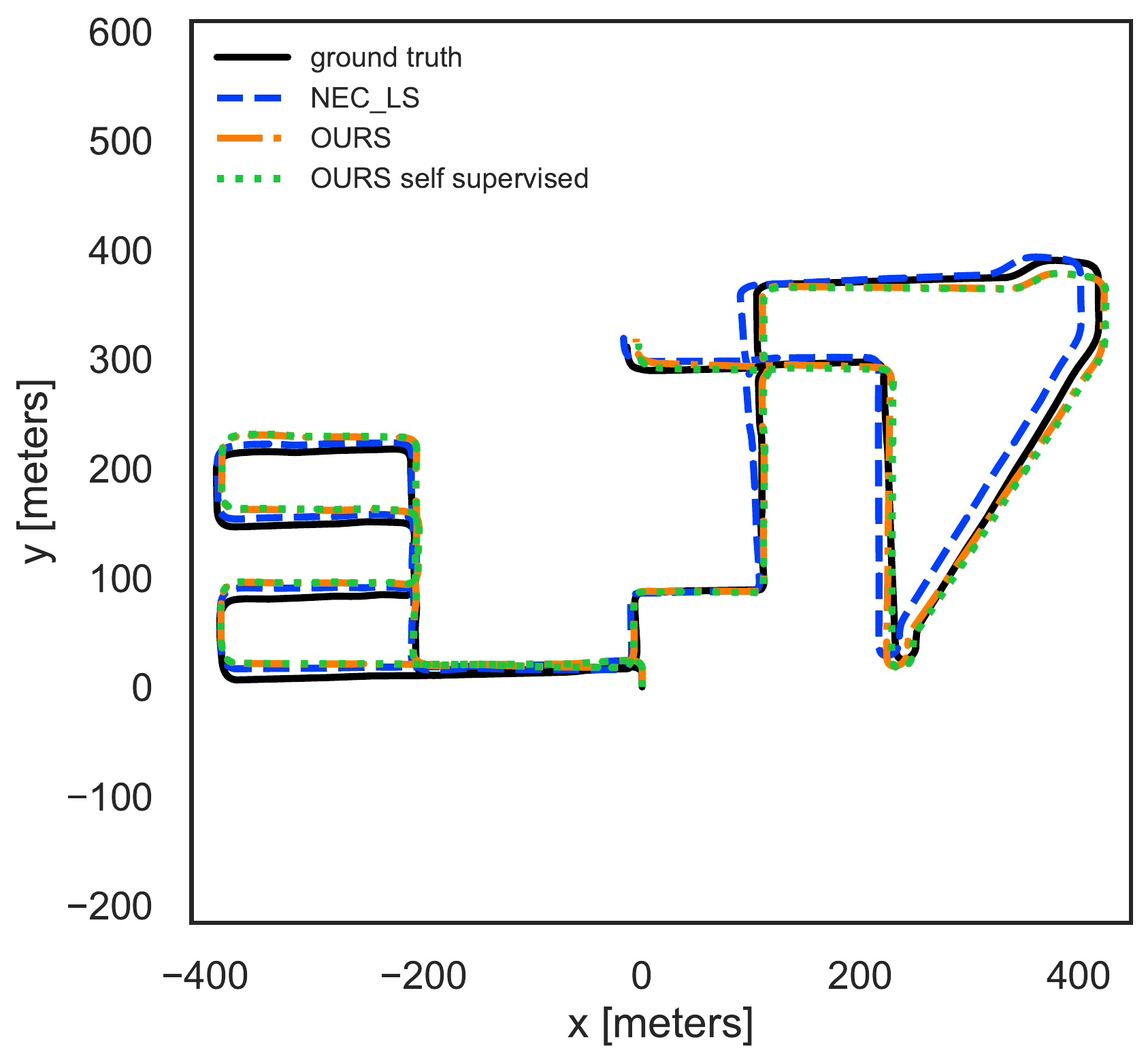}}
        \caption{sequence 08}
        \label{fig:traj_klt_08}
    \end{subfigure}
    \begin{subfigure}[b]{0.3\textwidth}
        \centering
        {\includegraphics[trim={0.25cm 0.25cm 0.25cm 0.2cm},clip,width=\textwidth]{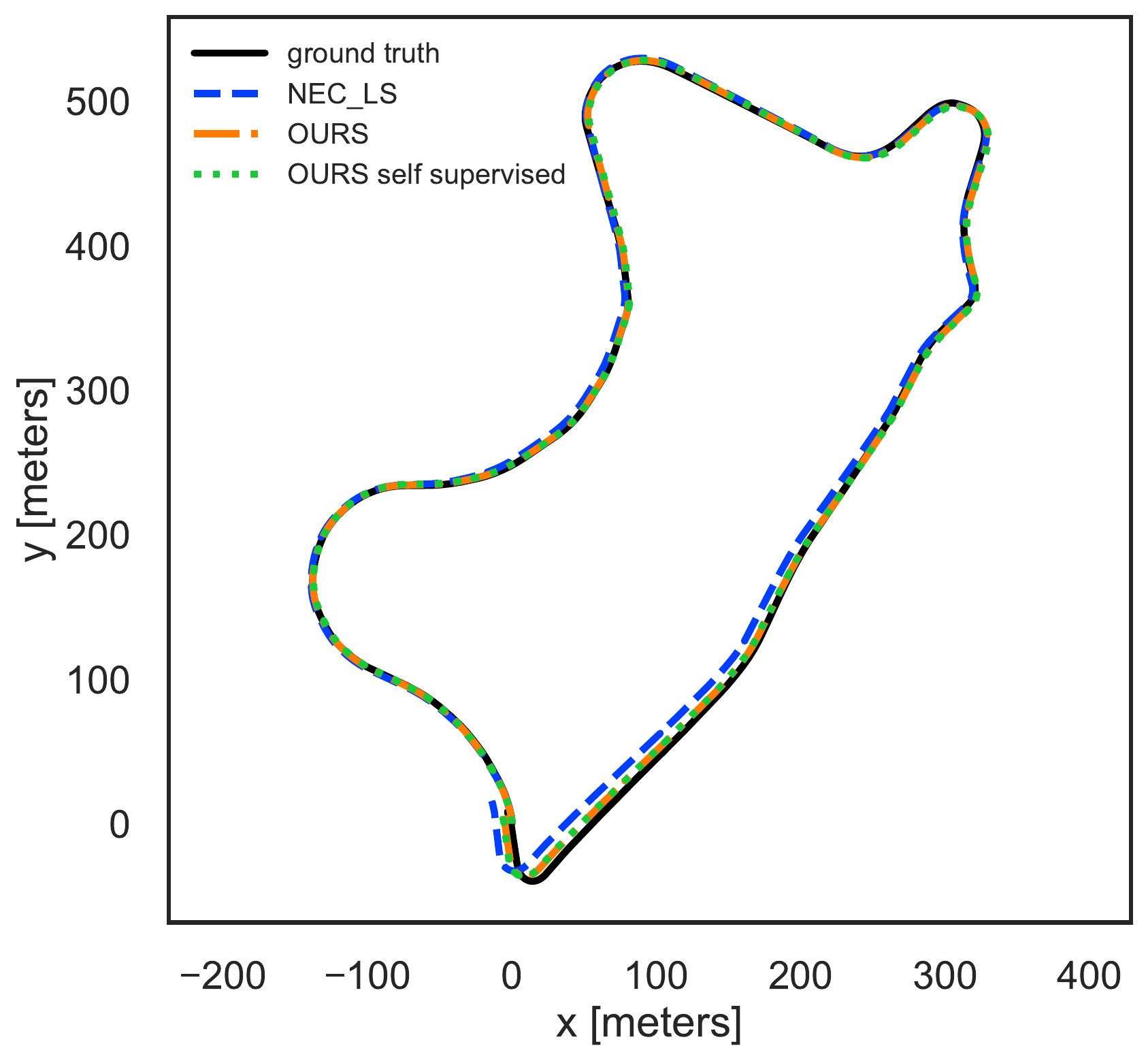}}
        \caption{sequence 09}
        \label{fig:traj_klt_09}
    \end{subfigure}
    \begin{subfigure}[b]{0.3\textwidth}
        \centering
        {\includegraphics[trim={0.25cm 0.25cm 0.25cm 0.2cm},clip,width=\textwidth]{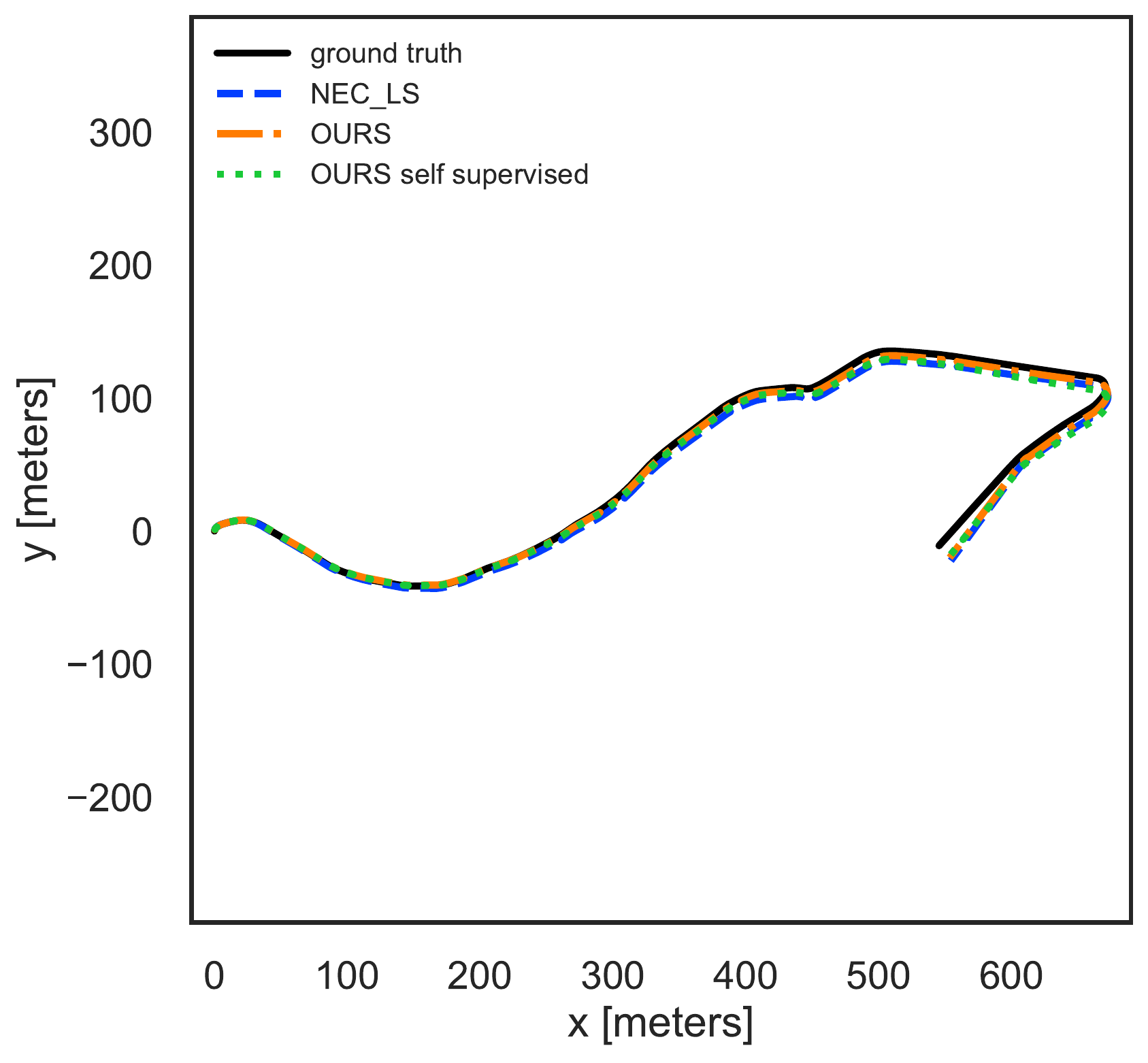}}
        \caption{sequence 10}
        \label{fig:traj_klt_10}
    \end{subfigure}
    \vspace{-0.25cm}
    \caption{
        Trajectory comparison for the KITTI visual odometry sequences for KLT-tracks. Since we compare monocular methods, that cannot estimate the correct scale from a pair of images, we use the scale of the ground truth translations for visualization purposes.
    }
\label{fig:traj_klt}
\vspace{-0.45cm}
\end{figure*}

 We show trajectories for all sequences of the KITTI dataset in \autoref{fig:traj_klt} and \autoref{fig:traj_sp}. Our method consistently achieves the smallest drift over all sequences.
 \newpage


 \begin{table*}[t]
\small
\centering
\sisetup{detect-weight,mode=text}
\renewrobustcmd{\bfseries}{\fontseries{b}\selectfont}
\renewrobustcmd{\boldmath}{}
\newrobustcmd{\B}{\bfseries}
\addtolength{\tabcolsep}{-3.5pt}
\begin{tabular} {p{0.6cm} r r r|r r r|r r r|r r r|r r r|r r r}
\toprule
& \multicolumn{3}{c}{\scshape 8pt \cite{longuet1987readings}}  
& \multicolumn{3}{c}{\scshape NEC \cite{EigenNEC_Kneip2013}} 
& \multicolumn{3}{c}{\scshape NEC-LS}  
& \multicolumn{3}{c}{\scshape OURS} 
& \multicolumn{3}{c}{\scshape OURS SELF-} 
& \multicolumn{3}{c}{\scshape REPROJECTION} 
\\
& \multicolumn{3}{c}{} 
& \multicolumn{3}{c}{} 
& \multicolumn{3}{c}{}
& \multicolumn{3}{c}{\scshape SUPERVISED} 
& \multicolumn{3}{c}{\scshape SUPERVISED} 
& \multicolumn{3}{c}{}  
\\
Seq.  &
  {\scshape $\text{RPE}_1$} & {\scshape $\text{RPE}_n$} & {\scshape $e_t$} &
  {\scshape $\text{RPE}_1$} & {\scshape $\text{RPE}_n$} & {\scshape $e_t$} &
  {\scshape $\text{RPE}_1$} & {\scshape $\text{RPE}_n$} & {\scshape $e_t$} &
  {\scshape $\text{RPE}_1$} & {\scshape $\text{RPE}_n$} & {\scshape $e_t$} &
  {\scshape $\text{RPE}_1$} & {\scshape $\text{RPE}_n$} & {\scshape $e_t$} &
  {\scshape $\text{RPE}_1$} & {\scshape $\text{RPE}_n$} & {\scshape $e_t$} \\
\midrule
00         &            0.185 &         7.203 &       2.61 &     0.153 &     5.505 &   9.32 &              0.121 &  \bfseries 2.403 &            1.42 &  \underline{0.115} &  \underline{2.994} &  \underline{1.31} &    \bfseries 0.113 &              3.110 &    \bfseries 1.30 &              0.117 &              3.080 &            1.29 \\
\rowcolor{Gray}
01         &  \bfseries 0.253 &         7.162 &       2.89 &     0.659 &    28.523 &   5.24 &  \underline{0.270} &            8.991 &  \bfseries 2.20 &              0.294 &  \underline{6.433} &  \underline{2.23} &              0.349 &    \bfseries 6.042 &              2.27 &              0.363 &              7.712 &            2.20 \\
02         &            0.159 &         7.451 &       1.85 &     0.115 &     6.891 &   7.69 &              0.079 &            3.751 &            1.06 &  \underline{0.078} &  \underline{3.411} &  \underline{0.99} &    \bfseries 0.075 &    \bfseries 3.342 &    \bfseries 0.99 &              0.083 &              4.410 &            0.99 \\
\rowcolor{Gray}
03         &            0.131 &         4.822 &       2.47 &     0.089 &     1.889 &   7.45 &  \underline{0.051} &            1.493 &            1.17 &              0.058 &  \underline{0.602} &  \underline{1.01} &    \bfseries 0.049 &    \bfseries 0.444 &    \bfseries 1.00 &              0.047 &              0.608 &            1.00 \\
04         &            0.126 &         1.899 &       1.08 &     0.037 &     0.846 &   6.42 &              0.037 &            0.816 &            0.50 &    \bfseries 0.030 &    \bfseries 0.387 &  \underline{0.44} &  \underline{0.030} &  \underline{0.428} &    \bfseries 0.43 &              0.028 &              0.549 &            0.33 \\
\rowcolor{Gray}
05         &            0.148 &         5.563 &       3.35 &     0.155 &    10.630 &   9.75 &              0.089 &            6.352 &            2.40 &    \bfseries 0.046 &  \underline{1.285} &    \bfseries 2.23 &  \underline{0.046} &    \bfseries 1.235 &  \underline{2.23} &              0.056 &              1.644 &            2.17 \\
06         &            0.142 &         3.376 &       1.55 &     0.066 &     1.984 &   7.30 &              0.044 &  \bfseries 1.325 &            0.63 &  \underline{0.032} &              1.576 &    \bfseries 0.50 &    \bfseries 0.032 &  \underline{1.569} &  \underline{0.50} &              0.031 &              1.467 &            0.45 \\
\rowcolor{Gray}
07         &            0.170 &         5.347 &       6.41 &     0.258 &    12.558 &  12.51 &              0.120 &            5.371 &            5.58 &    \bfseries 0.094 &  \underline{2.731} &    \bfseries 4.97 &  \underline{0.098} &    \bfseries 2.500 &  \underline{5.15} &              0.073 &              2.132 &            4.18 \\
08         &            0.144 &         8.508 &       3.49 &     0.088 &     3.902 &   8.91 &              0.053 &            2.908 &            2.49 &  \underline{0.048} &  \underline{2.373} &    \bfseries 2.36 &    \bfseries 0.047 &    \bfseries 1.706 &  \underline{2.36} &              0.047 &              2.454 &            2.31 \\
\rowcolor{Gray}
09         &            0.151 &         4.546 &       1.71 &     0.054 &     2.027 &   6.76 &              0.052 &            2.307 &            0.74 &  \underline{0.043} &  \underline{1.244} &    \bfseries 0.64 &    \bfseries 0.042 &    \bfseries 1.141 &  \underline{0.64} &              0.044 &              1.385 &            0.64 \\
10         &            0.148 &         6.540 &       2.88 &     0.119 &     8.302 &   8.53 &              0.066 &            4.576 &            1.78 &  \underline{0.058} &  \underline{3.789} &    \bfseries 1.58 &    \bfseries 0.056 &    \bfseries 3.623 &  \underline{1.60} &              0.057 &              2.615 &            1.37 \\
\midrule
\rowcolor{Gray}
train &            0.168 &         6.407 &       2.69 &     0.173 &     8.301 &   8.59 &              0.103 &            3.955 &            1.73 &    \bfseries 0.094 &  \underline{2.782} &    \bfseries 1.60 &  \underline{0.096} &    \bfseries 2.737 &  \underline{1.61} &              0.100 &              3.193 &            1.52 \\
test  &            0.146 &         7.246 &       2.97 &     0.085 &     4.237 &   8.34 &              0.055 &            3.060 &            1.96 &  \underline{0.048} &  \underline{2.359} &    \bfseries 1.82 &    \bfseries 0.048 &    \bfseries 1.910 &  \underline{1.83} &              0.048 &              2.234 &            1.76 \\
\bottomrule
\end{tabular}
\vspace{-0.25cm}
\caption{Quantitative comparison on the KITTI \cite{Geiger2012CVPR} dataset with \ac{klt} tracks \cite{Basalt_Usenko2020}. We replace the Nist\'er-5pt \cite{EM_Nister2003} with the 8pt \cite{longuet1987readings} algorithm to show more results. We also show, an approximation of the true error distance using reprojected points (this is excluded from being \textbf{bold} or \underline{underlined}). While the reprojection approximation achieves the best results on almost all sequences, our methods are often not far behind. This emphasises, that our method is able to effectively learn covariances.
}
\label{tab:kitti_klt_full}
\vspace{-0.5cm}
\end{table*}
 \begin{table*}[t]
\small
\centering
\sisetup{detect-weight,mode=text}
\renewrobustcmd{\bfseries}{\fontseries{b}\selectfont}
\renewrobustcmd{\boldmath}{}
\newrobustcmd{\B}{\bfseries}
\addtolength{\tabcolsep}{-3.5pt}

\begin{tabular} {p{0.6cm} r r r|r r r|r r r|r r r|r r r|r r r}
\toprule
& \multicolumn{3}{c}{\scshape 8pt \cite{longuet1987readings}}  
& \multicolumn{3}{c}{\scshape NEC \cite{EigenNEC_Kneip2013}} 
& \multicolumn{3}{c}{\scshape NEC-LS}  
& \multicolumn{3}{c}{\scshape OURS} 
& \multicolumn{3}{c}{\scshape OURS SELF-} 
& \multicolumn{3}{c}{\scshape REPROJECTION} 
\\
& \multicolumn{3}{c}{} 
& \multicolumn{3}{c}{} 
& \multicolumn{3}{c}{}
& \multicolumn{3}{c}{\scshape SUPERVISED} 
& \multicolumn{3}{c}{\scshape SUPERVISED} 
& \multicolumn{3}{c}{}  
\\
Seq.  &
  {\scshape $\text{RPE}_1$} & {\scshape $\text{RPE}_n$} & {\scshape $e_t$} &
  {\scshape $\text{RPE}_1$} & {\scshape $\text{RPE}_n$} & {\scshape $e_t$} &
  {\scshape $\text{RPE}_1$} & {\scshape $\text{RPE}_n$} & {\scshape $e_t$} &
  {\scshape $\text{RPE}_1$} & {\scshape $\text{RPE}_n$} & {\scshape $e_t$} &
  {\scshape $\text{RPE}_1$} & {\scshape $\text{RPE}_n$} & {\scshape $e_t$} &
  {\scshape $\text{RPE}_1$} & {\scshape $\text{RPE}_n$} & {\scshape $e_t$} \\
\midrule
00         &         0.216 &        11.650 &       3.40 &     0.132 &             12.483 &   3.20 &              0.116 &            8.728 &    \bfseries 1.35 &  \underline{0.114} &    \bfseries 2.277 &  \underline{1.38} &    \bfseries 0.114 &  \underline{2.522} &              1.38 &              0.113 &              2.363 &            1.28 \\
\rowcolor{Gray}
01         &         0.246 &         8.080 &       3.83 &     0.539 &             22.857 &   1.55 &              0.082 &            6.378 &              1.00 &  \underline{0.060} &  \underline{5.811} &  \underline{0.99} &    \bfseries 0.057 &    \bfseries 5.770 &    \bfseries 0.94 &              0.054 &              5.997 &            0.81 \\
02         &         0.188 &        12.003 &       2.06 &     0.093 &              7.594 &   1.76 &              0.069 &            4.050 &              1.01 &  \underline{0.066} &    \bfseries 2.224 &    \bfseries 0.99 &    \bfseries 0.066 &  \underline{2.237} &  \underline{1.00} &              0.065 &              2.679 &            0.95 \\
\rowcolor{Gray}
03         &         0.167 &         8.308 &       3.42 &     0.090 &              3.863 &   3.31 &    \bfseries 0.055 &            3.754 &  \underline{1.12} &              0.059 &  \underline{2.239} &              1.13 &  \underline{0.057} &    \bfseries 2.051 &    \bfseries 1.12 &              0.054 &              2.394 &            1.07 \\
04         &         0.160 &         2.682 &       1.45 &     0.040 &  \underline{0.486} &   0.81 &              0.041 &  \bfseries 0.434 &              0.49 &  \underline{0.038} &              1.041 &    \bfseries 0.46 &    \bfseries 0.037 &              0.808 &  \underline{0.46} &              0.027 &              0.526 &            0.30 \\
\rowcolor{Gray}
05         &         0.198 &         9.236 &       4.56 &     0.119 &             11.779 &   3.65 &              0.062 &           12.437 &              2.50 &  \underline{0.055} &    \bfseries 1.931 &    \bfseries 2.37 &    \bfseries 0.055 &  \underline{1.949} &  \underline{2.40} &              0.053 &              2.123 &            2.02 \\
06         &         0.193 &         5.244 &       2.89 &     0.059 &              6.901 &   1.43 &              0.050 &            6.634 &              0.76 &  \underline{0.042} &    \bfseries 1.178 &  \underline{0.70} &    \bfseries 0.041 &  \underline{1.242} &    \bfseries 0.70 &              0.035 &              0.964 &            0.58 \\
\rowcolor{Gray}
07         &         0.231 &         7.086 &       8.86 &     0.185 &              4.402 &   8.67 &              0.112 &  \bfseries 2.341 &              6.69 &    \bfseries 0.103 &  \underline{2.772} &    \bfseries 6.54 &  \underline{0.109} &              3.715 &  \underline{6.63} &              0.120 &              3.434 &            4.82 \\
08         &         0.183 &        10.423 &       4.21 &     0.081 &              8.284 &   3.66 &              0.056 &            7.004 &              2.50 &    \bfseries 0.050 &    \bfseries 4.067 &  \underline{2.46} &  \underline{0.050} &  \underline{4.118} &    \bfseries 2.46 &              0.048 &              3.623 &            2.30 \\
\rowcolor{Gray}
09         &         0.185 &         5.485 &       2.29 &     0.053 &              1.646 &   1.43 &              0.052 &            1.553 &              0.71 &  \underline{0.049} &  \underline{1.317} &  \underline{0.71} &    \bfseries 0.049 &    \bfseries 1.278 &    \bfseries 0.70 &              0.048 &              1.160 &            0.69 \\
10         &         0.198 &         8.960 &       4.09 &     0.167 &              9.264 &   4.43 &  \underline{0.064} &            4.787 &              1.79 &    \bfseries 0.063 &    \bfseries 3.513 &    \bfseries 1.64 &              0.065 &  \underline{3.821} &  \underline{1.65} &              0.060 &              2.404 &            1.21 \\
\midrule
\rowcolor{Gray}
train &         0.203 &        10.051 &       3.54 &     0.141 &             10.127 &   2.97 &              0.082 &            6.910 &              1.72 &  \underline{0.077} &    \bfseries 2.378 &    \bfseries 1.69 &    \bfseries 0.077 &  \underline{2.505} &  \underline{1.69} &              0.076 &              2.606 &            1.44 \\
test  &         0.186 &         9.023 &       3.74 &     0.089 &              6.917 &   3.28 &              0.056 &            5.353 &              1.96 &    \bfseries 0.052 &    \bfseries 3.333 &  \underline{1.91} &  \underline{0.053} &  \underline{3.408} &    \bfseries 1.91 &              0.050 &              2.839 &            1.73 \\
\bottomrule
\end{tabular}
\vspace{-0.25cm}
\caption{Full results on the KITTI \cite{Geiger2012CVPR} dataset with SuperPoint \cite{detone2018superpoint} keypoints. We replace the Nist\'er-5pt \cite{EM_Nister2003} with the 8pt \cite{longuet1987readings} algorithm to show more results. We also show, an approximation of the true error distance using reprojected points (this is excluded from being \textbf{bold} or \underline{underlined}). While the reprojection approximation achieves the best results on almost all sequences, our methods are often not far behind. This emphasises, that our method is able to effectively learn covariances.}
\label{tab:kitti_superpoint_full}
\vspace{-0.35cm}
\end{table*}


\clearpage

{\small
\bibliographystyle{ieee_fullname}
\bibliography{egbib}

\begin{thebibliography}{10}\itemsep=-1pt

\bibitem{ceres-solver}
Sameer Agarwal, Keir Mierle, and Others.
\newblock Ceres solver.
\newblock \url{http://ceres-solver.org}.

\bibitem{arandjelovic2016netvlad}
Relja Arandjelovic, Petr Gronat, Akihiko Torii, Tomas Pajdla, and Josef Sivic.
\newblock Netvlad: Cnn architecture for weakly supervised place recognition.
\newblock In {\em CVPR}, 2016.

\bibitem{bay2006surf}
Herbert Bay, Tinne Tuytelaars, and Luc Van~Gool.
\newblock Surf: Speeded up robust features.
\newblock In {\em ECCV}, 2006.

\bibitem{DBLP:journals/jei/BishopN07}
Christopher~M. Bishop.
\newblock {\em Pattern recognition and machine learning, 5th Edition}.
\newblock Springer, 2007.

\bibitem{brachmann2017dsac}
Eric Brachmann, Alexander Krull, Sebastian Nowozin, Jamie Shotton, Frank
  Michel, Stefan Gumhold, and Carsten Rother.
\newblock Dsac-differentiable ransac for camera localization.
\newblock In {\em CVPR}, 2017.

\bibitem{Briales2018globalNEC}
Jesus Briales, Laurent Kneip, and Javier Gonzalez-Jimenez.
\newblock A certifiably globally optimal solution to the non-minimal relative
  pose problem.
\newblock In {\em CVPR}, 2018.

\bibitem{pro_cov_Brooks2001}
M.J. Brooks, W. Chojnacki, D. Gawley, and A. van~den Hengel.
\newblock What value covariance information in estimating vision parameters?
\newblock In {\em ICCV}, 2001.

\bibitem{burnett2021radar}
Keenan Burnett, David~J Yoon, Angela~P Schoellig, and Timothy~D Barfoot.
\newblock Radar odometry combining probabilistic estimation and unsupervised
  feature learning.
\newblock {\em Robotics: Science and Systems}, 2021.

\bibitem{Burri2016euroc}
Michael Burri, Janosch Nikolic, Pascal Gohl, Thomas Schneider, Joern Rehder,
  Sammy Omari, Markus~W Achtelik, and Roland Siegwart.
\newblock The euroc micro aerial vehicle datasets.
\newblock {\em The International Journal of Robotics Research}, 2016.

\bibitem{cadena2016past}
Cesar Cadena, Luca Carlone, Henry Carrillo, Yasir Latif, Davide Scaramuzza,
  Jos{\'e} Neira, Ian Reid, and John~J Leonard.
\newblock Past, present, and future of simultaneous localization and mapping:
  Toward the robust-perception age.
\newblock {\em IEEE Transactions on robotics}, 32, 2016.

\bibitem{MRO_Chng2020}
Chee-Kheng Chng, {\'A}lvaro Parra, Tat-Jun Chin, and Yasir Latif.
\newblock Monocular rotational odometry with incremental rotation averaging and
  loop closure.
\newblock {\em Digital Image Computing: Techniques and Applications (DICTA)},
  2020.

\bibitem{detone2018superpoint}
Daniel DeTone, Tomasz Malisiewicz, and Andrew Rabinovich.
\newblock Superpoint: Self-supervised interest point detection and description.
\newblock In {\em Proceedings of the IEEE conference on computer vision and
  pattern recognition workshops}, 2018.

\bibitem{domke2012generic}
Justin Domke.
\newblock Generic methods for optimization-based modeling.
\newblock In {\em Artificial Intelligence and Statistics}. PMLR, 2012.

\bibitem{dorini2011unscented}
Leyza~Baldo Dorini and Siome~Klein Goldenstein.
\newblock Unscented feature tracking.
\newblock {\em Computer Vision and Image Understanding}, 115, 2011.

\bibitem{DSO_Engel2016}
J. Engel, V. Koltun, and D. Cremers.
\newblock Direct sparse odometry.
\newblock {\em IEEE Transactions on Pattern Analysis and Machine Intelligence},
  2018.

\bibitem{engel14eccv}
J. Engel, T. Schöps, and D. Cremers.
\newblock {LSD-SLAM}: Large-scale direct monocular {SLAM}.
\newblock In {\em ECCV}, 2014.

\bibitem{fathian2018quest}
Kaveh Fathian, J~Pablo Ramirez-Paredes, Emily~A Doucette, J~Willard Curtis, and
  Nicholas~R Gans.
\newblock Quest: A quaternion-based approach for camera motion estimation from
  minimal feature points.
\newblock {\em {{IEEE} Robotics and Automation Letters ({RAL})}}, 3, 2018.

\bibitem{Faugeras1989mul_solutions}
O.D. Faugeras and S. Maybank.
\newblock Motion from point matches: multiplicity of solutions.
\newblock In {\em Workshop on Visual Motion}, 1989.

\bibitem{fischler1981random}
Martin~A Fischler and Robert~C Bolles.
\newblock Random sample consensus: a paradigm for model fitting with
  applications to image analysis and automated cartography.
\newblock {\em Communications of the ACM}, 24, 1981.

\bibitem{forstner1987fast}
Wolfgang F{\"o}rstner and Eberhard G{\"u}lch.
\newblock A fast operator for detection and precise location of distinct
  points, corners and centres of circular features.
\newblock In {\em ISPRS intercommission conference on fast processing of
  photogrammetric data}, 1987.

\bibitem{Geiger2012CVPR}
Andreas Geiger, Philip Lenz, and Raquel Urtasun.
\newblock {Are we ready for Autonomous Driving? The KITTI Vision Benchmark
  Suite}.
\newblock In {\em CVPR}, 2012.

\bibitem{germain2020s2dnet}
Hugo Germain, Guillaume Bourmaud, and Vincent Lepetit.
\newblock S2dnet: Learning accurate correspondences for sparse-to-dense feature
  matching.
\newblock {\em arXiv preprint arXiv:2004.01673}, 2020.

\bibitem{hartley1997defense}
Richard~I Hartley.
\newblock In defense of the eight-point algorithm.
\newblock {\em IEEE Transactions on pattern analysis and machine intelligence},
  19, 1997.

\bibitem{Hartley2004}
R.~I. Hartley and A. Zisserman.
\newblock {\em Multiple View Geometry in Computer Vision}.
\newblock Cambridge University Press, second edition, 2004.

\bibitem{ctcnet}
Ganesh Iyer, {Krishna Murthy} Jatavallabhula, Gunshi Gupta, {Madhava Krishna}
  K, and Liam Paull.
\newblock Geometric consistency for self-supervised end-to-end visual odometry.
\newblock In {\em CVPR Workshops}, 2018.

\bibitem{jatavallabhula2020slam}
Krishna~Murthy Jatavallabhula, Ganesh Iyer, and Liam Paull.
\newblock {$\nabla$} slam: Dense slam meets automatic differentiation.
\newblock In {\em {{IEEE} International Conference on Robotics and Automation
  ({ICRA})}}, 2020.

\bibitem{kanatani2004geometric}
Kenichi Kanatani.
\newblock For geometric inference from images, what kind of statistical model
  is necessary?
\newblock {\em Systems and Computers in Japan}, 35, 2004.

\bibitem{kanatani2008statistical}
Kenichi Kanatani.
\newblock Statistical optimization for geometric fitting: Theoretical accuracy
  bound and high order error analysis.
\newblock {\em IJCV}, 80, 2008.

\bibitem{con_cov_Kanazawa2001}
Y. Kanazawa and K. Kanatani.
\newblock Do we really have to consider covariance matrices for image features?
\newblock In {\em ICCV}, 2001.

\bibitem{kendall2015posenet}
Alex Kendall, Matthew Grimes, and Roberto Cipolla.
\newblock Posenet: A convolutional network for real-time 6-dof camera
  relocalization.
\newblock In {\em ICCV}, 2015.

\bibitem{Kingma2015AdamAM}
Diederik~P. Kingma and Jimmy Ba.
\newblock Adam: A method for stochastic optimization.
\newblock {\em CoRR}, 2015.

\bibitem{Kneip2014opengv}
Laurent Kneip and Paul Furgale.
\newblock Opengv: A unified and generalized approach to real-time calibrated
  geometric vision.
\newblock In {\em {{IEEE} International Conference on Robotics and Automation
  ({ICRA})}}, 2014.

\bibitem{EigenNEC_Kneip2013}
Laurent Kneip and Simon Lynen.
\newblock Direct optimization of frame-to-frame rotation.
\newblock In {\em ICCV}, 2013.

\bibitem{OrigNEC_Kneip2012}
Laurent Kneip, Roland Siegwart, and Marc Pollefeys.
\newblock Finding the exact rotation between two images independently of the
  translation.
\newblock In {\em ECCV}, 2012.

\bibitem{kruppa1913ermittlung}
Erwin Kruppa.
\newblock {\em Zur Ermittlung eines Objektes aus zwei Perspektiven mit innerer
  Orientierung}.
\newblock H{\"o}lder, 1913.

\bibitem{kukelova2008polynomial}
Zuzana Kukelova, Martin Bujnak, and Tomas Pajdla.
\newblock Polynomial eigenvalue solutions to the 5-pt and 6-pt relative pose
  problems.
\newblock In {\em BMVC}, 2008.

\bibitem{levenberg1944method}
Kenneth Levenberg.
\newblock A method for the solution of certain non-linear problems in least
  squares.
\newblock {\em Quarterly of applied mathematics}, 2, 1944.

\bibitem{li2006five}
Hongdong Li and Richard Hartley.
\newblock Five-point motion estimation made easy.
\newblock In {\em {IEEE} International Conference on Pattern Recognition
  ({ICPR})}, 2006.

\bibitem{lim2010estimating}
John Lim, Nick Barnes, and Hongdong Li.
\newblock Estimating relative camera motion from the antipodal-epipolar
  constraint.
\newblock {\em IEEE TPAMI}, 32, 2010.

\bibitem{lindenberger2021pixsfm}
Philipp Lindenberger, Paul-Edouard Sarlin, Viktor Larsson, and Marc Pollefeys.
\newblock {Pixel-Perfect Structure-from-Motion with Featuremetric Refinement}.
\newblock In {\em ICCV}, 2021.

\bibitem{long2015fully}
Jonathan Long, Evan Shelhamer, and Trevor Darrell.
\newblock Fully convolutional networks for semantic segmentation.
\newblock In {\em CVPR}, 2015.

\bibitem{longuet1987readings}
HC Longuet-Higgins.
\newblock Readings in computer vision: issues, problems, principles, and
  paradigms.
\newblock {\em A computer algorithm for reconstructing a scene from two
  projections}, 1987.

\bibitem{lowe2004distinctive}
David~G Lowe.
\newblock Distinctive image features from scale-invariant keypoints.
\newblock {\em IJCV}, 60, 2004.

\bibitem{KLT_Lukas1981}
Bruce~D. Lucas and Takeo Kanade.
\newblock An iterative image registration technique with an application to
  stereo vision.
\newblock In {\em International Joint Conference on Artificial Intelligence
  ({IJCAI})}, 1981.

\bibitem{marquardt1963algorithm}
Donald~W Marquardt.
\newblock An algorithm for least-squares estimation of nonlinear parameters.
\newblock {\em Journal of the society for Industrial and Applied Mathematics},
  11, 1963.

\bibitem{meidow2009reasoning}
Jochen Meidow, Christian Beder, and Wolfgang F{\"o}rstner.
\newblock Reasoning with uncertain points, straight lines, and straight line
  segments in 2d.
\newblock {\em ISPRS Journal of Photogrammetry and Remote Sensing}, 64, 2009.

\bibitem{mildenhall2021nerf}
Ben Mildenhall, Pratul~P Srinivasan, Matthew Tancik, Jonathan~T Barron, Ravi
  Ramamoorthi, and Ren Ng.
\newblock Nerf: Representing scenes as neural radiance fields for view
  synthesis.
\newblock {\em Communications of the ACM}, 2021.

\bibitem{muhle2022pnec}
D Muhle, L Koestler, N Demmel, F Bernard, and D Cremers.
\newblock The probabilistic normal epipolar constraint for frame-to-frame
  rotation optimization under uncertain feature positions.
\newblock 2022.

\bibitem{ORB_SLAM2_Mur-Artal2017}
R. {Mur-Artal} and J.~D. {Tardós}.
\newblock Orb-slam2: An open-source slam system for monocular, stereo, and
  rgb-d cameras.
\newblock {\em IEEE Transactions on Robotics}, 33, 2017.

\bibitem{EM_Nister2003}
D. {Nister}.
\newblock An efficient solution to the five-point relative pose problem.
\newblock In {\em CVPR}, 2003.

\bibitem{Nistr2004VisualO}
D. Nistr, O. Naroditsky, and J. Bergen.
\newblock Visual odometry.
\newblock {\em CVPR}, 2004.

\bibitem{pineda2022theseus}
Luis Pineda, Taosha Fan, Maurizio Monge, Shobha Venkataraman, Paloma Sodhi,
  Ricky~TQ Chen, Joseph Ortiz, Daniel DeTone, Austin Wang, Stuart Anderson,
  Jing Dong, Brandon Amos, and Mustafa Mukadam.
\newblock {Theseus: A Library for Differentiable Nonlinear Optimization}.
\newblock {\em NeurIPS}, 2022.

\bibitem{ranftl2018deep}
Ren{\'e} Ranftl and Vladlen Koltun.
\newblock Deep fundamental matrix estimation.
\newblock In {\em ECCV}, 2018.

\bibitem{redmon2016you}
Joseph Redmon, Santosh Divvala, Ross Girshick, and Ali Farhadi.
\newblock You only look once: Unified, real-time object detection.
\newblock In {\em CVPR}, 2016.

\bibitem{ronneberger2015u}
Olaf Ronneberger, Philipp Fischer, and Thomas Brox.
\newblock U-net: Convolutional networks for biomedical image segmentation.
\newblock In {\em International Conference on Medical image computing and
  computer-assisted intervention}. Springer, 2015.

\bibitem{ORB_Rublee2011}
E. {Rublee}, V. {Rabaud}, K. {Konolige}, and G. {Bradski}.
\newblock Orb: An efficient alternative to sift or surf.
\newblock In {\em ICCV}, 2011.

\bibitem{sarlin2020superglue}
Paul-Edouard Sarlin, Daniel DeTone, Tomasz Malisiewicz, and Andrew Rabinovich.
\newblock Superglue: Learning feature matching with graph neural networks.
\newblock In {\em CVPR}, 2020.

\bibitem{sarlin2021back}
Paul-Edouard Sarlin, Ajaykumar Unagar, Mans Larsson, Hugo Germain, Carl Toft,
  Viktor Larsson, Marc Pollefeys, Vincent Lepetit, Lars Hammarstrand, Fredrik
  Kahl, et~al.
\newblock Back to the feature: Learning robust camera localization from pixels
  to pose.
\newblock In {\em CVPR}, 2021.

\bibitem{sheorey2014uncertainty}
Sameer Sheorey, Shalini Keshavamurthy, Huili Yu, Hieu Nguyen, and Clark~N
  Taylor.
\newblock Uncertainty estimation for klt tracking.
\newblock In {\em Asian Conference on Computer Vision}, 2014.

\bibitem{Steele2005Foerster}
R.M. Steele and C. Jaynes.
\newblock Feature uncertainty arising from covariant image noise.
\newblock In {\em CVPR}, 2005.

\bibitem{stewenius2006recent}
Henrik Stewenius, Christopher Engels, and David Nist{\'e}r.
\newblock Recent developments on direct relative orientation.
\newblock {\em ISPRS Journal of Photogrammetry and Remote Sensing}, 60, 2006.

\bibitem{szeliski2010computer}
Richard Szeliski.
\newblock {\em Computer vision: algorithms and applications}.
\newblock Springer Science \& Business Media, 2010.

\bibitem{KLT_Tomasi1991}
Carlo Tomasi and Takeo Kanade.
\newblock Detection and tracking of point features.
\newblock {\em IJCV}, 9, 1991.

\bibitem{torii201524}
Akihiko Torii, Relja Arandjelovic, Josef Sivic, Masatoshi Okutomi, and Tomas
  Pajdla.
\newblock 24/7 place recognition by view synthesis.
\newblock In {\em CVPR}, 2015.

\bibitem{triggs1999bundle}
Bill Triggs, Philip~F McLauchlan, Richard~I Hartley, and Andrew~W Fitzgibbon.
\newblock Bundle adjustment—a modern synthesis.
\newblock In {\em International workshop on vision algorithms}, 1999.

\bibitem{Basalt_Usenko2020}
Vladyslav Usenko, Nikolaus Demmel, David Schubert, J{\"{o}}rg St{\"{u}}ckler,
  and Daniel Cremers.
\newblock Visual-inertial mapping with non-linear factor recovery.
\newblock {\em {{IEEE} Robotics and Automation Letters ({RAL})}}, 5, 2020.

\bibitem{von2020gn}
Lukas Von~Stumberg, Patrick Wenzel, Qadeer Khan, and Daniel Cremers.
\newblock Gn-net: The gauss-newton loss for multi-weather relocalization.
\newblock {\em {{IEEE} Robotics and Automation Letters ({RAL})}}, 2020.

\bibitem{yang20d3vo}
N. Yang, L. von Stumberg, R. Wang, and D. Cremers.
\newblock D3vo: Deep depth, deep pose and deep uncertainty for monocular visual
  odometry.
\newblock In {\em CVPR}, 2020.

\bibitem{yang2018dvso}
N. Yang, R. Wang, J. Stueckler, and D. Cremers.
\newblock Deep virtual stereo odometry: Leveraging deep depth prediction for
  monocular direct sparse odometry.
\newblock In {\em ECCV}, 2018.

\bibitem{yi2016lift}
Kwang~Moo Yi, Eduard Trulls, Vincent Lepetit, and Pascal Fua.
\newblock Lift: Learned invariant feature transform.
\newblock In {\em ECCV}, 2016.

\bibitem{SIFTCOV_Zeisl2009}
Bernhard Zeisl, Pierre Georgel, Florian Schweiger, Eckehard Steinbach, and
  Nassir Navab.
\newblock Estimation of location uncertainty for scale invariant feature
  points.
\newblock In {\em BMVC}, 2009.

\bibitem{zhang2017uncertainty}
Hongmou Zhang, Denis Grie{\ss}bach, J{\"u}rgen Wohlfeil, and Anko B{\"o}rner.
\newblock Uncertainty model for template feature matching.
\newblock In {\em Pacific-Rim Symposium on Image and Video Technology}, pages
  406--420. Springer, 2017.

\end{thebibliography}
}

\end{document}